\documentclass{article} % For LaTeX2e
\usepackage{iclr2027_conference,times}

\usepackage{amsmath,amsfonts,bm}

\def\eqref#1{equation~\ref{#1}}
\def\1{\bm{1}}

\DeclareMathAlphabet{\mathsfit}{\encodingdefault}{\sfdefault}{m}{sl}
\SetMathAlphabet{\mathsfit}{bold}{\encodingdefault}{\sfdefault}{bx}{n}

\usepackage{hyperref}
\usepackage{url}
\usepackage{graphicx}
\usepackage{booktabs}
\usepackage{array}
\usepackage{longtable}
\usepackage{algorithm}
\usepackage{algpseudocode}
\usepackage{amsmath}
\usepackage{amssymb}
\usepackage{titletoc}
\usepackage{etoolbox}
\usepackage{placeins}
\usepackage{tikz}
\usetikzlibrary{arrows.meta,calc}
\usepackage{tcolorbox}
\tcbuselibrary{skins}

\definecolor{foregroundcolor}{HTML}{2A2623}
\definecolor{linkcolor}{HTML}{2A2623}
\definecolor{boxbackground}{HTML}{FAF7F2}
\definecolor{boxborder}{HTML}{DDD3C5}
\definecolor{linkcol}{HTML}{6E4C67}       % muted plum
\definecolor{citecol}{HTML}{8A5A7B}       % dusty magenta
\definecolor{urlcol}{HTML}{3F5E8C}        % muted blue

\hypersetup{
  colorlinks=true,
  linkcolor=linkcol,
  citecolor=citecol,
  urlcolor=urlcol
}

\newcommand{\email}[1]{\href{mailto:#1}{\texttt{#1}}}

\newcommand{\appendixtoc}{
  \clearpage
  \startcontents[appendix]

  \begin{center}
    {\huge\bfseries Zatom-2 Appendices}
    \vspace{8pt}
    \hrule height 1.5pt
    \vspace{16pt}
  \end{center}

  \titlecontents{section}[2em]
    {\addvspace{12pt}\bfseries\sffamily\large}
    {\contentslabel{2em}}
    {}
    {\titlerule*[0.6pc]{.}\contentspage}
    [\addvspace{2pt}]

  \titlecontents{subsection}[4.5em]
    {\addvspace{2pt}\sffamily\small}
    {\contentslabel{2.5em}}
    {}
    {\titlerule*[0.6pc]{.}\contentspage}

  \printcontents[appendix]{}{1}{\setcounter{tocdepth}{2}}

  \vspace{24pt}
  \hrule height 0.5pt
}

\makeatletter

\newcommand\addtolist[5][]{%
  \begingroup
    \if\relax#3\relax
      \def\sep{}%
    \else
      \def\sep{#5}%
    \fi
    \let\protect\@unexpandable@protect
    \xdef#3{\expandafter{#3}\sep #4[#1]{#2}}%
  \endgroup
}

\newcommand\authorlist{}
\newcommand\authorformat[2][]{{\rmfamily\bfseries #2$^{#1}$}}
\renewcommand\author[2][]{\addtolist[#1]{#2}{\authorlist}{\authorformat}{, }}

\newcommand\affiliationlist{}
\newcommand\affiliationformat[2][]{{\normalsize $^{#1}$#2}}
\newcommand\affiliation[2][]{\addtolist[#1]{#2}{\affiliationlist}{\affiliationformat}{, }}

\newcommand\contributionlist{}

\newcommand\metadatalist{}
\newcommand\metadataformat[2][]{{\small\rmfamily\bfseries #1:} {\small #2}}
\newcommand\metadata[2][]{\addtolist[#1]{#2}{\metadatalist}{\metadataformat}{\par\vspace{2mm}}}

\newcommand\abstractlist{}
\renewcommand{\abstract}[1]{\def\abstractlist{{#1}}}

\newcommand\correspondence[1]{\metadata[Correspondence]{#1}}

\newcommand\titlelist{}
\renewcommand{\title}[1]{%
  \def\titlelist{{\rmfamily\bfseries #1}}%
  \gdef\@title{#1}%
}

\newcommand{\@toptitlebar}{%
  {\color{foregroundcolor}\hrule height 0.5pt}
  \vskip 6mm
}

\newcommand{\@bottomtitlebar}{%
  \vskip 6mm
  {\color{foregroundcolor}\hrule height 0.5pt}
}

\newcommand{\mymaketitle}{%
  \thispagestyle{plain}%
  \lhead{Preprint}%
  {%
    \setlength{\parskip}{0mm}%
    \centering

    \noindent
    \begin{minipage}{\textwidth}
        \centering

        \@toptitlebar

        \vspace{1mm}

        {%
          \fontsize{18}{22}\selectfont
          \titlelist\par
        }

        \vspace{1mm}

        \@bottomtitlebar
    \end{minipage}%

    \vskip 5mm

    {\large \authorlist\par}

    \vskip 3mm

    {\normalsize \affiliationlist\par}

    \vskip 2mm

    {\small \contributionlist\par}
  }

  \vskip 5mm

  \begin{tcolorbox}[
    enhanced,
    width=\linewidth,
    colback=boxbackground,
    colframe=boxborder,
    boxrule=0.6pt,
    arc=3.5mm,
    outer arc=3.5mm,
    left=8mm,
    right=8mm,
    top=3mm,
    bottom=6mm,
    drop fuzzy shadow={black!12!white},
    before skip=0pt,
    after skip=0pt
  ]

    \setlength{\parindent}{0pt}
    \setlength{\parskip}{0.55em}

    {\rmfamily\bfseries\color{foregroundcolor} Abstract}

    \vspace{1mm}

    {\color{foregroundcolor}\abstractlist\par}

    \vspace{1mm}

    \ifdefempty{\metadatalist}
      {}
      {%
        \vspace{1mm}
        {\small \metadatalist\par}%
      }

  \end{tcolorbox}

  \vskip 7mm
}

\renewcommand{\maketitle}{\mymaketitle}

\makeatother

\title{Zatom-2: Multitask Pretraining on Atomistic Data for Generative Modeling across Domains}

\author[1,*]{Miruna Cretu}
\author[1,\dagger]{Alex Abrudan}
\author[2,\dagger]{Antonia Panescu}
\author[3,\dagger]{Tynan Perez}
\author[3,\dagger]{Rishabh Anand}
\author[4,5]{N.~Benjamin Erichson}
\author[4,5,6]{Michael W.~Mahoney}
\author[4]{Samuel Blau}
\author[3]{Joseph Jacobson}
\author[3]{Rafael G\'omez-Bombarelli}
\author[2]{Rex Ying}
\author[1]{Tuomas Knowles}
\author[1]{Pietro Li\`{o}}
\author[4,5,*]{Alex Morehead}

\affiliation[1]{University of Cambridge}
\affiliation[2]{Yale University}
\affiliation[3]{MIT}
\affiliation[4]{Lawrence Berkeley National Lab}
\affiliation[5]{International Computer Science Institute}
\affiliation[6]{UC Berkeley}

\correspondence{\email{mtc49@cam.ac.uk}; and \email{acmwhb@lbl.gov}}
\metadata[Project Page]{\url{https://zatom-ai.github.io/nucleus}}

\abstract{Unified atomistic modeling has the potential to accelerate discovery in chemistry, materials science, and biology by bridging data-rich chemical domains and data-scarce biological contexts. However, existing generative approaches to atomistic modeling remain highly specialized to scientific disciplines (chemistry vs. biology) or do not leverage both high-volume organic (molecule) and inorganic (material) data for general-purpose pretraining. To this end, we introduce Zatom-2, an atomistic generative model pretrained on approximately five million structures from the OMol25 and OMat24 electronic structure datasets. Zatom-2 features a multiscale Transformer architecture coupled with conditional flow matching that supports force conditioning and foundational pretraining tasks such as generation, structure prediction, and prediction of molecular and material energies and forces. Empirically, Zatom-2 achieves better molecular distribution fidelity than Zatom-1 and achieves strong performance on existing molecule and material generation benchmarks. Zatom-2 demonstrates the ability to control sample generation across low- and high-force regimes, and enhances protein generation in a low-data setting through joint generative-predictive pretraining and transfer learning, increasing protein backbone designability in a length extrapolation setting from 67.8\% without pretraining to 74.8\% after finetuning on 2,000 protein domains.}

\iclrfinalcopy % Uncomment for camera-ready version, but NOT for submission.
\begin{document}

\maketitle

\begingroup
\renewcommand{\thefootnote}{*}
\footnotetext{Equally contributing authors: Miruna Cretu (lead author) and Alex Morehead (senior author).}
\renewcommand{\thefootnote}{$\dagger$}
\footnotetext{Equal core computational contributors.}
\endgroup

\section{Introduction}

Atomistic machine learning has made significant progress across chemistry, materials science, and structural biology, yet these domains have largely developed separate modeling frameworks. For small molecular and materials systems, generative models operate directly on atom identities and three-dimensional coordinates~\citep{hoogeboom2022equivariant,tabasco,jiao2023crystal,zeni2025generative}. In contrast, models like RFdiffusion3 and BoltzGen~\citep{butcher2025rfdiffusion3,stark2025boltzgen} employ protein-specific tokenization to make all-atom generation tractable without a separate diffusion process over amino acid identities, while La-Proteina encodes sequence and side-chain information into learned per-residue latent variables \citep{laproteina}. These approaches are effective within their respective domains, but they do not explore training and finetuning a single model across molecules, periodic materials, and biomolecules. 
This separation is not intrinsic to the underlying physical systems. Molecules, materials, and proteins are all collections of interacting atoms in three-dimensional space, and many of the local environments that determine their structure and interactions recur across systems of varying size and type.

Recent large-scale quantum-chemical datasets make shared atomistic modeling increasingly practical. In particular, Open Molecules 2025 (OMol25) contains more than 100 million high-fidelity density functional theory calculations \citep{levine2025omol25}, and its diversity extends along two complementary axes. Chemically, OMol25 covers small organic molecules, biomolecules, metal complexes, electrolytes, varying charge and spin states, and a broad range of intra- and intermolecular interactions. Geometrically, it contains diverse conformers and reactive structures rather than restricting supervision to equilibrium geometries. Open Materials 2024 (OMat24) provides an analogous resource for periodic systems, comprising more than 110 million calculations over diverse inorganic compositions and configurations \citep{barrosoluque2024omat24}. Crucially, both datasets provide energy and force labels in addition to atomic structures.

We argue that the diversity of OMol25 creates two exciting opportunities for generative modeling. First, its broad chemical coverage makes it possible to move beyond the narrow distributions that dominate existing molecular generation benchmarks and train models capable of directly generating chemically diverse and industrially relevant systems. Second, its compositional, conformational, and energy diversity makes it a natural pretraining corpus for transfer to settings where structural data are scarce. The latter is particularly compelling for biomolecular generation: although quantum-chemical datasets predominantly contain much smaller systems than proteins and DNA/RNA structures, they provide dense supervision over the local atomic environments and geometric interactions from which larger structures are composed. This motivates a central question of our work: \emph{how should such large-scale atomistic data be used to learn representations that transfer effectively across both tasks and domains?}

To this end, we introduce \textbf{Zatom-2}, a unified atomistic architecture and pretraining framework for generative modeling across molecules, periodic materials and proteins. Uniquely, Zatom-2's pretraining curriculum includes the OMol25 (4M) and OMat24 (1M) datasets, and combines generation and structure prediction with auxiliary force and energy prediction objectives (see Figure~\ref{fig:zatom2-overview}). The latter objective is motivated by evidence that models trained to learn interatomic potentials (known as machine learning interatomic potentials, MLIPs)~\citep{Batatia2022mace,wood2025uma,neumann2024orb,qu2026allscaip}, have been demonstrated to learn meaningful, transferable representations of geometric systems~\citep{wedig2025rem3di,li2026platonic,didi2026r4g}, and we hypothesize that these auxiliary physics-grounded objectives help shape a generative model's latent representations in a way that is beneficial for downstream tasks. In particular, force and energy supervision encourages a backbone architecture to encode the local geometric structure of atomistic configurations influenced by the forces acting on the system. We further discuss related work in Appendix~\ref{app:related-work}.

\textbf{Contributions.} Based on these ideas, our main contributions in this work are as follows:
\begin{itemize}
    \item We introduce Zatom-2, a \emph{unified} model for generative modeling across atomistic domains, which supports molecule and material generation, structure prediction, and prediction of energies and forces by way of pretraining. Its architecture additionally supports protein generation, which we demonstrate in a finetuning task.
    \item Zatom-2: (\textbf{1}) achieves better molecular distribution learning than Zatom-1; (\textbf{2}) provides strong performance in molecule and material generation benchmarks; (\textbf{3}) introduces force conditioning to enable \emph{control} over generated samples' relaxation state; and (\textbf{4}) demonstrates transfer of atomistic pretraining to protein generation, including strong extrapolation to sequence lengths beyond the training distribution.
    \item The Zatom-2 architecture features a unified \texttt{atom1} + \texttt{atom14} tokenization scheme for atomistic data, thereby enabling molecule, material, and all-atom protein generative modeling tasks within a single, standardized Transformer architecture which demonstrates performance advantages with data and model scaling.
\end{itemize}

\section{Preliminaries}
\label{sec:preliminaries}

\subsection{All-atom representation and modeling}
\label{sec:all-atom-representation-and-modeling}
Diffusion models for small molecules and periodic materials often co-generate geometry and atom identity using separate corruption processes for coordinates and atom types \citep{vignac2023midi,zeni2025generative,joshi2025allatom,tabasco}. Our formulation draws inspiration from the protein literature \citep{qu2025pallatom, butcher2025rfdiffusion3}, which models all-atom protein coordinates and predicts residue identities from learned geometric representations. To accommodate variation in the number of atoms per residue, \citet{qu2025pallatom} introduces an \texttt{atom14} representation: each residue occupies 14 coordinate slots, with unused slots filled by virtual atoms. A protein with $L$ residues is thus represented by $\mathbf{X}_0 \in \mathbb{R}^{L \times 14 \times 3}$, whose shape is independent of residue identity. A sequence prediction head then recovers residue identities from the model's atom-level features, which enables coordinate diffusion without a separate diffusion process over amino acid types. 

Building on this principle, we pose the question, similarly to proteins, of whether molecule and material coordinates hold enough information to inform atom type prediction directly, and introduce an \texttt{atom1} representation for molecular systems, in which each atom constitutes a token with a three-dimensional coordinate. Like \citet{feng2025unigem}, we generate atomic coordinates through diffusion and infer atom identities from representations learned by the denoising network. Distinctly, our formulation enables a shared all-atom generation framework for biological (e.g., protein) and chemical (e.g., molecule, material) data by allowing residue- and atom-level tokenization, respectively.

% Zatom-2 overview with warm section panels and a teal computational backbone.
% Requires amsmath, amssymb, graphicx, tikz, and the arrows.meta and calc libraries.
\begin{figure*}[t]
  \centering
  \begingroup
  \definecolor{zInk}{HTML}{253A3C}
  \definecolor{zMuted}{HTML}{62676B}
  \definecolor{zGray}{HTML}{C8CCC8}
  \definecolor{zTeal}{HTML}{387E80}
  \definecolor{zTealFill}{HTML}{E4F0ED}
  \definecolor{zClay}{HTML}{AE6B51}
  \definecolor{zClayFill}{HTML}{F5E8DF}
  \definecolor{zOchre}{HTML}{A28038}
  \definecolor{zOchreFill}{HTML}{F4EDDA}
  \definecolor{zPanel}{HTML}{F8F7F3}
  \definecolor{zBackbonePanel}{HTML}{F4F8F6}

  \tikzset{
    zBox/.style={
      draw=zGray, fill=white, line width=0.55pt,
      rounded corners=3pt, inner xsep=6pt, inner ysep=4pt,
      align=center, text=zInk
    },
    zInput/.style={
      zBox, minimum width=3.80cm, minimum height=1.18cm,
      font=\sffamily\fontsize{10.5}{12}\selectfont
    },
    zTaskMark/.style={
      font=\fontsize{12}{12}\selectfont, inner sep=0pt, outer sep=0pt
    },
    zMaterialMark/.style={
      font=\sffamily\bfseries\fontsize{10}{11}\selectfont, inner sep=0pt
    },
    zFlow/.style={
      -{Latex[length=1.8mm,width=1.2mm]},
      draw=zInk!85!white, line width=0.75pt, rounded corners=3pt
    },
    zCondition/.style={
      zFlow, draw=zTeal, dash pattern=on 0.4pt off 2.2pt,
      line cap=round, line width=0.7pt
    },
    zFeedback/.style={
      zFlow, draw=zClay, dashed, line width=0.75pt
    },
    zStack/.style={
      zBox, draw=zTeal!70!white, fill=zTealFill,
      minimum width=2.80cm, minimum height=1.70cm,
      font=\sffamily\fontsize{10.8}{12.5}\selectfont
    },
    zHead/.style={
      zBox, draw=zTeal!65!white, fill=white,
      minimum width=4.15cm, minimum height=0.74cm,
      font=\sffamily\fontsize{10.8}{12.5}\selectfont
    },
    zProperty/.style={zHead, draw=zOchre!75!white, fill=zOchreFill},
    zAtomDot/.style={
      circle, draw=white, line width=0.35pt, fill=zTeal,
      minimum size=4.8pt, inner sep=0pt
    },
    zHeading/.style={
      font=\sffamily\bfseries\fontsize{12}{13.5}\selectfont,
      text=zInk
    },
    zNote/.style={
      font=\sffamily\fontsize{9.5}{11}\selectfont,
      text=zMuted, align=center, inner sep=1.5pt
    },
    zDatasetTitle/.style={
      zNote, text=black, font=\sffamily\bfseries\fontsize{10}{11.5}\selectfont
    },
    zDatasetLabel/.style={
      zNote, text=zInk, font=\sffamily\fontsize{8.5}{10}\selectfont
    },
    zDatasetIcon/.style={y=1cm,scale=0.90,transform shape}
  }

  \resizebox{\textwidth}{!}{%
  \begin{tikzpicture}[
    % Expand vertical spacing while keeping text and dataset icons proportional.
    x=1cm, y=1.25cm,
    font=\sffamily\fontsize{10}{11.5}\selectfont,
    text=zInk
  ]
    % Symmetric about x=9.225, the center of the token trunk.
    \useasboundingbox (-4.80,0.30) rectangle (23.25,9.40);

    % Split the existing left-hand space without changing overall proportions.
    % Quiet panels separate dataset coverage, inputs, computation, and readouts.
    \path[fill=zPanel,rounded corners=6pt]
      (-4.65,0.80) rectangle (-1.45,8.50);
    \path[fill=zPanel,rounded corners=6pt]
      (-1.15,0.80) rectangle (3.55,8.50);
    \path[fill=zBackbonePanel,rounded corners=6pt]
      (4.50,0.80) rectangle (13.95,8.50);
    \path[fill=zPanel,rounded corners=6pt]
      (14.90,0.80) rectangle (23.10,8.50);
    \node[zHeading] at (-3.05,9.00) {Datasets};
    \node[zHeading] at (1.20,9.00) {Conditioning inputs};
    \node[zHeading] at (9.225,9.00) {Multitask transformer backbone};
    \node[zHeading] at (19.00,9.00) {Task-specific readouts};
    \draw[zMuted,line width=1.3pt] (-4.40,8.65) -- (-1.70,8.65);
    \draw[zClay,line width=1.3pt] (-0.95,8.65) -- (3.15,8.65);
    \draw[zTeal,line width=1.3pt] (4.90,8.65) -- (13.55,8.65);
    \draw[zOchre,line width=1.3pt] (15.30,8.65) -- (22.70,8.65);

    % Schematic coverage examples, not additional model conditioning labels.
    \node[zDatasetTitle] at (-3.05,8.18) {OMol25 (4M)};

    % Biomolecule: a single cropped protein ribbon around a small ligand.
    \begin{scope}[shift={(-3.05,7.66)},zDatasetIcon]
      \clip (-1.12,-0.44) rectangle (1.12,0.33);
      \path[fill=zTeal!55!white,draw=zTeal,line width=0.5pt]
        (-1.30,0.30) .. controls (-0.83,0.43) and (-0.55,0.29) .. (-0.55,0.06)
        .. controls (-0.58,-0.28) and (-0.31,-0.37) .. (0.08,-0.37)
        .. controls (0.49,-0.37) and (0.68,-0.17) .. (1.30,-0.15)
        --(1.30,-0.30)
        .. controls (0.79,-0.30) and (0.56,-0.54) .. (0.05,-0.52)
        .. controls (-0.49,-0.52) and (-0.75,-0.26) .. (-0.74,0.02)
        .. controls (-0.73,0.20) and (-0.94,0.19) .. (-1.30,0.13)--cycle;
      % A clear gap separates the warm ligand from the teal binding pocket.
      \draw[zOchre,line width=0.95pt]
        (-0.27,0.10)--(-0.06,0.21)--(0.15,0.10)--(0.15,-0.10)
        --(-0.06,-0.21)--(-0.27,-0.10)--cycle
        (0.15,0.10)--(0.41,0.18);
      \foreach \x/\y in {-0.27/0.10,-0.06/0.21,0.15/0.10,-0.06/-0.21,-0.27/-0.10} {
        \node[zAtomDot,fill=zOchre,minimum size=3.8pt,line width=0.25pt] at (\x,\y) {};
      }
      \node[zAtomDot,fill=zClay,minimum size=4.2pt,line width=0.25pt] at (0.15,-0.10) {};
      \node[zAtomDot,fill=zClay,minimum size=4.2pt,line width=0.25pt] at (0.41,0.18) {};
    \end{scope}
    \node[zDatasetLabel] at (-3.05,7.08) {Biomolecules};

    % Drug-like molecule: a substituted heterocycle.
    \begin{scope}[shift={(-3.05,6.47)},zDatasetIcon,scale=0.90]
      \coordinate (m1) at (-0.35,0.34);
      \coordinate (m2) at (-0.05,0.17);
      \coordinate (m3) at (-0.05,-0.17);
      \coordinate (m4) at (-0.35,-0.34);
      \coordinate (m5) at (-0.65,-0.17);
      \coordinate (m6) at (-0.65,0.17);
      \draw[zMuted,line width=0.9pt]
        (m1)--(m2)--(m3)--(m4)--(m5)--(m6)--cycle
        (m2)--(0.28,0.32)--(0.59,0.12)--(0.89,0.28);
      \foreach \atom in {m1,m2,m3,m4,m6} {
        \node[zAtomDot] at (\atom) {};
      }
      \node[zAtomDot,fill=zClay] at (m5) {};
      \node[zAtomDot] at (0.28,0.32) {};
      \node[zAtomDot,fill=zClay] at (0.59,0.12) {};
      \node[zAtomDot,fill=zOchre] at (0.89,0.28) {};
    \end{scope}
    \node[zDatasetLabel] at (-3.05,5.94) {Drug-like molecules};

    % Electrolyte: separated ions surrounded by small solvent motifs.
    \begin{scope}[shift={(-3.05,5.33)},zDatasetIcon]
      \node[circle,draw=zTeal,fill=zTealFill,inner sep=1pt,
        minimum size=0.36cm,font=\fontsize{10}{10}\selectfont] at (-0.35,0) {$+$};
      \node[circle,draw=zClay,fill=zClayFill,inner sep=1pt,
        minimum size=0.36cm,font=\fontsize{10}{10}\selectfont] at (0.40,0) {$-$};
      \foreach \x/\y/\angle in {-0.87/0.17/30,-0.60/-0.28/160,0.02/0.27/-20,0.91/-0.16/200} {
        \begin{scope}[shift={(\x,\y)},rotate=\angle]
          \draw[zMuted,line width=0.7pt] (-0.13,0.10)--(0,0)--(0.13,0.10);
          \node[zAtomDot,fill=zClay,minimum size=3.5pt] at (0,0) {};
        \end{scope}
      }
    \end{scope}
    \node[zDatasetLabel] at (-3.05,4.80) {Electrolytes};

    % Metal complex: a central metal coordinated by four donor groups.
    \begin{scope}[shift={(-3.05,4.19)},zDatasetIcon]
      \foreach \x/\y in {-0.58/0.24,0.58/0.24,-0.58/-0.24,0.58/-0.24} {
        \draw[zMuted,line width=0.8pt] (0,0)--(\x,\y);
        \node[zAtomDot,fill=zClay] at (\x,\y) {};
      }
      \draw[zMuted,line width=0.8pt]
        (-0.58,0.24)--(-0.90,0.11) (0.58,0.24)--(0.90,0.11)
        (-0.58,-0.24)--(-0.90,-0.11) (0.58,-0.24)--(0.90,-0.11);
      \node[zAtomDot,fill=zOchre,minimum size=7pt] at (0,0) {};
    \end{scope}
    \node[zDatasetLabel] at (-3.05,3.66) {Metal complexes};

    % Transition state: partial bonds and the conventional double dagger.
    \begin{scope}[shift={(-3.05,3.05)},zDatasetIcon]
      \draw[zMuted,line width=0.9pt,dash pattern=on 2pt off 1.5pt]
        (-0.68,0)--(0,0.14)--(0.68,0);
      \node[zAtomDot,fill=zClay,minimum size=6pt] at (-0.68,0) {};
      \node[zAtomDot,minimum size=5.4pt] at (0,0.14) {};
      \node[zAtomDot,fill=zClay,minimum size=6pt] at (0.68,0) {};
      \draw[zMuted,line width=0.6pt]
        (-0.85,0.32)--(-1.00,0.32)--(-1.00,-0.23)--(-0.85,-0.23)
        (0.85,0.32)--(1.00,0.32)--(1.00,-0.23)--(0.85,-0.23);
      \node[inner sep=0pt,font=\fontsize{10}{10}\selectfont] at (1.15,0.30) {$\ddagger$};
    \end{scope}
    \node[zDatasetLabel] at (-3.05,2.52) {Transition states};

    \draw[zGray,line width=0.5pt] (-4.35,2.28)--(-1.75,2.28);
    \node[zDatasetTitle] at (-3.05,2.04) {OMat24 (1M)};

    % Periodic material: a projected unit cell with repeated lattice sites.
    \begin{scope}[shift={(-3.17,1.46)},zDatasetIcon]
      \coordinate (c1) at (-0.48,-0.147);
      \coordinate (c2) at (0.30,-0.147);
      \coordinate (c3) at (0.30,0.175);
      \coordinate (c4) at (-0.48,0.175);
      \coordinate (d1) at (-0.18,-0.014);
      \coordinate (d2) at (0.60,-0.014);
      \coordinate (d3) at (0.60,0.308);
      \coordinate (d4) at (-0.18,0.308);
      \draw[zGray,line width=0.7pt] (d1)--(d2)--(d3)--(d4)--cycle
        (c1)--(d1) (c2)--(d2) (c3)--(d3) (c4)--(d4);
      \draw[zMuted,line width=0.8pt] (c1)--(c2)--(c3)--(c4)--cycle;
      \foreach \atom in {c1,c3,d2,d4} {
        \node[zAtomDot,minimum size=4.2pt] at (\atom) {};
      }
      \foreach \atom in {c2,c4,d1,d3} {
        \node[zAtomDot,fill=zClay,minimum size=4.2pt] at (\atom) {};
      }
    \end{scope}
    \node[zDatasetLabel] at (-3.05,1.00) {Periodic materials};

    % Task strips expose which metadata the initializer receives.
    % Atom identities and molecular bonds have different MLIP masks.
    \node[zNote] at (1.20,8.22)
      {$\checkmark$ on\quad -- off};
    \foreach \inputnode/\inputlabel/\height/\genmark/\structmark/\mlipmark in {
      domain/{Molecule / material}/7.45/\checkmark/\checkmark/\checkmark,
      electronic/{Charge \& spin (Mol.)}/6.31/\checkmark/\checkmark/\checkmark,
      forcecond/{System force}/5.17/\checkmark/\checkmark/{-},
      chemistry/{Atom types}/4.03/{-}/\checkmark/\checkmark,
      bonds/{Bonds (Mol.)}/2.89/{-}/\checkmark/{-}} {
      \node[zInput] (\inputnode) at (0.95,\height) {};
      \node[font=\sffamily\fontsize{10.5}{12}\selectfont,inner sep=0pt]
        at (0.95,\height+0.20) {\inputlabel};
      \node[text=zTeal,font=\sffamily\fontsize{8.5}{10}\selectfont,inner sep=0pt]
        at (-0.25,\height-0.22) {Gen.\ $\genmark$};
      \node[text=zClay,font=\sffamily\fontsize{8.5}{10}\selectfont,inner sep=0pt]
        at (0.95,\height-0.22) {Struct.\ $\structmark$};
      \node[text=zOchre,font=\sffamily\fontsize{8.5}{10}\selectfont,inner sep=0pt]
        at (2.15,\height-0.22) {MLIP\ $\mlipmark$};
    }

    % Diffusion-state input; clean MLIP coordinates are specified in the caption.
    \node[zBox, draw=zClay!75!white, fill=zClayFill,
      minimum width=3.80cm, minimum height=0.96cm,
      font=\sffamily\fontsize{10.5}{12}\selectfont]
      (flowstate) at (0.95,1.63)
      {\textbf{Noisy coordinates}\\[2pt]
       $\mathbf{x}_t$ \& $t$};

    \node[zNote] at (1.20,0.53) {Mol.: molecule only};
    \node[zNote] at (21.40,0.53) {Mat.: materials only};

    % Lift the backbone and readout contents together within the fixed panels.
    \begin{scope}[shift={(0,0.25)}]
    % The initializer supplies the atom/token conditioning and pair features.
    \node[zBox, draw=zClay!75!white, fill=zClayFill,
      minimum width=8.50cm, minimum height=1.20cm]
      (initializer) at (9.25,7.00) {};
    \node[anchor=west, align=left,
      font=\sffamily\bfseries\fontsize{10.8}{12.5}\selectfont]
      at (5.20,7.00) {Input initializer};
    \node[align=center, inner sep=3pt,
      font=\sffamily\fontsize{9.5}{11}\selectfont]
      (atomContext) at (9.50,7.00) {Atom features\\[3pt]$C_L,\,P_{LL}$};
    \node[align=center, inner sep=3pt,
      font=\sffamily\fontsize{9.5}{11}\selectfont]
      (tokenContext) at (12.20,7.00) {Token features\\[3pt]$S_I,\,Z_{II}$};
    \draw[zClay!35!white,line width=0.5pt] (8.30,6.65) -- (8.30,7.35);

    % Available conditioning inputs enter the initializer together.
    \foreach \inputnode in {domain,electronic,forcecond,chemistry,bonds} {
      \draw[draw=zInk,line width=0.65pt]
        (\inputnode.east) -- (3.25,0 |- \inputnode.east);
    }
    \draw[draw=zInk,line width=0.65pt]
      (3.25,0 |- bonds.east) -- (3.25,0 |- domain.east);
    \draw[zFlow] (3.25,7.00) -- (initializer.west);

    % Flat cards retain explicit depth labels; the trunk anchors the U.
    \node[zStack] (atomEncoder) at (6.10,4.90)
      {\textbf{Atom encoder}\\[5pt]
       $3$ layers\\[3pt]
       {\color{zMuted}Local attention}};
    \node[zStack,minimum width=3.60cm,minimum height=1.85cm,
      draw=zTeal,fill=zTeal,text=white]
      (tokenTrunk) at (9.225,2.65)
      {\textbf{Token trunk}\\[5pt]
       $N$ layers\\[3pt]
       Global attention};
    \node[zStack] (atomDecoder) at (12.25,4.90)
      {\textbf{Atom decoder}\\[5pt]
       $3$ layers\\[3pt]
       {\color{zMuted}Local attention}};

    \draw[zCondition] (atomContext.south) -- (9.50,6.12) -| (atomEncoder.north);
    \draw[zCondition] (9.50,6.12) -| (atomDecoder.north);
    \draw[draw=zBackbonePanel,line width=2.5pt,rounded corners=3pt]
      (tokenContext.south) -- (12.20,6.32) --
      (9.95,6.32) -- (9.95,3.98) -| (tokenTrunk.north);
    \draw[zCondition]
      (tokenContext.south) -- (12.20,6.32) --
      (9.95,6.32) -- (9.95,3.98) -| (tokenTrunk.north);

    % Residual cross-attention carries information down and back up the U.
    \draw[zFlow] (atomEncoder.south) -- (tokenTrunk.west);
    \draw[zFlow] (tokenTrunk.east) -- (atomDecoder.south);

    % Coordinates and time enter at both atom and pooled-token resolution.
    \draw[zFlow] (flowstate.east) -- (3.00,0 |- flowstate.east)
      .. controls ($(flowstate.east)+(1.25,0)$) and (4.30,2.90) .. (4.30,4.40)
      .. controls (4.30,4.75) and (4.35,4.90) .. (atomEncoder.west);
    \draw[zFlow] ([yshift=-5pt]flowstate.east)
      .. controls (4.00,1.15) and (6.20,1.30) .. (tokenTrunk.west);
    \node[zNote] at (5.85,0.99) {Pooled $\mathbf{x}_t$, $t$};

    % Readout color distinguishes geometry/identity from optional properties.
    % Separate buses, not the colors, indicate atom versus token features.
    \node[zHead,minimum height=0.98cm] (velocityHead) at (17.30,6.95)
      {Euclidean geometry\\[1pt]
       {\fontsize{9.2}{10.6}\selectfont ($\hat{v}_\theta$ prediction)}};
    \node[zHead] (fractionalHead) at (17.30,5.88)
      {Fractional coordinates};
    \node[zProperty] (forceHead) at (17.30,4.81)
      {Force vectors};
    \node[zHead] (identityHead) at (17.30,3.67)
      {Atom types / sequence};
    \node[zHead] (latticeHead) at (17.30,2.60)
      {Lattice parameters};
    \node[zProperty] (energyHead) at (17.30,1.53)
      {Energy};

    % Pale task contours share the readout rows, without extra connectors.
    % Checkmarks apply to molecules and materials; Mat. marks materials only.
    \foreach \x/\taskcolor/\taskfill/\tasklabel in {
      20.30/zTeal/zTealFill/Gen.,
      21.40/zClay/zClayFill/Struct.,
      22.50/zOchre/zOchreFill/MLIP} {
      \path[draw=\taskcolor!50!white,fill=\taskfill!65!white,
        line width=0.55pt,rounded corners=4pt]
        (\x-0.46,1.02) rectangle (\x+0.46,7.51);
      \node[zNote,text=\taskcolor,
        font=\sffamily\bfseries\fontsize{9.5}{11}\selectfont]
        at (\x,7.88) {\tasklabel};
    }
    \foreach \head in {velocityHead,identityHead} {
      \node[zTaskMark,text=zTeal] at (20.30,0 |- \head.center) {$\checkmark$};
    }
    \node[zTaskMark,text=zClay] at (21.40,0 |- velocityHead.center) {$\checkmark$};
    \foreach \head in {fractionalHead,latticeHead} {
      \node[zMaterialMark,text=zTeal] at (20.30,0 |- \head.center) {Mat.};
      \node[zMaterialMark,text=zClay] at (21.40,0 |- \head.center) {Mat.};
    }
    \foreach \head in {forceHead,energyHead} {
      \node[zTaskMark,text=zOchre] at (22.50,0 |- \head.center) {$\checkmark$};
    }
    \coordinate (atomReadout) at (14.50,0 |- forceHead.west);
    \draw[zFlow,-] (atomDecoder.east |- forceHead.west) --
      node[above,inner sep=2pt] {$Q_L$} (atomReadout);
    \draw[draw=zInk,line width=0.7pt] (14.50,4.81) -- (14.50,6.95);
    \foreach \head/\height in {velocityHead/6.95,fractionalHead/5.88,forceHead/4.81} {
      \draw[zFlow] (14.50,\height) -- (\head.west);
    }

    \coordinate (tokenReadout) at (14.50,0 |- latticeHead.west);
    \draw[zFlow,-] (tokenTrunk.east |- latticeHead.west) --
      node[pos=0.60,below,inner sep=2pt] {$Q_I$} (tokenReadout);
    \draw[draw=zInk,line width=0.7pt] (14.50,1.53) -- (14.50,3.67);
    \foreach \head/\height in {identityHead/3.67,latticeHead/2.60,energyHead/1.53} {
      \draw[zFlow] (14.50,\height) -- (\head.west);
    }

    % The recycled signal is the endpoint distogram, not hidden features.
    \draw[zFeedback] (velocityHead.north) -- (17.30,7.82) --
      node[pos=0.50,above,inner sep=2pt,text=zClay,
        font=\sffamily\fontsize{9.5}{11}\selectfont]
        {Distogram recycling}
      (12.20,7.82) -- (tokenContext.north);
    \end{scope}

  \end{tikzpicture}%
  }

  \caption{\textbf{Zatom-2 multitask pretraining across domains.}
  Zatom-2 is jointly trained on OMol25 (4M) and OMat24 (1M) as a generative model. Its input initializer receives task-dependent conditioning, summarized by task strips: checkmarks indicate enabled inputs and dashes indicate disabled inputs. 
  % Mol.\ denotes molecule-specific inputs. 
  % Domain labels are supplied for all tasks, as are charges and spin multiplicities for OMol25. System force conditioning is optional for generation and structure prediction in both domains. Atom types are supplied for structure prediction and MLIP; molecular bonds are supplied only for structure prediction. 
  MLIP task supervision (force and energy prediction) uses clean coordinates at $t=1$ and known atom types, without bond or system force conditioning. The initializer provides atom $(C_L,P_{LL})$ and token $(S_I,Z_{II})$ conditioning through the dotted connections. Noisy coordinates $\mathbf{x}_t$ and flow time $t$ enter the U-shaped atom encoder---token trunk---atom decoder.
  % Readouts from atom features $Q_L$ predict normalized Cartesian velocities
  % $\widehat{\mathbf{v}}_\theta$, fractional coordinates, and force vectors.
  % Readouts from token features $Q_I$ predict atom types/sequences, lattice
  % parameters, and energy. 
  Atom features $Q_L$ and token features $Q_I$ feed task-specific readout heads. 
  % The outlined task columns mark primary readouts; Mat.\ denotes material-specific outputs. Materials also use the Cartesian velocity prediction for sampling. 
  The recycling path converts the clean-coordinate estimate obtained from $\widehat{\mathbf{v}}_\theta$ into a token-level distogram, updating token-pair conditioning for a second forward pass. In this work, we experiment with $N=9/18/36$ trunk layers.}
  \label{fig:zatom2-overview}
  \endgroup
\end{figure*}
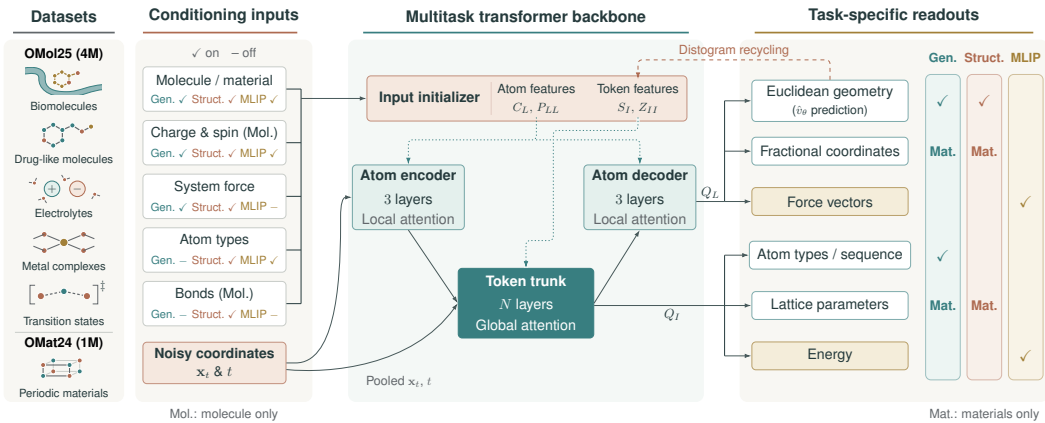

\subsection{Conditional flow matching and sampling}
\label{sec:zatom2-flow}

Flow matching generative models \citep{lipman2023flow,albergo} learn a continuous transport from a noise distribution at $t=0$ to the data distribution at $t=1$ through an ordinary differential equation (ODE). We adopt velocity prediction \citep{simplefold,laproteina}, with a network $\widehat{\mathbf{v}}_\theta(\mathbf{x}_t,t)$ that predicts a velocity normalized by a scale $\sigma_{\mathrm{data}}$. Given clean Cartesian coordinates $\mathbf{x}_1$ and a noise sample $\mathbf{x}_0=\sigma_{\mathrm{data}}\boldsymbol{\epsilon}$, where $\boldsymbol{\epsilon}\sim\mathcal{N}(\mathbf{0},\mathbf{I})$ and $\sigma_{\mathrm{data}}=16\,\text{\AA}$, we define the interpolation path and normalized velocity target as

\begin{equation}
\mathbf{x}_t=(1-t)\mathbf{x}_0+t\mathbf{x}_1,
\qquad
\mathbf{v}_t=
\frac{\mathbf{x}_1-\mathbf{x}_0}{\sigma_{\mathrm{data}}}.
\label{eq:zatom2-flow}
\end{equation}

The corresponding Cartesian velocity field is
$\sigma_{\mathrm{data}}\widehat{\mathbf{v}}_\theta(\mathbf{x}_t,t)$,
yielding the clean endpoint estimate

\begin{equation}
\widehat{\mathbf{x}}_1
=
\mathbf{x}_t
+
(1-t)\sigma_{\mathrm{data}}
\widehat{\mathbf{v}}_\theta(\mathbf{x}_t,t).
\label{eq:zatom2-endpoint}
\end{equation}

The model is trained via the $l_2$ regression objective 
$\mathbb{E}_{t,\mathbf{x}_0,\mathbf{x}_1}\!\left[
\left\|\widehat{\mathbf{v}}_\theta(\mathbf{x}_t,t)-\mathbf{v}_t\right\|_2^2
\right]$.

For inference, let $\mathbf{y}_\sigma$ denote noisy Cartesian coordinates at noise level
$\sigma$, corresponding to the additive corruption
$\mathbf{y}_\sigma=\mathbf{x}_1+\sigma\boldsymbol{\epsilon}$.
We map these coordinates to the flow parameterization as

\begin{equation}
t=\frac{\sigma_{\mathrm{data}}}{\sigma_{\mathrm{data}}+\sigma},
\qquad
\mathbf{x}_t=t\,\mathbf{y}_\sigma.
\label{eq:edm-flow-adapter}
\end{equation}

Since $t\sigma=(1-t)\sigma_{\mathrm{data}}$, this transformation recovers
the interpolation path in Equation~\ref{eq:zatom2-flow}. At each sampling step,
we evaluate the velocity network at $(\mathbf{x}_t,t)$ and supply the sampler
with the clean-coordinate estimate from Equation~\ref{eq:zatom2-endpoint}.
This change of variables enables EDM-style sampling\citep{karras2022edm,butcher2025rfdiffusion3,williams2026emyx} (n.b., which we observed to yield improved sample quality) while retaining a standard velocity-based flow matching training objective.

\section{Zatom-2}
\label{sec:zatom2}

\subsection{Multiscale Transformer backbone}
\label{sec:zatom2-backbone}

We formulate an architecture that supports multitask learning across generation, structure prediction, and property prediction tasks for molecules and periodic materials, as well as generation for proteins. To this end, we adopt our new multiscale tokenization scheme (Section~\ref{sec:all-atom-representation-and-modeling}) and define multitask input conditioning features, as well as bespoke output heads for each task and data domain. 

Zatom-2 contains an input initializer and a U-shaped atom--token--atom Transformer (Figure~\ref{fig:zatom2-overview}), similar to the multiscale
architectures
of RFdiffusion3 and Emyx \citep{butcher2025rfdiffusion3,williams2026emyx}; Appendix~\ref{app:additional-model-details} details Zatom-2's forward pass and compares it to that of RFdiffusion3 and Emyx. The model learns atom representations $Q_L$ for local geometric interactions and token representations $Q_I$ for global information exchange, where $L$ and $I$ denote atoms and tokens, respectively.

The initializer embeds a series of features: domain labels, spins and charges of the system, system force labels (see Section~\ref{sec:force-conditioning}), as well as fixed chemical identities, relative token positions and bonds for molecule structure prediction tasks. These embeddings are combined with coordinate and flow-time embeddings to initialize the atom and token representations and construct the conditioning features used throughout the denoising backbone (see Figure~\ref{fig:zatom2-overview}). 

The atom encoder updates atom features through local attention over token-index neighborhoods and nearby atoms in space, then downcasts these features into token representations through cross-attention. The Transformer trunk applies global attention between tokens, before upcasting the updated token features back to the atom stream through cross-attention. The atom decoder further refines the fused features
through local attention. All Transformer blocks use pair-biased
attention \citep{jumper2021highly} and DiT-style adaptive normalization
\citep{peebles2023scalable}.

\paragraph{Task-specific readouts.}
Dedicated heads map the shared representations to task-specific outputs.
Atom features are used for normalized Cartesian velocity and fractional coordinate
predictions, while token features are used for element identity and sequence type logits. Lattice lengths and angles are predicted from
token features pooled within each system. An optional property module
aggregates token contributions into a domain-normalized residual energy
per atom and predicts normalized force vectors from atom features.
For system $b$ in electronic-structure domain $d$, the physical energy is
reconstructed as

\begin{equation}
\widehat{E}_b =
\sum_{\ell \in b}\varepsilon_{d,z_\ell}
- n_b\left(\mu_d+s_d\widehat{e}_b\right),
\label{eq:energy-reconstruction}
\end{equation}

where $\varepsilon_{d,z}$ are elemental reference energies estimated
from the training split, $\mu_d$ and $s_d$ normalize the residual energy
per atom, and $n_b$ is the atom count. Forces are normalized by a
domain-specific RMS scale during training and restored to
$\mathrm{eV}\,\text{\AA}^{-1}$ at inference. The direct force readout
supports computational efficiency and comparison with prior work. However, note that we enforce neither energy-gradient consistency nor exact rotational equivariance.

\paragraph{Training loss.}
To train Zatom-2, we minimize the weighted multitask objective
\begin{equation}
\begin{aligned}
\mathcal{L} &= \lambda_{\mathrm{FM}}\mathcal{L}_{\mathrm{FM}}
+ \lambda_{\mathrm{lDDT}}\mathcal{L}_{\mathrm{lDDT}} \\
&\quad + g_{1\,\text{\AA}}(t)
  \sum_{k\in\mathcal{K}}
  \lambda_k\mathcal{L}_k \\
&\quad + m_E\lambda_E\mathcal{L}_E
+ m_F\lambda_F\mathcal{L}_F
+ \mathbf{1}_{\mathrm{MLIP}}\lambda_{\mathrm{eq}}\mathcal{L}_{\mathrm{eq}},
\end{aligned}
\label{eq:zatom2-objective}
\end{equation}
where $\mathcal{K}=\{\mathrm{element},\mathrm{sequence},\mathrm{fractional},\mathrm{lattice}\}$ indexes the time-gated training tasks, and $g_\delta(t)=\mathbf{1}[\sigma_{\mathrm{data}}(1-t)/t<\delta]$ gates tasks for near-endpoint supervision. $\mathcal{L}_{\mathrm{FM}}$ is the flow matching loss,
$\mathcal{L}_{\mathrm{lDDT}}$ is the smoothed lDDT loss \citep{mariani2013lddt, butcher2025rfdiffusion3}, and
$\mathcal{L}_E,\mathcal{L}_F$ are Huber losses. The energy/force gates $(m_E,m_F)$ equal $(1,1)$ for clean MLIP inputs, $(g_{1\,\text{\AA}},g_{0.25\,\text{\AA}})$ for force-unconditioned generation examples, and $(0,0)$ otherwise, preventing target-force leakage in force-conditioned sample generation. Both gates vanish unless input examples are uncropped and, for OMol25, have valid charge and spin annotations. Inspired by \citep{elhag2025learning}, the MLIP-only latent equivariance loss $\mathcal{L}_{\mathrm{eq}}$ trains a two-layer network to predict one rotated input view's pooled representation from another, conditioned on the two views' relative rotation. Empirically, we find that $\mathcal{L}_{\mathrm{eq}}$ enables accurate atomic force prediction during joint generative-predictive pretraining.

\subsection{Force conditioning}
\label{sec:force-conditioning}

OMol25 and OMat24 contain substantial structural diversity, spanning high-force, rattled configurations to fully relaxed geometries (Appendix \ref{app:datasets}). We hypothesize that explicitly informing the model of a system’s force state can help it distinguish between geometrically distinct configurations that share the same discrete composition and implicitly inferred topology. To provide this information, we condition both generation and structure prediction tasks on the mean force norm of the system, using it as a global descriptor of the degree of structural relaxation:
\begingroup
\small
\begin{equation}
  s_F = \log_{10}\!\left(
  \frac{1}{L}\sum_{\ell=1}^{L}
  \lVert \mathbf{F}_{\ell}\rVert_2\right).
  \label{eq:force-condition}
\end{equation}
\endgroup

\section{Experiments}
\label{sec:experiments}

We use the OMol25 4M subset (3,986,754 examples) \citep{levine2025omol25} and the OMat24 1M subset (1,009,850 examples) \citep{barrosoluque2024omat24} for pretraining. For details on the datasets, including their composition and properties, we refer the reader to Appendix \ref{app:datasets} and their original publications \citep{levine2025omol25, barrosoluque2024omat24}.

The Experiments section is organized as follows. In Section~\ref{sec:results-pretraining}, we present multiple pretraining configurations for OMol25 and OMat24, combining generation, structure prediction, and force and energy prediction tasks. To show the empirical effects of multitask and multi-domain training, we evaluate each configuration for generation and force conditioning quality. In Section~\ref{sec:results-standard-benchmarks}, we perform a comparison study of Zatom-2 against established benchmarks, to offer a relative evaluation of the model's architecture. In Section~\ref{sec:results-scaling}, we conduct a scaling study to investigate the effects of model and data size on performance. Lastly, in Section~\ref{sec:results-protein-transfer}, we explore the transferability of Zatom-2 to protein generation tasks, demonstrating its potential for broader applications in biomolecular design.

\paragraph{Metrics.} We introduce force-consistency metrics to probe whether our training recipe enables Zatom-2 to capture the degree of rattling/relaxation of each generated sample, and, more broadly, the configurational diversity of OMol25 and OMat24. For each generated sample, we compute atomic forces using UMA-S-1p2~\citep{wood2025uma} and quantify agreement between the resulting mean force norm and the conditioning value using AUROC and MAE (Appendix~\ref{app:evaluation-metrics}). We argue that forces provide a more relevant measure for the degree of rattling across OMol25 and OMat24 systems of varying sizes than unnormalized total energies.  The rest of the metrics used in this section primarily assess generation quality, and are explained in Appendix~\ref{app:evaluation-metrics}.

\subsection{Atomistic pretraining on OMol25 \& OMat24}
\label{sec:results-pretraining}

\begin{table}[htbp]
  \centering
  \scriptsize
  \newcommand{\tablecheckmark}{%
    \textcolor{green!50!black}{\scalebox{1.2}{$\checkmark$}}%
  }
  \renewcommand{\arraystretch}{0.85}
  \caption{\textbf{Pretraining configurations and generation quality.}
  Panel (a) shows multiple training configurations for OMol25 and OMat24 mixed training. Panel (b) reports their performance, where we also compare to 10,000 dataset samples and Zatom-1. All models are trained for 80 epochs. Forces are calculated using UMA~\citep{wood2025uma} and metrics are reported in eV\,\AA$^{-1}$. AUROC uses a conditioning mean force norm cutoff of 1 eV\,\AA$^{-1}$. Structure prediction results are reported in Appendix
  Table~\ref{tab:pretraining-structure-prediction}.}
  \label{tab:pretraining-quality}
  \vspace{0.4em}

  {\textbf{a) Pretraining configurations}\par}
  \vspace{0.25em}
  \begin{tabular}{@{}lccccccc@{}}
    \toprule
    & & \multicolumn{2}{c}{Datasets} & \multicolumn{3}{c}{Tasks} & \\
    \cmidrule(lr){3-4}\cmidrule(lr){5-7}
    Model & Size & OMol25 & OMat24 & Generation & Structure prediction &
    \shortstack{Force \& energy\\prediction} & \shortstack{Force\\conditioning} \\
    \midrule
    Zatom-1 (released) & Large & \tablecheckmark & -- & \tablecheckmark & -- & -- & -- \\
    \midrule
    Zatom-2 (I) & Small & \tablecheckmark & -- & \tablecheckmark & -- & -- & -- \\
    Zatom-2 (II) & Small & \tablecheckmark & -- & \tablecheckmark & -- & -- & \tablecheckmark \\
    Zatom-2 (III) & Small & \tablecheckmark & -- & \tablecheckmark & \tablecheckmark & -- & \tablecheckmark \\
    \midrule
    Zatom-2 (IV) & Base & \tablecheckmark & \tablecheckmark & \tablecheckmark & \tablecheckmark & -- & \tablecheckmark \\
    Zatom-2 (V) & Base & \tablecheckmark & \tablecheckmark & \tablecheckmark & \tablecheckmark & \tablecheckmark & \tablecheckmark \\
    \midrule
    Zatom-2 (VI)--\textit{final} & Large & \tablecheckmark & \tablecheckmark & \tablecheckmark & \tablecheckmark & \tablecheckmark & \tablecheckmark \\
    \bottomrule
  \end{tabular}

  \vspace{0.3em}

  {\textbf{b) Generation quality and conditioning adherence}\par}
  \vspace{0.25em}
  \resizebox{\linewidth}{!}{%
  \begin{tabular}{@{}lrrrrrrrrr@{}}
    \toprule
    & \multicolumn{6}{c}{Sample quality} &
    \multicolumn{3}{c}{\shortstack{Force quality and\\conditioning adherence}} \\
    \cmidrule(lr){2-7}\cmidrule(l){8-10}
    Model & \shortstack{RDKit\\Valid. $\uparrow$} & Unique $\uparrow$ & Diversity $\uparrow$ &
    FP novelty $\uparrow$ & AFD $\downarrow$ & \shortstack{OMat24\\Valid. $\uparrow$} &
    \shortstack{$\overline{\|F\|}$ $\downarrow$} &
    \shortstack{$\overline{\|F\|}$\\MAE $\downarrow$} & \shortstack{$\overline{\|F\|}$\\AUROC $\uparrow$} \\
    \midrule
    Dataset & 0.846 & -- & 0.946 & -- & 0.0025 & 0.791 & 1.311 & -- & -- \\
    \midrule
    Zatom-1 (released) & 0.941 & 0.614 & 0.940 & 0.218 & 0.047 & -- &
    29.547 & -- & -- \\
    \midrule
    Zatom-2 (I) & 0.840 & 0.923 & 0.937 & 0.349 & 0.016 & -- &
    1.154 & 0.827 & 0.586 \\
    Zatom-2 (II) & 0.823 & 0.898 & 0.940 & 0.354 & 0.017 & -- &
    1.252 & 0.791 & 0.662 \\
    Zatom-2 (III) & 0.846 & 0.885 & 0.939 & 0.368 & 0.019 & -- &
    1.627 & 0.722 & 0.730 \\
    \midrule
    Zatom-2 (IV) & 0.836 & 0.884 & 0.938 & 0.354 & 0.018 & 0.727 &
    1.595 & 0.740 & 0.713 \\
    Zatom-2 (V) & 0.851 & 0.899 & 0.940 & 0.365 & 0.019 & 0.748 &
    1.599 & 0.688 & 0.757 \\
    \midrule
    Zatom-2 (VI)--\textit{final} & 0.850 & 0.895 & 0.941 & 0.373 & 0.016 & 0.748 &
    1.269 & 0.616 & 0.817 \\
    \bottomrule
  \end{tabular}%
  }
\end{table}

We investigate a series of training configurations on OMol25, progressively augmenting the generation objective with structure prediction, OMat24 data, and force and energy supervision. We first train a small variant of Zatom-2 on OMol25 alone (70M parameters), with and without force conditioning (see Table~\ref{tab:pretraining-quality}). Compared to Zatom-1, Zatom-2 improves generation quality, especially as seen in lower AFD scores and mean force norms ($\overline{\|F\|}$) predicted by UMA~\citep{wood2025uma}, reflecting better training set distribution learning. Our method for force conditioning improves alignment to the dataset's average $\overline{\|F\|}$, and enables the model to generate force-consistent structures.

We then introduce structure prediction as an additional task. For OMol25, structure prediction conditions on atom types and molecular connectivity and predicts the corresponding geometry. Because connectivity could not be inferred reliably for all OMol25 structures, we restrict this task to a subset of 1,581,810 examples for which bonds could be assigned reliably (Appendix~\ref{app:pretraining-structure-prediction}). For OMat24, structure prediction is performed on the full dataset by conditioning on atom types and predicting both the atomic geometry and lattice parameters.

Adding OMol25 structure prediction (Zatom-2 (II) $\rightarrow$ Zatom-2 (III)) largely preserves generation quality, with a slight reduction in adherence to the dataset's average $\overline{\|F\|}$. We report results for the structure prediction task below, and in Appendix~\ref{app:additional-results}. Incorporating OMat24 supervision and increasing the training budget (Zatom-2 (III) $\rightarrow$ Zatom-2 (IV), 140M parameters) partially restores this force consistency. Finally, we add force and energy prediction as auxiliary tasks and increase the model's capacity (Zatom-2 (IV) $\rightarrow$ Zatom-2 (V) $\rightarrow$ Zatom-2 (VI), 270M parameters), yielding our final configuration and the strongest overall performance across the evaluated metrics.

\paragraph{Force conditioning consistency.} We evaluate how closely Zatom-2 generated samples adhere to their specified force conditioning first by measuring the absolute deviation between the mean force norm of each generated structure and its conditioning target using mean absolute error (MAE, see Appendix~\ref{app:force-sc-metrics}). Second, we assess whether the model can controllably generate structures in low- versus high-force regimes of a given molecular system using AUROC. For the AUROC evaluation, we define dataset-specific force thresholds based on the corresponding OMol25 mean force norm distributions:
$\tau=1$~eV\,\AA$^{-1}$ for \texttt{geom\_orca6}, and
$\tau=2.5$~eV\,\AA$^{-1}$ for
\texttt{ani2x} (see Appendix Figure~\ref{fig:omol25-force-norms}).
Beyond evaluating whether Zatom-2 distinguishes between rattled and relaxed configurations, we use force consistency as the primary metric for structure prediction on OMol25. 

\begin{figure}[htbp]
  \centering
  \includegraphics[height=0.3\linewidth]{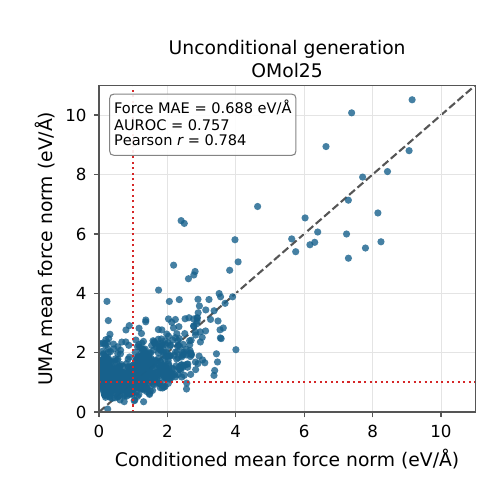}\hfill%
  \includegraphics[height=0.3\linewidth]{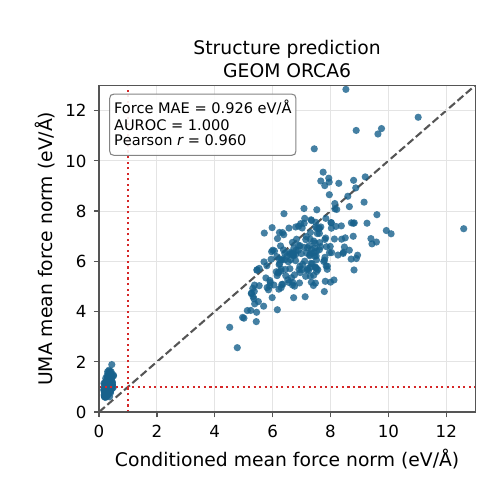}\hfill%
  \includegraphics[height=0.3\linewidth]{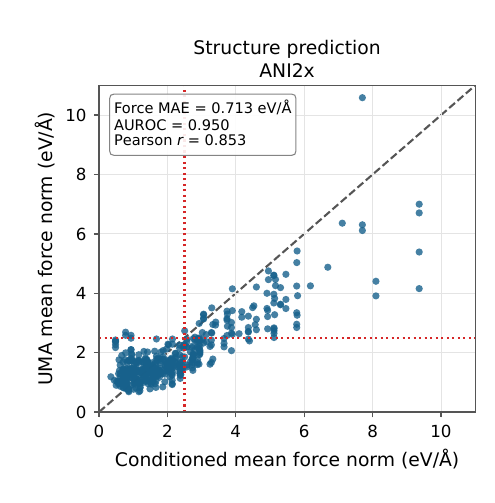}
  \caption{\textbf{Force-conditioning consistency.} Panel (a) compares
  the conditioned mean force norm with the mean force norm evaluated by UMA for
  unconditional OMol25 molecule generation. Panels (b) and (c) give the corresponding
  comparison for OMol25 structure prediction on the GEOM ORCA6 and ANI2x
  subsets, respectively. Red dotted lines mark the AUROC cutoff
  (1\,eV\,\AA$^{-1}$ in (a)--(b), and 2.5\,eV\,\AA$^{-1}$ in (c)).}
  \label{fig:force-conditioning-self-consistency}
\end{figure}

Small molecule structure prediction, more commonly known as conformer generation, is typically assessed using precision and recall based on the root mean square deviation (RMSD) between sets of generated and reference conformers for a given molecular graph~\citep{liu2026goflow}. This evaluation is not directly applicable to OMol25: the dataset contains both equilibrium and non-equilibrium configurations and does not provide a canonical mapping from a molecular graph to a reference conformer ensemble (in fact, the dataset does not provide any molecular graphs or bond topologies). Moreover, conditioned on a molecular graph and target force magnitude, there is generally no unique ground-truth geometry, making RMSD to an individual reference structure an inappropriate measure of prediction quality. Force consistency instead evaluates whether a predicted geometry represents a physically compatible configuration under a specified force regime. 
Figure~\ref{fig:force-conditioning-self-consistency} shows that generated structures show an MAE ranging from 0.69 to 0.93~eV\AA$^{-1}$, and the model reliably distinguishes between low- and high-force regimes, achieving AUROC values between 0.76 and 1.00 across the three evaluated settings. Further results for structure prediction are presented in Appendix Table~\ref{tab:pretraining-structure-prediction}. This indicates that force conditioning provides reliable separation between high- and low-force regions, while precise force matching within each regime remains more challenging.

\paragraph{Energy and force prediction.} By way of pretraining, Zatom-2 uniquely supports MLIP energy and force prediction for molecules and materials. We treat energy and force prediction as an auxiliary objective, motivated by the hypothesis that supervision on forces and energies encourages the model to better learn local geometric structure and improve representation quality, which previous methods have shown leads to improved generative performance~\citep{yan2026bettergeometricrepresentationsmolecule}. We report performance on MLIP prediction and tune generative / structure prediction / force \& energy prediction task mixtures in Appendix~\ref{sec:results-mlip-finetuning}. As expected from a non-specialized model, MLIP accuracy remains below specialized potentials, however the task improves generative quality and especially enhances transfer learning, which is studied in Section~\ref{sec:results-protein-transfer}. We further observe that an MLIP-pretrained model shows better separation of atomic embeddings in PCA space (see Appendix~\ref{app:visualization-embeddings}) compared to a model which is not trained with force and energy supervision.

\subsection{Generation on GEOM-Drugs and MP20}
\label{sec:results-standard-benchmarks}

\begin{table}[htbp]
  \centering
  \scriptsize
  \caption{\textbf{Generation benchmarks on GEOM-Drugs and MP20.}
  Panel (a) reports unconditional GEOM-Drugs molecule generation metrics,
  and panel (b) reports MetaSUN yield for unrelaxed MP20 material generation.}
  \label{tab:geom-drugs-mp20-benchmark}
  \vspace{0.4em}

  \begin{minipage}[t]{0.6\linewidth}
  \centering
  \vspace{0pt}
  \scriptsize
  \setlength{\tabcolsep}{2pt}
  {\textbf{(a) GEOM-Drugs molecule generation}\par}
  \vspace{0.25em}

  \begin{tabular*}{\linewidth}{@{\extracolsep{\fill}}lrrrrr@{}}
    \toprule
    Model & Valid $\uparrow$ & Unique $\uparrow$ &
    \shortstack{PB-valid$\uparrow$} &
    \shortstack{Valid \&\\unique $\uparrow$} &
    \shortstack{Valid \&\\PB-valid $\uparrow$} \\
    \midrule
    EQGAT-diff & 94.6 & 100.0 & 59.7 & 94.60 & 56.48 \\
    SemlaFlow & 93.9 & 100.0 & 87.5 & 93.90 & 82.16 \\
    ADiT & 95.3 & 100.0 & 85.3 & 95.30 & 81.29 \\
    TABASCO & 97.6 & 99.03 & 91.6 & 96.65 & 89.40 \\
    Zatom-1 & 93.6 & 99.93 & 94.1 & 93.53 & 88.08 \\
    \shortstack[l]{GEOM-Drugs\\(reference)}
      & -- & -- & 94.0 & -- & -- \\
    \midrule
    Zatom-2 & 96.90 & 99.61 & 91.47 & 96.52 & 88.63 \\
    \bottomrule
  \end{tabular*}
  \end{minipage}\hfill%
  \begin{minipage}[t]{0.38\linewidth}
  \centering
  \vspace{0pt}
  \scriptsize
  \setlength{\tabcolsep}{2pt}
  {\textbf{(b) MP20 material generation (unrelaxed)}\par}
  \vspace{0.25em}

  \begin{tabular*}{\linewidth}{@{\extracolsep{\fill}}lr@{}}
    \toprule
    Model & \shortstack{MetaSUN $\uparrow$} \\
    \midrule
    ADiT (joint QM9+MP20) & 1.0 \\
    Crystal-GFN & 0.0 \\
    Crystalformer & 3.1 \\
    LLaMat2-CIF & 2.1 \\
    SymmCD & 2.4 \\
    \shortstack[l]{Zatom-1 (MP20-only, 80M)} & 0.20 \\
    \shortstack[l]{Zatom-1 (joint QM9+MP20, 80M)} & 0.36 \\
    \shortstack[l]{Zatom-1-L (joint QM9+MP20, 160M)} & 0.64 \\
    \midrule
    Zatom-2 (joint QM9+MP20) & 4.92 \\
    \bottomrule
  \end{tabular*}
\end{minipage}
\end{table}

To contextualize Zatom-2's performance and enable comparison with prior architectures, we evaluate the model on three established generation benchmarks: joint QM9 molecule generation and MP20 material generation, and GEOM-Drugs molecule generation. As reported in Table~\ref{tab:geom-drugs-mp20-benchmark} and Appendix Table~\ref{tab:qm9-generation}, Zatom-2 achieves 95.72\% validity on QM9, 96.90\% validity on GEOM-Drugs, and a 4.92\% yield of unrelaxed metastable, unique, and novel MP20 materials (MetaSUN) on LeMat-GenBench \citep{betala2025lematgenbench}, improving or approaching state-of-the-art performance across these benchmarks.
These results provide empirical support for our formulation in
Sections~\ref{sec:zatom2-flow} and ~\ref{sec:zatom2-backbone}, that atom identities and lattice parameters can be inferred from coordinate representations without separate
generative trajectories over these modalities. We further evaluate Zatom-2 for structure prediction on GEOM-Drugs and MP20, and report comparisons to baselines in Appendix~\ref{app:geom-drugs-conformer-benchmark} and ~\ref{app:mp20-material-structure-prediction}.
Appendix~\ref{app:visualization} provides MolStar visualizations of generated
molecules and materials.

\subsection{Scaling model capacity and data coverage}
\label{sec:results-scaling}

\begin{figure}[htbp]
  \centering
  \includegraphics[width=\linewidth]{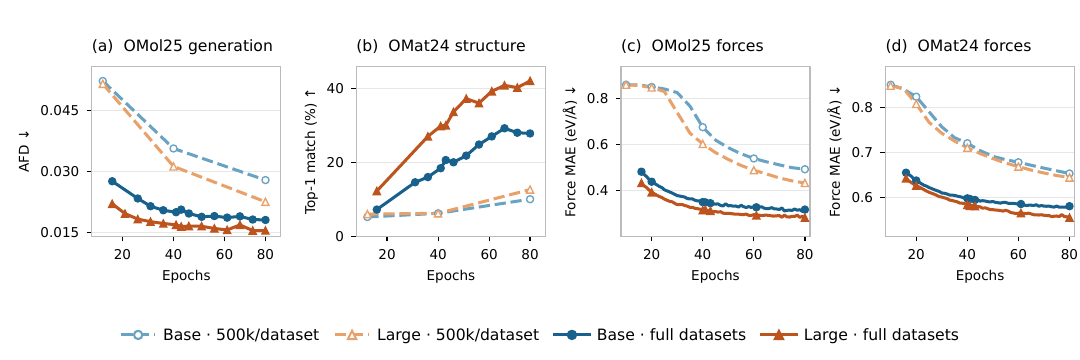}
  \vspace{-12pt}
  \caption{\textbf{Model and data scaling across training epochs.} We evaluate Zatom-2 models trained either on 500k examples per dataset or on the full OMol25 and OMat24 datasets. We report (a) OMol25 molecular generation quality measured by AFD, (b) OMat24 structure prediction measured by top-1 match rate, and force prediction MAE on (c) OMol25 and (d) OMat24. Training on the full dataset with more model parameters consistently improves validation performance across tasks.}
  \label{fig:model-data-scaling}
\end{figure}
\FloatBarrier

We further investigate the effects of scaling Zatom-2 along its data and model parameter axes. Figure~\ref{fig:model-data-scaling} shows how validation sample quality and force prediction performance evolve across training. At an equal number of epochs, across all four validation metrics, training on the full dataset consistently outperforms restricting each dataset domain to 500k examples. Parameter scaling provides an additional improvement under both data regimes, with the largest effect on OMat24 structure prediction. Similar, though smaller, gains are observed for molecular generation and force prediction fidelity. Overall, these results indicate that data coverage and model capacity provide complementary gains, and motivate further scaling to the full OMol25 (100M) and OMat24 (100M) datasets.

\subsection{Transfer to low-data protein generation}
\label{sec:results-protein-transfer}
Finally, we study whether Zatom-2's atomistic pretraining transfers to a low-data protein generation setting. We curate a diverse subset of PDB structures from SCOPe (Appendix~\ref{app:datasets-scope}), which we refer to as SCOPe-2k, and compare training from scratch against finetuning from three pretrained initializations: Zatom-2 (IV) (generation + structure prediction), Zatom-2 (V) (generation + structure prediction + force \& energy prediction), and a variant of Zatom-2 (IV) without force conditioning. We also train RFdiffusion3~\citep{butcher2025rfdiffusion3} on the SCOPe-2k dataset. For each method, we select its checkpoint with the highest designability and novelty in a 96-backbone sweep (Appendix~\ref{app:scope2k-transfer-convergence}).

\begin{table}[htbp]
  \centering
  \scriptsize
  \caption{\textbf{SCOPe-2k protein generation and transfer.} \textbf{(a):} We evaluate 1,027 protein backbones for each configuration, 13 corresponding to each length between 50 and 128 (with means $\pm$ standard deviations over 3 random seeds). \textbf{(b):} The same protocol is repeated for 129--256 residue lengths. FC denotes force-conditioning, and F/E pred. refers to a model trained with force and energy prediction as auxiliary tasks (Zatom (V)). The Zatom (V)-finetuned model substantially outperforms all configurations on the length extrapolation task. Metrics are explained in Appendix~\ref{app:metrics-protein}.}
  \label{tab:scope2k-transfer}
  \setlength{\tabcolsep}{4pt}
  \resizebox{\linewidth}{!}{%
  \begin{tabular}{@{}lrrrr@{}}
    \toprule
    Model / Zatom-2 initialization
      & \shortstack{Designability\\(\%) $\uparrow$}
      & \shortstack{Per-sequence\\success (\%) $\uparrow$}
      & \shortstack{Diversity\\(\#clusters) $\uparrow$}
      & \shortstack{Novelty (\%) $\uparrow$} \\
    \midrule
    \multicolumn{5}{@{}l}{\textbf{(a) In-distribution: 50--128 residues}} \\
    \midrule
    RFdiffusion3 & $\mathbf{91.07 \pm 0.63}$ & $\mathbf{67.61 \pm 1.18}$ & $435.0 \pm 11.5$ & $13.33 \pm 1.25$ \\
    \midrule
    None & $88.22 \pm 0.26$ & $61.35 \pm 0.14$ & $400.3 \pm 5.0$ & $11.41 \pm 0.77$ \\
    Zatom-2 (IV): Generation + structure (w/o FC) & $86.30 \pm 1.12$ & $59.99 \pm 1.61$ & $432.0 \pm 6.1$ & $16.11 \pm 1.32$ \\
    Zatom-2 (IV): Generation + structure & $89.68 \pm 0.84$ & $62.40 \pm 2.08$ & $439.0 \pm 5.0$ & $14.26 \pm 0.71$ \\
    Zatom-2 (V): Generation + structure + F/E pred. & $89.09 \pm 1.61$ & $63.39 \pm 1.98$ & $\mathbf{453.7 \pm 9.0}$ & $\mathbf{16.77 \pm 1.04}$ \\
    \midrule
    \multicolumn{5}{@{}l}{\textbf{(b) Length extrapolation: 129--256 residues}} \\
    \midrule
    RFdiffusion3 & $54.04 \pm 1.81$ & $28.37 \pm 1.19$ & $455.3 \pm 8.3$ & $90.44 \pm 1.11$ \\
    \midrule
    None & $67.81 \pm 1.33$ & $34.68 \pm 0.20$ & $470.7 \pm 17.6$ & $88.81 \pm 1.66$ \\
    Zatom-2 (IV): Generation + structure (w/o FC) & $62.96 \pm 1.18$ & $33.91 \pm 0.67$ & $498.3 \pm 7.4$ & $94.63 \pm 1.59$ \\
    Zatom-2 (IV): Generation + structure & $68.00 \pm 2.22$ & $37.69 \pm 0.86$ & $539.3 \pm 20.4$ & $93.89 \pm 0.90$ \\
    Zatom-2 (V): Generation + structure + F/E pred. & $\mathbf{74.80 \pm 1.06}$ & $\mathbf{45.79 \pm 0.82}$ & $\mathbf{618.7 \pm 3.2}$ & $\mathbf{95.17 \pm 0.25}$ \\
    \bottomrule
  \end{tabular}%
  }
\end{table}

We evaluate protein generation in-distribution, using sequence lengths represented in the training set (50--128 residues), as well as out-of-distribution (OOD), using sequences up to twice as long (129--256 residues). In Table~\ref{tab:scope2k-transfer}, joint generative-predictive pretraining---combining generation, structure prediction, and force \& energy prediction---has the strongest performance across most metrics when compared to other initializations for Zatom-2. In the in-distribution setting, it yields 454 distinct designable clusters on average, compared with 400 when training from scratch, with generation + structure pretraining offering similar performance improvements.

The benefits of pretraining are more pronounced in the OOD regime: Zatom-2 (V) substantially outperforms the other Zatom-2 initializations, improving designability from 67.81\% to 74.80\% compared to random initialization.
Interestingly, pretraining is only beneficial to designability when force conditioning is enabled, supporting our hypothesis in Section~\ref{sec:force-conditioning} that force-aware pretraining is particularly beneficial when learning from data with high configurational diversity. Moreover, initializing from a vanilla generative model on OMol25 and OMat24 (Generation + structure (w/o FC)) even achieves worse performance than a model trained from scratch. All Zatom-2 configurations outperform RFdiffusion3 in the OOD setup, and in turn RFdiffusion3 outperforms Zatom-2's sample designability rates in-distribution.

Together, these results indicate that broad atomistic pretraining can transfer to data-limited generation tasks in biomolecular design. Interestingly, the strongest gains of pretraining arise when predictive objectives for forces and energies are incorporated alongside generation, providing evidence that joint generative-predictive training is a promising recipe for future atomistic generative models. In Appendix~\ref{app:broader-impacts-and-limitations}, we provide a detailed discussion of this work's broader impacts and limitations.

\subsection{Latent space analysis of unseen proteins}
\label{sec:results-protein-embeddings}
\begin{figure}[t]
  \centering
  \includegraphics[width=\linewidth]{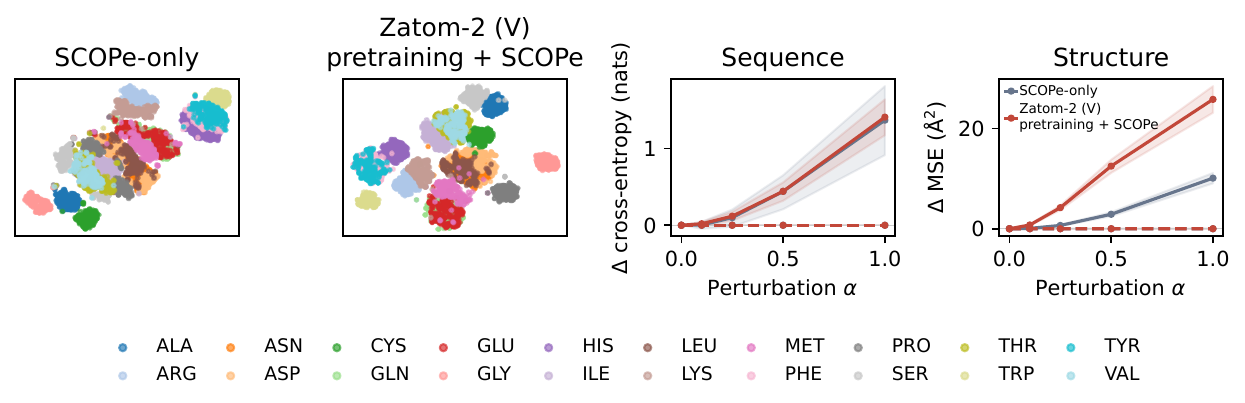}
  \caption{\textbf{Protein latent space organization and localized decoding responses.} Left two: t-SNE views of final token features, colored by amino acid. Right two: increases in sequence and structure reconstruction losses after final-token perturbation; solid/dashed lines denote the target/other amino acids. Bands show 95\% intervals from 2,000 protein bootstrap resamples. Pretrained Zatom-2 is notably more sensitive to local amino acid-specific perturbations than without pretraining.}
  \label{fig:protein-embeddings}
\end{figure}

We examine Zatom-2's amino acid representations of proteins excluded from the SCOPe-2k finetuning dataset through t-SNE visualizations \citep{van2008visualizing,laproteina}. Since Zatom-2's \texttt{atom14} input geometry already encodes amino acid identity, its final token features naturally cluster by amino acid type (Figure~\ref{fig:protein-embeddings}). However, Zatom-2 (V), with joint generative-predictive pretraining, produces clearer separation between amino acid clusters while retaining structurally (e.g., ASN/ASP, GLN/GLU) and chemically (e.g., PHE/TRP/TYR) similar groupings, compared to Zatom-2 trained only on SCOPe. Quantitatively, with such pretraining, Zatom-2's original latent space achieves a higher cross-protein amino acid 10-nearest-neighbor purity score (88.1\% versus 84.8\% for SCOPe-only training) and a higher Calinski-Harabasz index (374.9 versus 228.1), indicating better overall token organization in the model's latent space \citep{calinski1974dendrite}.

\FloatBarrier

To further investigate Zatom-2's latent space, we perform an analysis similar to~\citet{laproteina} and test whether Zatom-2 exhibits desirable amino acid locality after perturbing a \emph{single} amino acid's final token representation $Q_I[j]$ before atom decoding. Namely, 25\%, 50\%, and 75\% along each protein sequence, we add to $Q_I[j]$ the perturbation $\alpha\lVert Q_I[j]\rVert_2\mathbf{u}$ using uniformly random unit directions $\mathbf{u}$ and $\alpha\in\{0,0.1,0.25,0.5,1\}$. Here, models receive the same noise and random directions and hold all other features fixed. Interestingly, even though Zatom-2's decoder for each atom can attend to different tokens' atoms, token-specific perturbation effects are strongly localized in Zatom-2's decoder, with joint generative-predictive pretraining inducing notably more perturbation sensitivity. This suggests that Zatom-2 does not undesirably propagate per-token corruptions amongst atoms.

\section{Conclusion}

In this work, we introduced Zatom-2, an atomistic generative model across molecules and materials, pretrained on the OMol25 (4M) and OMat24 (1M) datasets. We proposed a training recipe for generative models incorporating predictive objectives, and empirically found that force and energy supervision offers better separation of atomic embeddings in PCA space, boosts generative performance, and enables stronger transfer to protein generation tasks. We discuss the importance of force-guided generation on datasets with high configurational diversity, and propose a force consistency metric to assess the ability of the model to distinguish between rattled and relaxed configurations. Zatom-2 performs strongly across existing molecule and material generation benchmarks and achieves strong performance for in-distribution and out-of-distribution protein generation tasks through its generative-predictive pretraining recipe. The architecture also exhibits performance advantages with data and model scaling, motivating larger-scale pretraining efforts in future work.

\subsection*{Reproducibility Statement}

To ensure that this work is reproducible, we have accompanied it with freely available source code at \url{https://github.com/Zatom-AI/nucleus} containing all documentation, data loading infrastructure, and training/inference materials necessary to reproduce the results presented in this manuscript. Additionally, Section~\ref{sec:zatom2} of the main text introduces the Zatom-2 architecture, and Appendix~\ref{app:additional-model-details} provides additional model details, including an in-depth examination of Zatom-2's forward pass, its hyperparameters, and its required computing resources. Lastly, Appendix~\ref{app:datasets} outlines the datasets referenced in this work and how they were sourced/used, as well as what licenses apply to them.

\subsection*{AI Use Statement}

In this work, we used generative AI tools to refine hypotheses, design or provide feedback on research methodology or experiments, implement methods and support qualitative data analysis. We have not used generative AI tools to generate synthetic datasets, help develop theoretical models or conceptual frameworks, or clean and reformat datasets. Formulating mathematical claims, providing critical ingredients for proving mathematical claims, assisting in the writing of proofs, and assisting with translation do not apply to this work. As additional context, we used generative AI tools to create or edit figures or images, suggest experimental parameters, create or edit software code, draft parts of the research paper, brainstorm or source/search for information. We have reviewed all AI-assisted work. For instance, we checked LLM-generated research ideas for potential plagiarism through a manual literature survey, and LLM-generated code was verified and tested for correctness by at least 2 authors. We take responsibility for the final content of this work, including text, claims, or artifacts produced with the aid of generative AI.

\section{Acknowledgments}
This research used resources of the National Energy Research Scientific Computing Center, a DOE Office of Science User Facility supported by the Office of Science of the U.S. Department of Energy (DOE) under Contract No. DE-AC02-05CH11231, using the AI4Sci@NERSC award NERSC DDRERCAP0036206 awarded to AM. NBE would also like to acknowledge that this work was supported in part by the U.S. Department of Energy's Genesis Mission and the Office of Science, Office of Advanced Scientific Computing Research's ModCon under Contract No. DE-AC02-05CH11231 at Lawrence Berkeley National Laboratory. Additionally, MC's PhD is funded by the EPSRC Centre of Doctoral Training in Automated Chemical Synthesis Enabled by Digital Molecular Technologies (SynTech CDT).

\clearpage
\bibliography{iclr2027_conference}
\bibliographystyle{iclr2027_conference}

\appendix
\appendixtoc
\clearpage

\section{Related Work}
\label{app:related-work}

\subsection{Generative modeling of 3D molecules}

3D molecule generation methods commonly model atom identities and coordinates together. For example, equivariant diffusion models have done so while imposing Euclidean symmetry in their denoiser networks \citep{hoogeboom2022equivariant}, and subsequent works have examined optimal architectures and loss functions supporting such equivariant molecular diffusion \citep{vignac2023midi,le2024navigating,irwin2025semlaflow}. TABASCO instead demonstrated that a standard Transformer can support high-accuracy molecule generation~\citep{tabasco}, and Zatom-1 demonstrated the same capabilities for molecule-material generation \citep{morehead2026zatom1}. Zatom-2 retains a Transformer backbone without exact rotation equivariance, uses rotational data augmentations, and extends its range of pretraining tasks to structure prediction, force-conditioned sample generation,
and energy/force prediction. The model's QM9 and GEOM-Drugs results establish its competitiveness with well-known molecule generation baselines, while its OMol25 experiments test substantially broader chemical and conformational coverage and generative capabilities \citep{ramakrishnan2014quantum,levine2025omol25}.

\subsection{Generative modeling of periodic materials}

Materials generation additionally requires fractional coordinates and lattice geometry. Towards this end, CDVAE combines latent materials representations with diffusion-based sample generation \citep{xie2022crystal}. Similarly, DiffCSP performs joint equivariant diffusion of material modalities \citep{jiao2023crystal}, while FlowMM formulates materials generation in the framework of Riemannian flow matching \citep{miller2024flowmm}. Uniquely, FlowLLM uses a language model to provide a base distribution for a materials generation flow matching model \citep{sriram2024flowllm}. All-atom Diffusion Transformers and Zatom-1, however, shift toward supporting shared molecule-material generation capabilities, with the latter supporting prediction tasks as well \citep{joshi2025allatom,morehead2026zatom1}. In this spirit, Zatom-2 supports domain-specific fractional coordinate and lattice geometry prediction while retaining a common Transformer trunk. The model's MP20 results demonstrate that its new \texttt{atom1} tokenizer enables competitive performance for materials generation without dedicated generation tracks for materials-specific input modalities, such as lattice parameters.

\subsection{Representation learning and multitask atomistic prediction}

Molecular representation learning has previously used paired 2D/3D views of atomistic data, masked modeling, and coordinate denoising to improve performance for downstream prediction tasks \citep{liu2022graphmvp,zhou2023unimol,zaidi2023pretraining}. Analogously, Zatom-1 introduced a shared generative model for 3D molecules and materials and studied generative pretraining as a route to obtaining transferable atomistic representations \citep{morehead2026zatom1}. Related efforts have coupled 3D molecule generation and property prediction within a single model framework \citep{feng2025unigem}.

Machine learning interatomic potential models (MLIPs)~\citep{Batatia2022mace,neumann2024orb,wood2025uma} are shown to learn rich representations of chemical domains, which are useful for downstream tasks \citep{wedig2025remdi} and converge in representation space as the improve in performance~\citep{edamadaka2025universallyconvergingrepresentationsmatter,li2026platonic}. This poses the question whether they learn universal descriptors of local atomic geometries which can support generative modeling. This question has been explored using representation alignment (REPA)~\citep{yu2025representation} to align the hidden representations of Boltzmann emulators~\citep{pinede2025unifying} and crystal generative models~\citep{zhang2026crystalrepatransferringphysicalpriors} with those of pretrained MLIPs.

In contrast, Zatom-2 places generative flow matching, structure prediction, and energy/force prediction in a single mixture of pretraining tasks. This framing tests whether generative and structural objectives can coexist with an energy/force (MLIP) objective and whether their shared features remain useful during MLIP-heavy finetuning. Notably, Zatom-2 differs from a standalone universal potential such as UMA \citep{wood2025uma}, which has the sole objective of atomistic property prediction.

Overall, Zatom-2 is positioned around a specific unification question: whether one scalable atomistic backbone can benefit from unified generative, structural, and property prediction capabilities across atomistic 3D systems. The experiments in Section~\ref{sec:experiments} and throughout this work attempt to answer this question from a variety of angles.

\section{Datasets}
\label{app:datasets}

\subsection{OMol25 and OMat24}

\begin{figure*}[t]
  \centering
  \includegraphics[width=\textwidth]{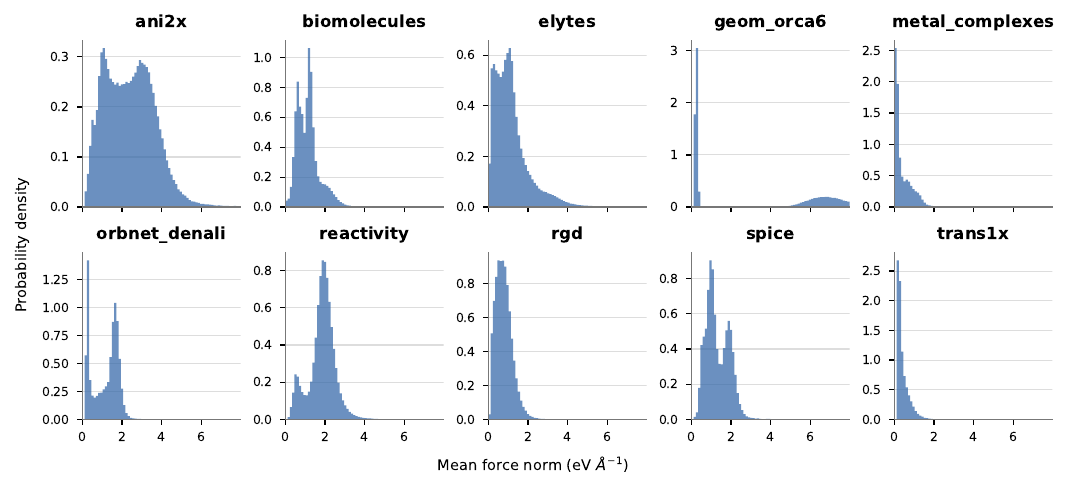}
  \caption{\textbf{Mean force norm distributions across OMol25-4M subsets.}}
  \label{fig:omol25-force-norms}
\end{figure*}

OMol25 is a molecular dataset with more than 100 million single-point density functional theory (DFT) calculations at a high level of theory \citep{levine2025omol25}. OMol25 contains 83 different atom types and combines small molecules, biomolecules, metal complexes, electrolytes, transition state geometries and reactive trajectories. Charge and spin multiplicity vary across the corpus, and a breakdown of the subsets used in the 4M training split (which we adopt in this work), together with their mean force norm distributions, is shown in Figure~\ref{fig:omol25-force-norms}. The values were obtained by computing the mean of the Euclidean norms of the molecules' per-atom force vectors (i.e., the quantity inside the logarithm in Equation~\ref{eq:force-condition} of the main text). 

Other than the composition diversity reflected in Figure~\ref{fig:omol25-force-norms}, OMol25 contains per-system conformational diversity. For example, the GEOM-derived subset of OMol25 captures conformational variation within approximately 300{,}000 molecular families, with around 30\% of each family's available conformers selected for DFT calculations. For its protein pocket--ligand subset, OMol25 performs molecular dynamics (MD) on extracted, cropped binding pockets and their ligands at 300 or 400~K, with restraints on the protein backbone. After filtering and subsampling each trajectory, the first and last retained frames are selected for DFT calculations, providing two configurations of the same pocket--ligand complex with variation in ligand conformation and relative ligand-pocket geometry. Similarly, for metal complexes, five configurations are randomly sampled from each 1--2~ps MLIP-based MD trajectory for DFT calculations. Detailed descriptions of OMol25 composition, rattling, and DFT calculations are presented in \citet{levine2025omol25}.

\begin{figure*}[t]
  \centering
  \includegraphics[width=\textwidth]{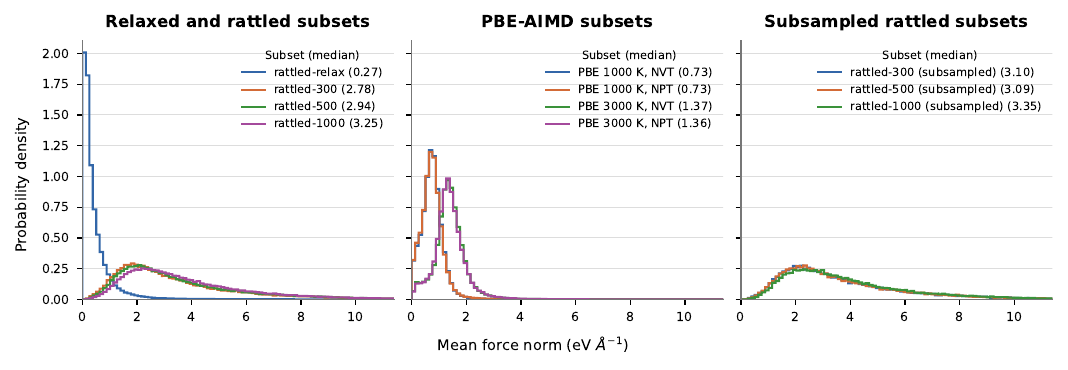}
  \caption{\textbf{Mean force norm distributions across OMat24-1M subsets.}
  Data subsets are grouped by generation protocol (rattled and then re-relaxed, rattled at various temperatures, or ab initio molecular dynamics trajectories; see \citet{barrosoluque2024omat24} for more details). The legend shows subset medians in eV~\AA$^{-1}$.}
  \label{fig:omat24-force-norms}
\end{figure*}

In a complementary manner, OMat24 is a materials dataset containing approximately 118 million DFT calculations with energy and force labels \citep{barrosoluque2024omat24}. Starting structures are taken from the Alexandria corpus and diversified with rattled Boltzmann samples, rattled relaxation trajectories, and ab initio MD. The dataset therefore includes both near-equilibrium and substantially distorted structures, with 89 different elements throughout. We use its public 1M training split (which contains 88 distinct elements) and, similarly to OMol25, report mean force norm distributions in Figure~\ref{fig:omat24-force-norms}.

\begin{figure}[t]
  \centering
  \includegraphics[width=\linewidth]{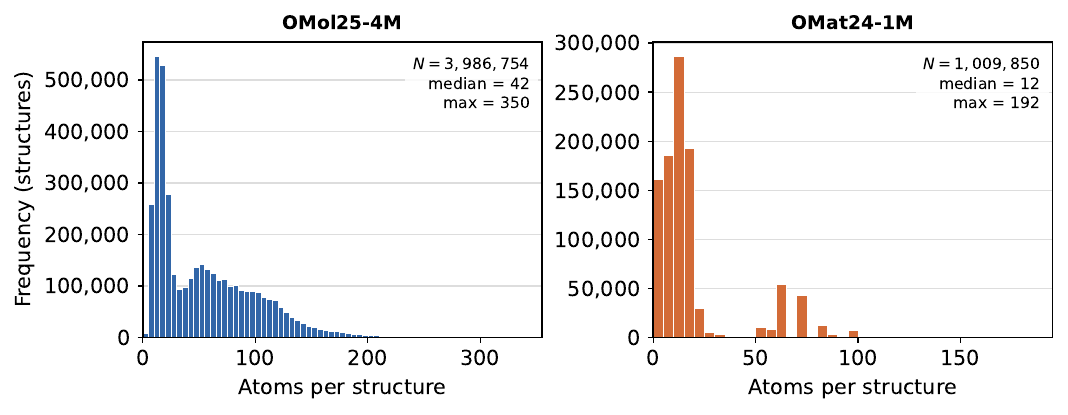}
  \caption{\textbf{Atom count distributions of OMol25-4M \& OMat24-1M.} Hydrogens are included.}
  \label{fig:dataset-atom-counts}
\end{figure}

Figure~\ref{fig:dataset-atom-counts} shows the total atom count distributions of
both datasets. Atom counts include hydrogens.

\FloatBarrier

\FloatBarrier

\subsection{QM9}
\label{app:datasets-qm9}

QM9 contains approximately 134,000 small organic molecules with quantum-chemically optimized geometries and molecular properties \citep{ramakrishnan2014quantum}. Each molecule has at most nine heavy atoms (C, N, O, or F), with hydrogens included. For joint molecule and material generation, we use Zatom-1's corresponding 100,000 QM9 training examples alongside its MP20 training subset described below, matching the corresponding training protocol of Zatom-1 \citep{morehead2026zatom1}.

\subsection{MP20}
\label{app:datasets-mp20}

MP20 consists of 45,231 inorganic material structures from the Materials Project, with at most 20 atoms per unit cell and 89 elements across the dataset \citep{xie2022crystal,morehead2026zatom1}. In this work, we use Zatom-1's corresponding 27,138 MP20 training structures and match its MP20 training conventions \citep{morehead2026zatom1}.

For the structure prediction experiments in Appendix~\ref{app:mp20-material-structure-prediction}, we also prepared a Crystalite partitioning of the dataset \citep{hadziveljkovic2026crystalite}. To do so, we processed Crystalite's published training/validation/testing CSV data splits separately with Niggli reduction enabled and primitive cell reduction and graph construction disabled.

\subsection{GEOM-Drugs}
\label{app:datasets-geom-drugs}

GEOM-Drugs provides energy-annotated conformer ensembles of drug-like molecules,
covering larger and more structurally flexible systems than QM9
\citep{axelrod2022geom}. We use it for unconditional molecule generation and a
separate structure prediction (i.e., conformer generation) task. The latter uses ConfGF's processed training split \citep{shi2021learning} and the 200-molecule test split adopted by GO-Flow \citep{liu2026goflow}, containing 14,324 reference conformers.

\subsection{SCOPe}
\label{app:datasets-scope}

SCOPe (Structural Classification of Proteins---extended)
\citep{chandonia2022scope} takes experimentally determined structures largely
from the PDB, splits proteins into structural domains (up to 256 residues), and
organizes those domains hierarchically according to structural and evolutionary
relationships (e.g., fold, family, and superfamily).
We use the SCOPe ASTRAL-40 subset, which is obtained by clustering SCOPe domains
at 40\% sequence identity and selecting one representative from each cluster,
ensuring that no two domains share more than 40\% sequence identity. Genie
\citep{lin2023genie} showed that using the short variant of SCOPe ASTRAL-40,
containing only 3,936 domains of up to 128 residues after six domains were
deleted because of parsing incompatibilities, is sufficient to train a
well-balanced protein generative model. We call this the SCOPe-4k dataset.

We further sample selectively from the
488 SCOPe-4k fold groups to reduce the dataset size while preserving structural diversity:
\begin{enumerate}
  \item Group domains by their SCOPe fold.
  \item Randomize the order of the folds and the order of domains within each
  fold using a fixed random seed.
  \item Cycle through the folds, taking one domain from each fold per cycle. If
  a fold runs out of domains, skip it.
  \item Use the first 1,000 domains as the 1k dataset, the first 2,000 as the 2k
  dataset, and the first 4,000 as the 4k dataset.
\end{enumerate}

\subsection{Dataset licenses}
\label{app:dataset-licenses}

Table~\ref{tab:zatom2-dataset-licenses} records the licensing terms for the datasets described in this appendix as well as the LeMat-Bulk reference dataset. These terms concern only the data; third-party software and pretrained models used in method evaluations retain their separate licenses.

\begin{table}[htbp]
  \centering
  \small
  \caption{\textbf{Dataset licenses and access terms.} Dataset names link to
  the corresponding release or the provider's usage statement.}
  \label{tab:zatom2-dataset-licenses}
  \vspace{0.4em}
  \begin{tabular}{@{}p{0.30\linewidth}p{0.66\linewidth}@{}}
    \toprule
    Dataset & Source data terms \\
    \midrule
    \href{https://huggingface.co/facebook/OMol25}{OMol25}
      & Creative Commons Attribution 4.0 (CC BY 4.0) \\
    \href{https://huggingface.co/datasets/facebook/OMAT24}{OMat24}
      & CC BY 4.0 \\
    \href{https://doi.org/10.6084/m9.figshare.1057646}{QM9}
      & CC0 1.0 (original Figshare data deposit) \\
    \href{https://huggingface.co/datasets/chaitjo/MP20_ADiT}{MP20}
      & MIT for the processed ADiT release used by Zatom-1;
        underlying Materials Project data are CC BY 4.0 \\
    \href{https://doi.org/10.7910/DVN/JNGTDF}{GEOM-Drugs}
      & CC0 1.0 (Harvard Dataverse deposit) \\
    \href{https://scop.berkeley.edu/about/ver=2.08}{SCOPe / ASTRAL}
      & Provider states that all data are freely available to all users;
        no standard named license is specified \\
    \href{https://www.wwpdb.org/about/usage-policies}{PDB source structures}
      & CC0 1.0; attribution to structure authors is encouraged \\
    \href{https://huggingface.co/datasets/LeMaterial/LeMat-Bulk}{LeMat-Bulk}
      & CC BY 4.0, including attribution to upstream data providers \\
    \bottomrule
  \end{tabular}
\end{table}

\clearpage
\section{Additional Model Details}
\label{app:additional-model-details}

\subsection{Model forward pass}
\label{app:forward-pass-comparison}

Algorithm~\ref{alg:zatom2-forward} details Zatom-2's model forward pass and task readouts. Throughout this subsection, $q,a$ are atom/token representations, $c,s$ denote conditioning, and $p,z$ are pair representations/biases. Subscripts $0$ and $-$ denote initial values and the previous pass, respectively. $D$ computes distograms of representative-token coordinates, $W$ denotes a learned projection, and $\operatorname{sg}$ stops gradients. Enc, Trunk, and Dec are attention stacks; Down/Up are gated cross-attention between resolutions.

The feature dictionary $f$ contains conditioning, atom-to-token and token-to-system maps, and fixed-coordinate masks. $L,I,B$ index atom, token, and system levels, respectively: $\operatorname{MeanPool}_{L\to I}$ averages atoms within tokens, and $\operatorname{MeanPool}_{I\to B}$ averages tokens within systems. $\operatorname{Time}_{L/I}$ combines Fourier time features, RMSNorm, and a linear projection. Each $\operatorname{Head}$ includes its own adaptive normalization and output projection; the energy readout additionally sums domain-specific token contributions and divides by the system's atom count. The velocity, energy, and forces returned are, by construction, normalized; physical velocity is $\sigma_{\rm data}\hat v$, and property rescaling follows Section~\ref{sec:zatom2-backbone}, namely Equation~\ref{eq:energy-reconstruction}.

\begin{algorithm}[htbp]
  \caption{Zatom-2 forward pass with coordinate recycling and multitask readouts.}
  \label{alg:zatom2-forward}
  \small
  \begin{algorithmic}[1]
    \Require Coordinates $x_t\in\mathbb{R}^{L\times3}$, flow time $t$, features $f$,
      total passes $R\geq1$ (default $2$)
    \State $(c_0,s_0,p,z_0)\gets\operatorname{Initializer}(f)$
      \Comment{Static atom/token features and pair biases}
    \State $u\gets x_t/\sigma_{\rm data}$;
      $q_0\gets\operatorname{RMSNorm}(c_0+W_Lu)$
    \State $a_0\gets\operatorname{RMSNorm}(s_0+\operatorname{MeanPool}_{L\to I}(W_Iu))$
    \State $(t_L,t_I)\gets\operatorname{BroadcastTime}(t;f)$;
      set fixed atoms and fully fixed tokens to time $1$
    \State $c\gets(q_0+\operatorname{Time}_L(t_L))/2$;
      $s\gets(a_0+\operatorname{Time}_I(t_I))/2$
    \State $G_L\gets\operatorname{LocalNeighbors}(x_t,f;128)$;
      $M_I\gets\operatorname{SameSystemMask}(f)$
    \For{$r=1,\ldots,R$}
      \State Evaluate this pass without gradients unless training and $r=R$
      \State $q\gets q_0$; $a\gets a_0$; $z\gets z_0$
        \Comment{Restart node states on every pass}
      \If{$r>1$}
        \State $z\gets z_0+W_D D(x_{\rm rec})$
          \Comment{65-bin distogram of representative-token coordinates}
      \EndIf
      \State $q\gets\operatorname{Enc}(q;c,p,G_L)$
        \Comment{Local atom DiT stack}
      \State $a\gets\operatorname{Down}(q,a;f)$
        \Comment{Tokens attend to their constituent atoms}
      \State $a\gets\operatorname{Trunk}(a;s,z,M_I)$
        \Comment{Global token DiT stack within each system}
      \State $q\gets\operatorname{Up}(q,a;f)$
        \Comment{Atoms attend to 24 channel groups of their token}
      \State $q\gets\operatorname{Dec}(q;c,p,G_L)$
        \Comment{Local atom DiT stack}
      \State $\hat v\gets\operatorname{Head}_{\rm velocity}(q,c)$;
        $\hat v[\text{fixed atoms}]\gets0$
      \State $\hat x\gets x_t+(1-t_L)\sigma_{\rm data}\hat v$
        \Comment{Endpoint estimate; fixed coordinates stay unchanged}
      \State $\ell_{\rm element}\gets\operatorname{Head}_{\rm element}(a,s)$;
        $\ell_{\rm sequence}\gets\operatorname{Head}_{\rm sequence}(a,s)$
      \State $\hat x_{\rm frac}\gets\operatorname{Head}_{\rm fractional}(q,c)$
      \State $g\gets\operatorname{Down}_{I\to B}(a,g_0;f)$;
        $\bar s\gets\operatorname{MeanPool}_{I\to B}(s)$
        \Comment{Learned system query $g_0$}
      \State $(\hat l,\hat\alpha)\gets\operatorname{Head}_{\rm lattice}(g,\bar s)$
        \Comment{Lattice lengths and angles}
      \State $\mathcal O\gets\{\hat x,\hat v,\ell_{\rm element},\ell_{\rm sequence},
        \hat x_{\rm frac},\hat l,\hat\alpha\}$
      \If{property heads are enabled}
        \State $\hat e\gets\operatorname{Head}_{\rm energy}(a,s;f)$;
          $\hat F\gets\operatorname{Head}_{\rm force}(q,c)$;
          $\mathcal O\gets\mathcal O\cup\{\hat e,\hat F\}$
      \EndIf
      \State $x_{\rm rec}\gets\operatorname{sg}(\hat x)$
        \Comment{Only coordinates feed the next pass}
    \EndFor
    \State \Return $\mathcal O$
      \Comment{Readouts from the final pass}
  \end{algorithmic}
\end{algorithm}
\FloatBarrier

To complement Algorithm ~\ref{alg:zatom2-forward}, Table~\ref{tab:forward-pass-comparison} compares one model forward pass (including recycling) in Zatom-2, RFdiffusion3's public source code (with standard memory mode enabled) \citep{butcher2025rfdiffusion3}\footnote{RFdiffusion3 source and configuration: \href{https://github.com/RosettaCommons/foundry/tree/b02eed6a6bdf8f44d14a80cc36e3da13c9f2291c/models/rfd3}{\texttt{foundry/models/rfd3}}, revision \texttt{b02eed6a6bdf}.} and Emyx's published pseudocode (Appendix~J, Algorithms~2--17 of \citealp{williams2026emyx}, arXiv v1). The table lists graph construction first to align stages across models, while Algorithm~\ref{alg:zatom2-forward} follows Zatom-2's implementation order. This reordering leaves the model's outputs unchanged.

\begingroup
  \small
  \setlength{\tabcolsep}{3pt}
  \setlength{\LTcapwidth}{\textwidth}
  \renewcommand{\arraystretch}{1.0}
  \begin{longtable}{@{}p{0.035\linewidth}*{3}{>{\raggedright\arraybackslash}p{0.305\linewidth}}@{}}
  \caption{\textbf{Aligned forward pass operations.} Enc, Trunk, and Dec are attention stacks; Down/Up are gated cross-attention between input resolutions. By definition, no recycling feedback is used in the first pass.}
  \label{tab:forward-pass-comparison}\\
    \toprule
    & RFdiffusion3 & Emyx & Zatom-2 \\
    \midrule
    \endfirsthead
    \multicolumn{4}{l}{\tablename~\thetable\ (continued)}\\
    \toprule
    & RFdiffusion3 & Emyx & Zatom-2 \\
    \midrule
    \endhead
    \bottomrule
    \endlastfoot
    1 & Build local atom neighborhoods (128 keys); use global token attention.
      & Build sparse atom/token graphs prioritizing sequence, bonds, ligand, and motif edges, then spatial neighbors.
      & Build local atom neighborhoods (128 keys, token-index/spatial neighbors); use global token attention. \\
    \addlinespace
    2 & Embed atom/token metadata; mix dense pair features with two triangle-free Pairformer blocks; pool atom pairs into token pairs.
      & Bottleneck-embed atom14 atom/token and global features; project relative positions and bonds to edge biases.
      & Bottleneck-embed atom1/atom14 metadata, domain, and charge/spin; residually-embed mean force norm; map bonds, relative positions, and any available fixed/reference geometry to head biases. \\
    \addlinespace
    3 & Add linear embeddings of $x_\sigma/\sqrt{\sigma^2+\sigma_{\rm data}^2}$; condition on Fourier features of $\frac14\log(\sigma/\sigma_{\rm data})$.
      & Add Fourier coordinate embeddings; average geometry-aware conditioning with sinusoidal time features.
      & Add linear atom and mean-pooled linear token embeddings of $x_t/\sigma_{\rm data}$; average each conditioning stream with Fourier features of $t$. \\
    \addlinespace
    4 & Cache $q_e=\mathrm{Enc}(q_0;c,p)$ and $a_e=\mathrm{Down}(q_e,a_0;s)$ \emph{before} recycling.
      & Begin recycling before the atom encoder.
      & Begin recycling before the atom encoder. \\
    \addlinespace
    5 & Reset $q,a\leftarrow q_e,a_e$; concatenate $z_0$, noisy-input and recycled distograms; apply transitions and two triangle-free Pairformer blocks to obtain $s,z$.
      & Reset $q,a$ to initial states plus projected, normalized $\operatorname{sg}(q_-,a_-)$; set $z=z_0+WD(\hat x_-)$ on sparse edges.
      & Reset $q,a\leftarrow q_0,a_0$; set $z=z_0+WD(\hat x_-)$ for all token pairs. No node-state feedback. \\
    \addlinespace
    6 & Reuse cached encoder/downcast outputs.
      & $q\leftarrow\mathrm{Enc}(q;c,p)$; $a\leftarrow\mathrm{Down}(q,a)$.
      & $q\leftarrow\mathrm{Enc}(q;c,p)$; $a\leftarrow\mathrm{Down}(q,a)$. \\
    \addlinespace
    7 & $a\leftarrow\mathrm{Trunk}(a;s,z)$, globally over tokens.
      & $a\leftarrow\mathrm{Trunk}(a;s,z)$, along token edges.
      & $a\leftarrow\mathrm{Trunk}(a;s,z)$, globally within each packed system. \\
    \addlinespace
    8 & For each decoder block: $q\leftarrow\mathrm{Dec}_j(\mathrm{Up}_j(q,a);c,p)$. Downcast detached states for sequence logits.
      & Upcast once using learned token copies; then $q\leftarrow\mathrm{Dec}(q;c,p)$.
      & Upcast once using 24 channel groups per token; then $q\leftarrow\mathrm{Dec}(q;c,p)$. \\
    \addlinespace
    9 & $u=W\mathrm{RMSNorm}(q)$; $\hat x=c_{\rm skip}x_\sigma+c_{\rm out}u$. Predict sequence logits from the final downcast.
      & $v=\mathrm{AdaLNHead}(q,c)$; $\hat x=x_t+(1-t)v$; zero motif velocities.
      & $\hat v=\mathrm{AdaLNHead}(q,c)$; zero fixed-atom velocities; $\hat x=x_t+(1-t_L)\sigma_{\rm data}\hat v$. Read out element/sequence, fractional coordinates, lattice, and optional energy/forces. \\
    \addlinespace
    10 & Recycle coordinates/distograms; repeat lines 5--9 (two passes by default).
       & Recycle detached nodes and endpoint geometry; repeat lines 5--9 ($N_r+1=3$ inference passes).
       & Recycle endpoint geometry only; repeat lines 5--9 (two passes here). \\
  \end{longtable}
\endgroup

\subsection{Hyperparameters}
\label{app:hyperparameters}

Table~\ref{tab:zatom2-hyperparameters} summarizes Zatom-2's architecture and OMol25/OMat24 pretraining settings. Joint pretraining samples the two domains (i.e., OMol25 molecules and OMat24 materials) equally with replacement, using $\sim$5 million draws per configured epoch for 80 epochs; an epoch is therefore a sampling budget rather than a full pass through each source dataset. Generation/structure prediction samples the two training tasks equally. Adding MLIP supervision changes the generation/structure prediction/MLIP task probabilities to 0.34/0.33/0.33. Force conditioning is supplied for 80\% of eligible generation and structure prediction examples and disabled in the non-force-conditioned model ablations.

\begin{table}[htbp]
  \centering
  \small
  \caption{\textbf{Key Zatom-2 hyperparameters.} S/base/L vary the token trunk depth while retaining the same feature widths. Token budgets are per GPU before diffusion batch/view augmentation; a smaller token budget is paired with twice as much gradient accumulation.}
  \label{tab:zatom2-hyperparameters}
  \vspace{0.4em}
  \begin{tabular}{@{}p{0.43\linewidth}p{0.53\linewidth}@{}}
    \toprule
    Hyperparameter & Setting \\
    \midrule
    Approximate model size (S/base/L) & 70M / 140M / 270M parameters \\
    Token-trunk blocks (S/base/L) & 9 / 18 / 36 \\
    Atom encoder / decoder blocks & 3 / 3 \\
    Atom / token / token-pair widths & 128 / 768 / 256 \\
    Atom / token attention heads & 4 / 12 \\
    Attention & Local atom attention (128 keys); global token attention \\
    Transformer dropout / drop-path rate & 0.1 / 0.1 \\
    Denoiser passes & 2 (one coordinate-derived recycling update) \\
    Optimizer & Adam; constant learning rate $10^{-4}$ \\
    Adam $(\beta_1,\beta_2)$ / $\epsilon$ & $(0.9,0.95)$ / $10^{-8}$ \\
    Gradient norm clipping value / EMA decay rate & 10 / 0.999 \\
    Training precision & \texttt{bf16-mixed} (default) / \texttt{32-true} (MLIP finetuning) \\
    Packed token budget per step (S/base; L) & 7,168; 3,584 \\
    Crop limit / augmented views per system & 512 tokens / 2 views \\
    Coordinate prior scale $\sigma_{\mathrm{data}}$ & 16\,\AA \\
    Training-time distribution & $0.98\,\mathrm{Beta}(1.9,1)+0.02\,\mathrm{Uniform}(0,1)$,
      clipped to $[10^{-3},1-10^{-3}]$ \\
    \bottomrule
  \end{tabular}
\end{table}

The Euclidean coordinate loss has a weight of 4, with a smoothed lDDT loss coefficient of 0.25; atom element type and token sequence type cross-entropy losses each have weight 0.1. Joint generative/structure-MLIP pretraining uses energy/force/latent equivariance loss weights of 0.1/0.3/0.1 and a 10,000-step MLIP loss warmup. To balance gradient magnitudes between generative and predictive tasks, fractional coordinates, lattice length, and lattice angle weights are 0.25/0.05/0.05 with MLIP supervision and 10/1/1 otherwise. Notably, our SCOPe transfer learning experiments use a batch size of 4 input systems per GPU, four augmented views per system, and crops of at most 128 tokens and 1,792 atoms.

Our configuration for OMol25/OMat24 generated sample evaluation uses 200 sampling steps with force-only classifier-free guidance at a scale of 2. In contrast, our QM9/MP20 results use 100 steps without classifier-free guidance (n.b., to fairly compare these inference settings to those of Zatom-1) and a 3,584-token training budget per GPU. Both sampling configurations use EMA weights, two denoiser passes, an EDM schedule exponent of 7, a noise scale of 1.003, and a step scale of 1.5. Furthermore, their churn coefficients are 0.6 and 0.8, respectively.

\subsection{Compute resources}
\label{app:compute-resources}

Training and evaluation use a computing cluster with NVIDIA
A100 GPUs containing 80\,GBs of GPU memory. Distributed launchers
configure four GPUs per node, for either four-node (16-GPU) jobs or sixteen-node
(64-GPU) jobs. For pretraining, gradient accumulation preserves
an approximate maximum of $64\times7{,}168=458{,}752$ tokens per optimizer
update before the two augmented diffusion views, including when the per-GPU token budget is halved for Zatom-2-L's experiments.

Table~\ref{tab:zatom2-compute} estimates training costs from training run metadata by multiplying elapsed run time by the number of GPUs used for each experiment.

\begin{table}[htbp]
  \centering
  \small
  \caption{\textbf{Approximate 80GB A100 GPU hours by training configuration.}
  G/S/P denote generation, structure prediction, and MLIP prediction; FC denotes
  force conditioning. Epoch budgets count completed epochs;
  MLIP finetuning runs end at raw epoch 119. All finetuning costs exclude pretraining the corresponding initial model weights.}
  \label{tab:zatom2-compute}
  \vspace{0.4em}
  \begin{tabular}{@{}p{0.50\linewidth}lrr@{}}
    \toprule
    Model training configuration & Budget / endpoint & GPUs & GPU hours \\
    \midrule
    Small, OMol25 G, no FC & 80 epochs & 16 & 1,830 \\
    Small, OMol25 G, FC & 80 epochs & 16 & 1,820 \\
    Small, OMol25 G+S, FC & 80 epochs & 64 & 2,280 \\
    Base, OMol25+OMat24 G+S, FC & 80 epochs & 64 & 2,740 \\
    Base, OMol25+OMat24 G+S+P, FC & 80 epochs & 64 & 2,850 \\
    Large, OMol25+OMat24 G+S+P, FC & 80 epochs & 64 & 4,090 \\
    Base, OMol25+OMat24 G+S, no FC & 80 epochs & 64 & 2,810 \\
    Base, fixed 500k per OMol25/OMat24 dataset, G+S+P & 16 * 5 = 80 epochs & 64 & 540 \\
    Large, fixed 500k per OMol25/OMat24 dataset, G+S+P & 16 * 5 = 80 epochs & 64 & 790 \\
    Base, QM9+MP20 G & 10,000 epochs & 16 & 2,380 \\
    Base, GEOM-Drugs G & 80 epochs & 64 & 970 \\
    \midrule
    Base, MLIP finetuning from G+S & 40 epochs & 64 & 1,520 \\
    Base, MLIP finetuning from G+S+P, Part 1 & 40 epochs & 64 & 1,520 \\
    Base, MLIP finetuning from G+S+P, Part 2 & +120 epochs & 64 & 4,560 \\
    \midrule
    SCOPe, no pretraining & Step 25,000 & 16 & 210 \\
    SCOPe, G+S initialization, no FC & Step 125,000 & 16 & 270 \\
    SCOPe, G+S initialization, FC & Step 120,000 & 16 & 180 \\
    SCOPe, G+S+P initialization, FC & Step 130,000 & 16 & 320 \\
    \bottomrule
  \end{tabular}
\end{table}

\section{Evaluation Metrics}
\label{app:evaluation-metrics}

In this section, we describe the metrics defined in the \texttt{nucleus}
and \texttt{zevals} code implementations underlying Zatom-2. Unless stated
otherwise, lower error or distance is better. 

\subsection{Force self-consistency metrics}
\label{app:force-sc-metrics}

In this subsection, we define how we measure the consistency between a molecule or material sample's predicted forces and those used as conditioning for that sample.

\begin{itemize}
    \item \textbf{Force self-consistency.} For a generated sample, we compute its mean atomic force norm, $N^{-1}\sum_{\ell=1}^{N}\lVert\mathbf F_\ell\rVert_2$, without relaxation. To predict a sample's forces, UMA-S-1p2 uses its OMol and OMat task heads for molecules and materials, respectively \citep{wood2025uma}. We report the mean of this per-structure statistic in eV\,\AA$^{-1}$. When running UMA on 1000 random structures from OMol25 and 1000 structures from OMat24, we get MAEs between calculated mean force norms and those extracted from the metadata of 0.00587 and 0.02462~eV\,\AA$^{-1}$, respectively.
    \item \textbf{UMA target MAE.} This is the mean absolute deviation of UMA-S-1p2's predicted mean force norms and the requested conditioning value over samples successfully scored by UMA. As a reference for this force-norm proxy, we score 1,000 structures sampled uniformly without replacement from each of the OMol25 and OMat24 validation splits (seed 42), using their original geometries without relaxation.
    \item \textbf{UMA AUROC.} This measures the area under the receiver operating characteristic curve (AUROC) for distinguishing low- and high-force conditioning targets, using a dataset's ground-truth mean force norm values with cutoff $\geq\tau$ as the positive class. We use $\tau=1$~eV\,\AA$^{-1}$ in all contexts, except for OMol25's ANI2x data subset, where $\tau=2.5$~eV\,\AA$^{-1}$, based on the subset distributions shown in Figure~\ref{fig:omol25-force-norms} and Figure~\ref{fig:omat24-force-norms}.
\end{itemize}

\subsection{Molecule generation metrics}
\label{app:metrics-molecule-generation}

In this subsection, we define the default metrics used to assess molecule sample quality, which differ slightly from those used in the main text's Section \ref{sec:results-standard-benchmarks} for GEOM-Drugs/QM9 baseline experiments.

\begin{itemize}
  \item \textbf{RDKit validity and uniqueness.} Validity is the fraction of
  generated molecule samples for which RDKit produces a valid SMILES representation of the sample's largest molecular fragment \citep{landrum2013rdkit}. Uniqueness
  is the number of distinct valid SMILES divided by the number of valid
  samples.
  \item \textbf{Internal diversity.} For the set of unique valid SMILES,
  diversity is the mean pairwise Tanimoto distance, $1-s_{ij}$, between
  radius-2, 2,048-bit Morgan fingerprints.
  \item \textbf{Novelty.} To calculate novelty, the experiments in the main text query each valid generated SMILES against a precomputed (OMol25) training set database of fingerprints generated by FPSim2 \citep{eloy_felix_2025_14652272}. In this setting, a method's corresponding novelty value is one minus the mean top-1 Morgan fingerprint Tanimoto similarity to the training set of all its generated molecule samples.
  \item \textbf{PoseBusters pass rate.} PoseBusters applies standardized
  loading, sanitization, connectivity, valence, geometry, clash, flatness, and
  internal energy tests \citep{buttenschoen2024posebusters}. The inputs for each test are RDKit-valid molecules; a method's aggregate "PoseBusters-valid" rate is its fraction of RDKit-valid molecules passing every configured PoseBusters test.
  \item \textbf{Force diagnostics.} UMA force norms and force conditioning
  metrics follow Appendix~\ref{app:force-sc-metrics}, using UMA's OMol head. Additionally, Table~\ref{tab:pretraining-quality} of the main text reports UMA's OMol25 summary metrics.
\end{itemize}

\subsection{Material generation metrics}
\label{app:metrics-material-generation}

In this subsection, we define the default metrics used to assess material sample quality, which differ slightly from those used in the main text's Section \ref{sec:results-standard-benchmarks} for MP20 baseline experiments.

\begin{itemize}
  \item \textbf{Compositional validity.} A material's composition is considered valid if SMACT finds that its oxidation states satisfy charge neutrality and the Pauling electronegativity test; note that single-element systems and all-metal alloys are accepted by the metric \citep{davies2019smact}.
  \item \textbf{Structural and overall validity.} A material's structure is considered valid if it has a periodic cell with volume at least $0.1$\,\AA$^3$ and no periodic interatomic distance below $0.5$\,\AA. Overall validity of a material requires both compositional and structural validity and a successfully constructed CrystalNN fingerprint \citep{pan2021benchmarking}.
  \item \textbf{Uniqueness.} To calculate uniqueness, valid materials are grouped using Pymatgen's \texttt{StructureMatcher} utility with $(\mathrm{ltol},\mathrm{stol},\mathrm{angle\_tol})=(0.3,0.5,10^\circ)$ \citep{ong2013pymatgen}. Uniqueness is the number of groups divided by the number of valid materials.
  \item \textbf{Internal diversity.} Each valid material is represented
  by its normalized mean atom-wise CrystalNN fingerprint. Diversity is then one minus the mean off-diagonal cosine similarity over all successfully fingerprinted generated materials.
  \item \textbf{Novelty.} To calculate novelty, one representative material from each unique valid group is compared with cached OMat24 training fingerprints. A material is considered novel if its cosine similarity to every reference fingerprint is below $0.99$; novelty is reported over only successfully fingerprinted representatives.
  \item \textbf{Force diagnostics.} UMA force norms and force conditioning
  metrics follow Appendix~\ref{app:force-sc-metrics}, using UMA's OMat head.
\end{itemize}

\subsection{QM9 and MP20 sample generation metrics}
\label{app:metrics-qm9-mp20}

The QM9 and MP20 benchmarks of Section~\ref{sec:results-standard-benchmarks} in the main text follow Appendix~B of Zatom-1 \citep{morehead2026zatom1}. Here, we describe only the differences from Appendices~\ref{app:metrics-molecule-generation} and~\ref{app:metrics-material-generation}.

\paragraph{QM9.}
Table~\ref{tab:qm9-generation} reports individual PoseBusters pass rates for
connectivity, bond angles, bond lengths, aromatic ring flatness, double bond
flatness, internal energy, and internal steric clashes. Each constituent PoseBusters test result is the fraction of generated molecules passing that test among RDKit-valid molecules. Molecular validity follows Appendix~\ref{app:metrics-molecule-generation}.

\paragraph{MP20.}
The LeMat-GenBench metrics in
Appendix Table~\ref{tab:geom-drugs-mp20-benchmark-full} differ from the
material metrics in Appendix~\ref{app:metrics-material-generation} as follows
\citep{betala2025lematgenbench}:
\begin{itemize}
  \item \textbf{Overall validity.} A valid material must pass charge balance,
  interatomic distance, coordination, and physical unit tests. These tests assess oxidation states and bond valences, radii-based separation, element-specific coordination, and density, lattice, format, and symmetry constraints.
  \item \textbf{Uniqueness.} The number of structurally distinct valid
  materials is divided by all attempted materials, rather than only valid
  materials.
  \item \textbf{Novelty.} A unique valid material is novel if it has no
  structural match in the (5M) LeMat-Bulk reference dataset. Reported novelty rates divide the number of such materials by all attempted samples; they do
  not use the default OMat24 fingerprint similarity criterion.
  \item \textbf{Metastable, unique, and novel (MetaSUN).} The fraction of
  all attempted materials that are valid, unique, novel, and satisfy
  $0<E_{\mathrm{hull}}\leq0.1$~eV/atom, where $E_{\mathrm{hull}}$ is the mean
  energy above the hull across an MLIP ensemble of Orb, MACE, and UMA. Notably, this metastable test excludes stable materials with $E_{\mathrm{hull}}\leq0$, a portion of which can often be recovered via post hoc MLIP-based relaxation (n.b., which we exclude in this work to assess direct method output quality).
\end{itemize}

\subsection{GEOM-Drugs sample generation metrics}
\label{app:metrics-geom-drugs-generation}

GEOM-Drugs validity, uniqueness, and aggregate PoseBusters validity follow
Appendix~\ref{app:metrics-molecule-generation}; individual PoseBusters pass
rates use the definition in Appendix~\ref{app:metrics-qm9-mp20}, following
Zatom-1 \citep{morehead2026zatom1}. The additional metric definitions needed
for the results in Appendix~\ref{app:complete-generation-benchmarks} are:
\begin{itemize}
  \item \textbf{Valid \& unique and Valid \& PB-valid.} These joint (intersection) metrics divide the number of distinct valid SMILES and the number of RDKit-valid molecules passing all PoseBusters checks, respectively, by all attempted samples. They therefore include the effect of RDKit validity, whereas uniqueness and PoseBusters validity are conditional on RDKit validity.
\end{itemize}

\subsection{Atomistic Fr\'echet Distance}
\label{app:metrics-afd}

\textbf{Background.} The Atomistic Fr\'echet Distance (AFD) is the Fr\'echet Inception Distance (FID) applied to 3D atomistic structures \cite{perez2026atomisticgenevals}. FID scores an image generator by the Fr\'echet distance between Gaussian fits to embeddings of real and generated images created with a pretrained model (Inception-v3) \cite{FID-paper}. AFD replaces the Inception network with a self-supervised pretrained embedding model for atomistic data and applies the same distance to sets of molecules or materials. AFD scores provide a label-free measure of distributional similarity between two sets of atomistic structures - a ground truth reference set and a generated set. Unlike other binary evaluation tests like molecule 'validity' or PoseBusters geometry checks, AFD scores are sensitive to both compositional and conformational diversity and better represent a generative model's ability to produce realistic and diverse atomistic structures. Lower AFD scores suggest generated data is less distinguishable from real data (lower is better). However, AFD does not certify that any single structure is valid and should be reported in combination with validity checks.
\begin{equation}
\mathrm{AFD}^2 \;=\; \lVert \mu_r - \mu_g \rVert_2^{\,2}
\;+\; \operatorname{Tr}\!\Big( \Sigma_r + \Sigma_g - 2\big(\Sigma_r \Sigma_g\big)^{1/2} \Big)
\label{eq:afd}
\end{equation}
Here $(\mu_r,\Sigma_r)$ and $(\mu_g,\Sigma_g)$ are the mean and covariance of the reference and generated embeddings, and Equation~\ref{eq:afd} is the Fr\'echet distance between the two Gaussians. Embeddings are the 256-dimensional graph-level output of a frozen Conditional Equivariant Transformer (CT) \cite{perez2026scd}. The encoder defines the feature space and therefore what the score can detect. Notably, the CT embedding is $E(3)$-invariant and responds monotonically to coordinate noise and to atom mutation or removal. It is also insensitive to chirality and unit-cell multiplicity.

\textbf{Choice of encoder.} The encoder checkpoint chosen for AFD must match one's evaluation domain (e.g., molecules, materials) because cross-domain checkpoints are far less sensitive. AFD is not an absolute score and requires a ground-truth reference set to define its target distribution. Additionally, the Fr\'echet estimator is biased upward at finite $N$, so scores are comparable only at equal $N$ with the same reference set and checkpoint. In this work, we use the pretrained CT checkpoint \href{https://huggingface.co/Ty-Perez/ct-scd-omol25}{\texttt{ct-scd-omol25}}  (pretrained on OMol25-4M) to evaluate generated samples from the OMol25 distribution and the checkpoint \href{https://huggingface.co/Ty-Perez/ct-scd-amp20}{\texttt{ct-scd-amp20} } (pretrained on Alex-MP-20) to evaluate models trained on OMat24 (see Table~\ref{tab:afd_baselines}).

\begin{table}[H]
\centering
\small
\caption{AFD baselines for encoder dataset pairs used in this work (finite-$N$ floor): Each entry is the mean $\pm$ standard deviation of the AFD score between two non-overlapping ground-truth reference sets of size $N$, over 5 replicates (no structure is shared within or across replicates at a given $N$).}

\label{tab:afd_baselines}
\begin{tabular}{r c c}
\toprule
$N$ & OMol25 \texttt{4m} / \texttt{ct-scd-omol25} & OMat24 \texttt{1m} / \texttt{ct-scd-amp20} \\
\midrule
100     & $0.02321 \pm 0.00021$   & $0.0776 \pm 0.0042$    \\
500     & $0.005461 \pm 0.000078$ & $0.01691 \pm 0.00091$  \\
1{,}000 & $0.002527 \pm 0.000032$ & $0.008268 \pm 0.00034$ \\
5{,}000 & $0.0000916 \pm 0.0000047$ & $0.001236 \pm 0.000027$ \\
\bottomrule
\end{tabular}
\end{table}

\subsection{Structure prediction metrics}
\label{app:metrics-structure-prediction}

\textbf{Molecules.} For the OMol25 structure prediction evaluation in Table~\ref{tab:pretraining-structure-prediction}, each target has one reference structure and one generated candidate. We additionally report symmetry-aware best heavy-atom RMSD after rigid alignment, removing hydrogens and requiring the generated and reference molecular graphs to agree. Note that force consistency is the primary metric discussed in Section~\ref{sec:results-pretraining}. We report the mean and median molecular RMSD in \AA. In this work, molecule structure prediction evaluations use 500 held-out OMol25 test set targets unless otherwise specified. The GEOM-Drugs structure prediction evaluation instead follows Appendix~\ref{app:geom-drugs-conformer-benchmark}.

\textbf{Materials.} For structure prediction of periodic materials, we use Pymatgen's \texttt{StructureMatcher} utility with $(\mathrm{ltol},\mathrm{stol},\mathrm{angle\_tol})=(0.3,0.5,10^\circ)$ and volume rescaling enabled \citep{ong2013pymatgen}. Here, a match rate is the fraction of all prediction targets that match their predicted structures within the predefined geometric tolerances above. Normalized root mean square (RMS) displacement divides predicted-reference coordinate deviations by the characteristic length per atom, $(V/N)^{1/3}$, and is summarized only over matched targets; note that it is not an Angstrom RMSD per se. Failed conversions and matcher errors count as unmatched predictions.

\subsection{Interatomic potential metrics}
\label{app:metrics-mlip}

\textbf{Energies and forces.} For systems $b=1,\ldots,B$ with $n_b$ labeled atoms and ground-truth energy $E_b$ and forces $F_b$, we report the MLIP metrics
\begin{align}
  \mathrm{MAE}_E
  &= \frac{1}{B}\sum_b |\widehat E_b-E_b|, &
  \mathrm{MAE}_{E/\text{atom}}
  &= \frac{1}{B}\sum_b
     \frac{|\widehat E_b-E_b|}{n_b}, \\
  \mathrm{MAE}_F
  &= \frac{1}{3\sum_b n_b}
     \sum_{b,\ell,k}
     |\widehat F_{b\ell k}-F_{b\ell k}|.
  \label{eq:mlip-metrics}
\end{align}
Total energy and energy-per-atom errors are in eV and eV/atom, respectively; force component errors are in eV\,\AA$^{-1}$. Energy coverage is the fraction of systems with finite predictions and valid targets after applying elemental reference masks; force coverage is defined analogously over atoms with finite vector labels.

For reference, MLIP task results using these metrics are reported in Appendix~\ref{sec:results-mlip-finetuning}.

\subsection{Protein generation metrics}
\label{app:metrics-protein}

The metrics used for Table~\ref{tab:scope2k-transfer} are defined below. For
each generated backbone, we use
SolubleMPNN~\citep{dauparas2022proteinmpnn} to design eight sequences and
ESMFold2~\citep{candido2026esmfold2} to refold each sequence. We apply no
pLDDT confidence gate.

\begin{itemize}
  \item \textbf{Self-consistency RMSD (scRMSD).} Each refold is aligned to its generated backbone by a Kabsch superposition of corresponding C$\alpha$ atoms. Best-of-eight scRMSD is the minimum value for each backbone; means and medians are then taken across backbones.
  \item \textbf{Designability.} A protein backbone is considered designable
  when at least one of its eight refolds has C$\alpha$ scRMSD below
  $1.5$\,\AA{} (the best-of-eight criterion).
  \item \textbf{Per-sequence success rate.} For backbone $i$, let $m_i$
  be the number of its eight SolubleMPNN refolds with scRMSD below 1.5\,\AA.
  The reported rate is $N^{-1}\sum_{i=1}^{N}m_i/8$, the expected success of
  choosing one designed sequence uniformly per backbone.
  \item \textbf{Distinct designable clusters.} These are computed by restricting the symmetric TM-align matrix of all generated backbones to the best-of-eight designable subset, applying clustering at TM$\geq 0.5$, and counting the
  resulting clusters.
  \item \textbf{Novelty.} We select the best-of-eight designable backbones and compare
  every retained backbone against the canonical SCOPe-2k training set with
  TM-align \citep{zhang2005tmalign}. For each query, we take the maximum TM-score
  normalized by query length. Novelty is defined as the number of backbones with score below
  0.5, divided by the number of designable backbones.
  \item \textbf{Joint sequence--structure clusters.} For the checkpoint-sweep
  diagnostic in Appendix~\ref{app:scope2k-transfer-convergence}, we again use
  the best-of-eight designable subset. A pair is similar only when its
  min-symmetrized TM-align score is at least 0.5 and its generated sequences
  have at least 10\% identity with 70\% bidirectional coverage. We apply
  complete-linkage clustering to this joint relation and count the resulting
  clusters, following the sequence-and-structure diversity motivation of
  Proteina and RFD3 \citep{geffner2025proteina,butcher2025rfdiffusion3}.
\end{itemize}

\section{Additional Results}
\label{app:additional-results}

\subsection{OMol25 and OMat24 structure prediction}
\label{app:pretraining-structure-prediction}

To enable bond conditioning for the structure prediction task on OMol25, we infer bonds using OpenBabel \citep{oboyle2011openbabel}. Because OMol25 contains high-energy structures, reactive compounds and biomolecules, for which OpenBabel cannot correctly infer bonds, we filter out all compounds which contain disconnected fragments after bond inference, as a proxy for failure of inferring the correct bonds. This excludes 60.32\% of OMol25 training examples, and we train on the remaining subset for structure prediction. 

Table~\ref{tab:pretraining-structure-prediction} reports molecule and material structure prediction results for the pretrained configurations in Table~\ref{tab:pretraining-quality}. Each target has one generated candidate and evaluation metrics are defined in Appendix~\ref{app:metrics-structure-prediction}.

\begin{table}[htbp]
  \centering
  \scriptsize
  \renewcommand{\arraystretch}{0.85}
  \caption{\textbf{Molecule/material top-1 structure prediction after
  pretraining.} All models use the 80-epoch pretrained checkpoints described
  in Table~\ref{tab:pretraining-quality}. OMol25 RMSDs are reported in \AA\ and forces are calculated with UMA.
  Compared to OMol25, OMat24 structure prediction is more sensitive to outliers, so we report median metrics for OMat24. 
  OMol25 AUROC uses a target mean force norm cutoff of 1~eV\,\AA$^{-1}$.}
  \label{tab:pretraining-structure-prediction}
  \vspace{0.4em}
  \begin{tabular*}{\linewidth}{@{\extracolsep{\fill}}lrrrrr@{}}
    \toprule
    & \multicolumn{5}{c}{OMol25} \\
    \cmidrule(l){2-6}
    Model & Mean RMSD $\downarrow$ & Median RMSD $\downarrow$ &
    \shortstack{Mean\\$\overline{\|F\|}$ $\downarrow$} &
    \shortstack{Target\\MAE $\downarrow$} & AUROC $\uparrow$ \\
    \midrule
    Zatom-2 (III) & 1.739 & 1.714 & 2.271 & 0.725 & 0.798 \\
    Zatom-2 (IV) & 1.708 & 1.675 & 2.138 & 0.787 & 0.787 \\
    Zatom-2 (V) & 1.762 & 1.729 & 1.953 & 0.637 & 0.801 \\
    Zatom-2 (VI) & 1.723 & 1.743 & 1.211 & 1.214 & 0.818 \\
    \bottomrule
  \end{tabular*}
  \par\vspace{0.5em}
  \begin{tabular*}{\linewidth}{@{\extracolsep{\fill}}lrrrr@{}}
    \toprule
    & \multicolumn{4}{c}{OMat24} \\
    \cmidrule(l){2-5}
    Model & Match $\uparrow$ & \shortstack{Median norm.\\RMS disp. $\downarrow$} &
    \shortstack{Median\\$\overline{\|F\|}$ $\downarrow$} &
    \shortstack{Target\\Median AE $\downarrow$} \\
    \midrule
    Zatom-2 (IV) & 0.446 & 0.225 & 2.765 & 1.703 \\
    Zatom-2 (V) & 0.294 & 0.349 & 32.458 & 30.553 \\
    Zatom-2 (VI) & 0.460 & 0.321 & 18.867 & 17.380 \\
    \bottomrule
  \end{tabular*}
\end{table}
\FloatBarrier

\subsection{MLIP performance before and after finetuning}
\label{sec:results-mlip-finetuning}

Zatom-2 predicts energies and per-atom forces for both molecules and periodic materials with the same model used for generation and structure prediction. Table~\ref{tab:mlip-finetuning} compares pretrained and MLIP-finetuned versions of Zatom-2 for energy and force prediction. Here, evaluation uses 2.76 million OMol25 and 107,732 OMat24 validation examples, with fixed atom identities and geometries and no force conditioning. The finetuned Zatom-2 base models start from generation/structure pretraining (Zatom-2 (IV)) or joint generation/structure/MLIP pretraining (Zatom-2 (V)); Zatom-2 (VI) provides a reference point for how a larger pretrained model fairs for MLIP prediction.

\paragraph{Finetuning improves prediction, especially for molecules.} After 40 epochs of finetuning, Zatom-2 (V)'s force MAE falls from 0.253 to 0.130\,eV\,\AA$^{-1}$ on OMol25 (49\%) and from 0.637 to 0.587\,eV\,\AA$^{-1}$ on OMat24 (8\%), with lower aggregate energy errors in both domains. Extended finetuning further reduces force MAE to 0.097 and 0.564\,eV\,\AA$^{-1}$, respectively. Starting from joint MLIP pretraining also yields lower aggregate errors than starting from generation/structure pretraining alone, suggesting that generative pretraining can still produce a rich model latent space but nonetheless one that is not as informative of atomistic properties as a latent space produced via MLIP supervision.

\paragraph{Generative capabilities are largely retained.}
Keeping generation and structure prediction examples in Zatom-2's finetuning data mixture preserves the model's capabilities for these tasks while improving MLIP accuracy. For Zatom-2 (V), OMol25 generation AFD improves from 0.019 to 0.016 and structure prediction RMSD changes from 1.762 to 1.729\,\AA; OMat24 generation validity and structure match rate remain comparable to their pretrained values. Nonetheless, retention is metric-dependent: for example, molecular structure prediction's force conditioning MAE worsens with extended finetuning even as UMA mean force norms and MLIP errors continue to decrease. Moreover, materials generation and structure prediction metrics degrade with MLIP finetuning, however, interestingly, the performance is in part recovered with more finetuning epochs (Zatom-2 (V) + FT (extended)).

\paragraph{Accuracy and physical consistency remain limiting.} Zatom-2's energy and force errors remain substantially above specialized MLIP baselines such as eSEN, AllScAIP, and UMA \citep{morehead2026zatom1,qu2026allscaip,wood2025uma}. These comparisons provide context rather than a controlled ranking, since training budgets and evaluation protocols partially differ between these methods (including our 1M OMat24 training subset versus AllScAIP's 100M OMat24 training set size). Zatom-2's improvements are also uneven across subsets of OMol25 and OMat24; for instance, biomolecular total-energy error increases slightly after initial finetuning, and rattled materials remain difficult. Lastly, Zatom-2's direct force prediction head enforces neither consistency between predicted energies and their gradients nor exact rotational equivariance with respect to the model's input geometries. These results demonstrate that Zatom-2 supports multitask energy and force prediction, but they currently do not establish Zatom-2 as a suitable method for high-accuracy energy-conserving molecular dynamics or reliable physical simulations.

\begin{table}[!t]
  \centering
  \scriptsize
  \renewcommand{\arraystretch}{0.95}
  \caption{\textbf{MLIP accuracy and retention of generative capabilities.} ``+ FT'' denotes 40 epochs of finetuning with 80\% of training steps allocated to MLIP prediction and 10\% for generation and structure prediction each; here, ``extended'' means adding 120 additional MLIP-heavy finetuning epochs. Model names follow Table~\ref{tab:pretraining-quality}. Panel (a) reports generation and top-1 structure prediction results after finetuning, with forces calculated using UMA; corresponding pretrained results appear in Tables~\ref{tab:pretraining-quality} and~\ref{tab:pretraining-structure-prediction}. For OMat24, we observe that generation and structure prediction are sensitive to outliers and we report median values. Panels (b)--(c) report validation MAEs (Appendix~\ref{app:metrics-mlip}); external baselines are published values with partially differing training and evaluation protocols.}
  \label{tab:mlip-finetuning}
  \vspace{0.4em}

  {\textbf{(a) Generation and top-1 structure prediction retention}\par}
  \vspace{0.25em}
  \resizebox{\linewidth}{!}{%
  \begin{tabular}{@{}lrrrrrrrrr@{}}
    \toprule
    & \multicolumn{9}{c}{OMol25} \\
    \cmidrule(l){2-10}
    & \multicolumn{4}{c}{Generation} &
    \multicolumn{5}{c}{Structure prediction} \\
    \cmidrule(lr){2-5}\cmidrule(l){6-10}
    Model & AFD $\downarrow$ &
    \shortstack{Mean\\$\overline{\|F\|}$ $\downarrow$} &
    \shortstack{Target\\MAE $\downarrow$} &
    \shortstack{AUROC\\$\uparrow$} &
    \shortstack{Mean\\RMSD $\downarrow$} &
    \shortstack{Median\\RMSD $\downarrow$} &
    \shortstack{Mean\\$\overline{\|F\|}$ $\downarrow$} &
    \shortstack{Target\\MAE $\downarrow$} &
    \shortstack{AUROC\\$\uparrow$} \\
    \midrule
    Zatom-2 (IV) + FT & 0.017 & 1.205 & 0.540 & 0.807 & 1.771 & 1.747 & 1.570 & 0.632 & 0.890 \\
    Zatom-2 (V) + FT & 0.016 & 1.137 & 0.530 & 0.809 & 1.729 & 1.706 & 1.458 & 0.690 & 0.828 \\
    \midrule
    Zatom-2 (V) + FT (extended) & 0.015 & 1.040 & 0.526 & 0.840 & 1.720 & 1.713 & 1.178 & 0.871 & 0.882 \\
  \end{tabular}%
  }
  \par\vspace{0.4em}
  \resizebox{\linewidth}{!}{%
  \begin{tabular}{@{}lrrrrrrrr@{}}
    \toprule
    & \multicolumn{8}{c}{OMat24} \\
    \cmidrule(l){2-9}
    & \multicolumn{4}{c}{Generation} &
    \multicolumn{4}{c}{Structure prediction} \\
    \cmidrule(lr){2-5}\cmidrule(l){6-9}
    Model & Valid $\uparrow$ & AFD $\downarrow$ &
    \shortstack{Median\\$\overline{\|F\|}$ $\downarrow$} &
    \shortstack{Target\\Median AE $\downarrow$} &
    Match $\uparrow$ & \shortstack{Median norm.\\RMS disp. $\downarrow$} &
    \shortstack{Median\\$\overline{\|F\|}$ $\downarrow$} &
    \shortstack{Target\\Median AE $\downarrow$} \\
    \midrule
    Zatom-2 (IV) + FT & 0.812 & 1.114 & 11.167 & 9.651 & 0.406 & 0.269 & 10.834 & 8.790 \\
    Zatom-2 (V) + FT & 0.759 & 1.182 & 38.826 & 36.866 & 0.308 & 0.354 & 32.128 & 30.098 \\
    \midrule
    Zatom-2 (V) + FT (extended) & 0.758 & 1.177 & 25.783 & 24.169 & 0.382 & 0.310 & 20.843 & 19.019 \\
    \bottomrule
  \end{tabular}%
  }
  \par\vspace{0.25em}
  \begin{minipage}{0.98\linewidth}
  \end{minipage}

  \vspace{0.4em}
  {\textbf{(b) OMol25 MLIP prediction}\par}
  \vspace{0.25em}
  \begin{tabular*}{\linewidth}{@{\extracolsep{\fill}}lrrrr@{}}
    \toprule
    Model & $E$ MAE (eV) $\downarrow$ &
    $E$/atom MAE (eV/atom) $\downarrow$ &
    Force MAE (eV/\AA) $\downarrow$ & $E$ coverage $\uparrow$ \\
    \midrule
    Zatom-1 (joint) & 2.512780 & 0.034450 & 0.062990 & -- \\
    eSEN-md-d. (4M) & 0.077130 & 0.001320 & 0.006780 & -- \\
    AllScAIP-md-ft-cons. (4M) & 0.044400 & -- & 0.007510 & -- \\
    \midrule
    Zatom-2 (V) & 1.366503 & 0.016426 & 0.252603 & 100\% \\
    Zatom-2 (VI) & 1.253679 & 0.015344 & 0.228265 & 100\% \\
    \midrule
    Zatom-2 (IV) + FT & 1.709557 & 0.018713 & 0.164550 & 100\% \\
    Zatom-2 (V) + FT & 1.178403 & 0.013088 & 0.130012 & 100\% \\
    \midrule
    Zatom-2 (V) + FT (extended) & 1.015206 & 0.011174 & 0.096865 & 100\% \\
    \bottomrule
  \end{tabular*}

  \vspace{0.4em}
  {\textbf{(c) OMat24 MLIP prediction}\par}
  \vspace{0.25em}
  \begin{tabular*}{\linewidth}{@{\extracolsep{\fill}}lrrrr@{}}
    \toprule
    Model & $E$ MAE (eV) $\downarrow$ &
    $E$/atom MAE (eV/atom) $\downarrow$ &
    Force MAE (eV/\AA) $\downarrow$ & $E$ coverage $\uparrow$ \\
    \midrule
    UMA-L & -- & 0.009700 & 0.043500 & -- \\
    AllScAIP-md-ft-cons. & -- & 0.010700 & 0.054300 & -- \\
    \midrule
    Zatom-2 (V) & 1.634724 & 0.100531 & 0.636802 & 99.9991\% \\
    Zatom-2 (VI) & 1.637466 & 0.098387 & 0.617528 & 99.9991\% \\
    \midrule
    Zatom-2 (IV) + FT & 1.618015 & 0.095861 & 0.604039 & 99.9991\% \\
    Zatom-2 (V) + FT & 1.395262 & 0.089066 & 0.586901 & 99.9991\% \\
    \midrule
    Zatom-2 (V) + FT (extended) & 1.338368 & 0.087149 & 0.563980 & 99.9991\% \\
    \bottomrule
  \end{tabular*}
\end{table}
\FloatBarrier

\subsection{SCOPe-2k checkpoint trajectories}
\label{app:scope2k-transfer-convergence}

\begin{figure}[!htbp]
  \centering
  \includegraphics[width=\linewidth]{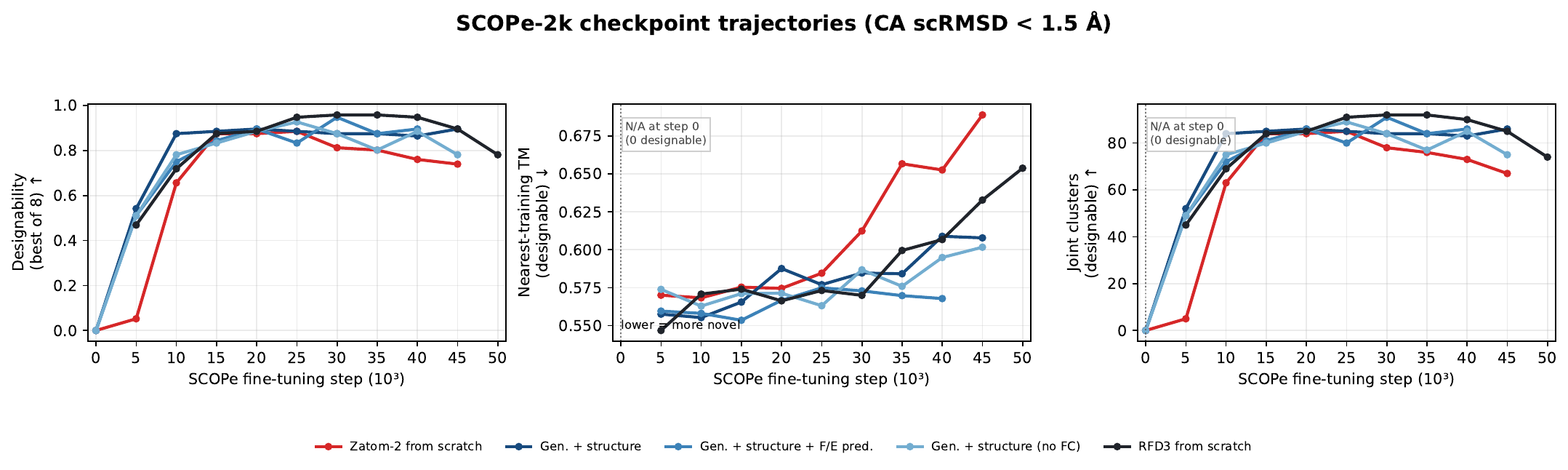}
  \caption{\textbf{SCOPe-2k convergence} From left to right, the panels show
  best-of-eight designability (CA scRMSD
  $<1.5$\,\AA{}), the mean maximum exact, query-normalized
  TM-align score against the canonical SCOPe-2k training set among designable
  backbones (lower is more novel), and distinct joint sequence--structure
  clusters among designable backbones. Joint clusters require TM $\geq 0.5$
  together with at least 10\% sequence identity and 70\% coverage. Each point
  uses 96 generated backbones and eight refolds per backbone.}
  \label{fig:scope2k-transfer-convergence}
\end{figure}

Because we train our protein generation models on a small dataset which is prone to overfitting, we evaluate intermediary model checkpoints during training and finetuning to select the checkpoint with the best designability / novelty tradeoff. We perform the study for all four initializations of Zatom-2 in Table~\ref{tab:scope2k-transfer} and for RFdiffusion3~\citep{butcher2025rfdiffusion3}. Each checkpoint
was evaluated using 96 unconditional backbones, eight SolubleMPNN sequences per
backbone, and ESMFold2 refolding. The pretrained checkpoints begin at raw step
100k, so Figure~\ref{fig:scope2k-transfer-convergence} reports their local
finetuning step after subtracting 100k, and scratch-trained Zatom-2 and RFD3 start at 0 steps. Metric definitions are given in
Appendix~\ref{app:metrics-protein}.

Based on convergence trends, we selected the following checkpoints for evaluation: Zatom-2 from scratch at \texttt{step-00025000}; Zatom-2 (IV) without force conditioning at \texttt{step-00125000} (25k local finetuning steps); Zatom-2 (IV) with force conditioning at \texttt{step-00120000} (20k local steps); Zatom-2 (V) at \texttt{step-00130000} (30k local steps); and RFdiffusion3 at \texttt{step-00030000}.

The trajectories show faster convergence for pretrained models. The
generation-and-structure-pretrained model exceeds 85\% designability at 10k
local steps, whereas Zatom-2 trained from scratch and RFdiffusion3 first exceed 85\% designability 15k. The two models trained from scratch also deteriorate more
sharply after they peak. The pretrained models are generally
preserving high designability and novelty over the same late-training region. We interpret this
pattern of models trained from scratch as evidence consistent with overfitting, whereas the pretrained models show more resilience against overfitting. 

\FloatBarrier

\subsection{Comparison to generative baselines}

\subsubsection{Generation on QM9, GEOM-Drugs, and MP20}
\label{app:complete-generation-benchmarks}

Table~\ref{tab:qm9-generation} reports results on QM9 generation for a model trained jointly on QM9 and MP20, and Table~\ref{tab:geom-drugs-mp20-benchmark-full} contains full generation results for MP20. In Table~\ref{tab:geom-drugs-posebusters-baselines} we report individual
PoseBusters checks for GEOM-Drugs generation.

\begin{table}[htbp]
  \centering
  \scriptsize
  \caption{\textbf{QM9 generation results.}
  Zatom-2 is trained jointly on QM9 and MP20 (Appendix~\ref{app:datasets}). Published baselines are from
  Table~2 of \citet{morehead2026zatom1} (arXiv v4). }
  \label{tab:qm9-generation}
  \vspace{0.4em}

  \begin{minipage}[t]{0.44\linewidth}
  \centering
  \vspace{0pt}
  \setlength{\tabcolsep}{2pt}
  {\textbf{(a) Validity (\%)}\par}
  \vspace{0.25em}
  \begin{tabular}{@{}lr@{}}
    \toprule
    Model & Validity $\uparrow$
    \\
    \midrule
    Equivariant Diffusion & 91.90
    \\
    Symphony & 83.50
    \\
    GeoLDM & 93.80
    \\
    ADiT (QM9-only) & 92.19
    \\
    ADiT (joint QM9+MP20) & 94.45
    \\
    Zatom-1 (QM9-only, 80M) & 92.88
    \\
    Zatom-1 (joint QM9+MP20, 80M) & 94.94
    \\
    Zatom-1-L (joint QM9+MP20, 160M) & 95.26
    \\
    \midrule
    Zatom-2 (joint QM9+MP20) & 95.72
    \\
    \bottomrule
  \end{tabular}
  \end{minipage}\hfill%
  \begin{minipage}[t]{0.54\linewidth}
  \centering
  \vspace{0pt}
  {\textbf{(b) PoseBusters sanity checks (\% pass)}\par}
  \vspace{0.25em}
  \tiny
  \setlength{\tabcolsep}{2pt}
  \begin{tabular}{@{}lrrrrr@{}}
    \toprule
    Test $\uparrow$ & Symphony & Eq. Diff. &
    \shortstack{ADiT\\(joint)} &
    \shortstack{Zatom-1\\(joint, 80M)} &
    \shortstack{Zatom-2\\(joint)} \\
    \midrule
    Atoms connected & 99.92 & 99.88 & 99.70 & 99.98 & 100.00 \\
    Bond angles & 99.56 & 99.98 & 99.85 & 99.95 & 99.93 \\
    Bond lengths & 98.72 & 100.00 & 99.41 & 99.97 & 99.93 \\
    Aromatic ring flat & 100.00 & 100.00 & 100.00 & 100.00 & 100.00 \\
    Double bond flat & 99.07 & 98.58 & 99.98 & 99.99 & 99.99 \\
    Internal energy & 95.65 & 94.88 & 95.86 & 99.78 & 99.61 \\
    No steric clash & 98.16 & 99.79 & 99.79 & 99.81 & 99.98 \\
    \bottomrule
  \end{tabular}
  \par\vspace{0.2em}
  \raggedright
  \end{minipage}
\end{table}
\FloatBarrier

\begin{table}[htbp]
  \centering
  \scriptsize
  \caption{\textbf{MP20 generation results (unrelaxed).}}
  \label{tab:geom-drugs-mp20-benchmark-full}
  \vspace{0.4em}

  \begin{tabular}{@{}lrrrr@{}}
    \toprule
    Model & \shortstack{Overall valid\\$\uparrow$} &
    \shortstack{Unique\\$\uparrow$} & \shortstack{Novel\\$\uparrow$} &
    \shortstack{MetaSUN\\$\uparrow$} \\
    \midrule
    ADiT (joint QM9+MP20) & 90.6 & 87.8 & 26.0 & 1.0 \\
    Crystal-GFN & 51.7 & 51.7 & 51.7 & 0.0 \\
    Crystalformer & 69.9 & 69.4 & 31.8 & 3.1 \\
    LLaMat2-CIF & 84.4 & 81.4 & 30.0 & 2.1 \\
    LLaMat3-CIF & 15.4 & 15.2 & 10.5 & 0.2 \\
    SymmCD & 73.4 & 73.0 & 47.0 & 2.4 \\
    \shortstack[l]{Zatom-1\\(MP20-only, 80M)} & 73.2 & 70.4 & 21.0 & 0.20 \\
    \shortstack[l]{Zatom-1\\(joint QM9+MP20, 80M)} & 88.5 & 84.4 & 8.1 & 0.36 \\
    \shortstack[l]{Zatom-1-L\\(joint QM9+MP20, 160M)} & 95.0 & 90.4 & 3.7 & 0.64 \\
    \midrule
    Zatom-2 (joint QM9+MP20) & 70.24 & 53.44 & 32.24 & 4.92 \\
    \bottomrule
  \end{tabular}
\end{table}
\FloatBarrier

\begin{table}[htbp]
  \centering
  \scriptsize
  \caption{\textbf{Individual PoseBusters checks for GEOM-Drugs generation.}
  Published baseline values are reported from Table~3 of
  \citet{morehead2026zatom1} (arXiv v4). Results are percentages among RDKit-valid molecules.}
  \label{tab:geom-drugs-posebusters-baselines}
  \vspace{0.4em}
  \resizebox{\linewidth}{!}{%
  \begin{tabular}{@{}lrrrrrrr@{}}
    \toprule
    Model & \shortstack{Atoms\\connected} & \shortstack{Bond\\angles} &
    \shortstack{Bond\\lengths} & \shortstack{Aromatic\\ring flat} &
    \shortstack{Double\\bond flat} & \shortstack{Internal\\energy} &
    \shortstack{No steric\\clash} \\
    \midrule
    EQGAT-diff & 84.4 & 86.9 & 87.0 & 87.0 & 87.0 & 86.8 & 82.9 \\
    SemlaFlow & 92.3 & 94.8 & 94.6 & 94.9 & 94.2 & 94.8 & 92.0 \\
    ADiT & 93.0 & 92.3 & 92.5 & 95.4 & 95.3 & 91.3 & 91.8 \\
    TABASCO & 99.9 & 99.2 & 99.4 & 100.0 & 99.8 & 99.4 & 94.3 \\
    Zatom-1 & 99.6 & 99.6 & 99.7 & 100.0 & 100.0 & 99.3 & 96.4 \\
    GEOM-Drugs (reference) & -- & -- & -- & -- & -- & -- & -- \\
    \midrule
    Zatom-2 & 99.18 & 99.51 & 99.17 & 100.00 & 99.97 & 99.31 & 94.34 \\
    \bottomrule
  \end{tabular}%
  }
\end{table}
\FloatBarrier

\subsection{GEOM-Drugs conformer generation benchmark}
  \label{app:geom-drugs-conformer-benchmark}

To further characterize Zatom-2's architecture, we evaluate it on the GEOM-Drugs conformer generation benchmark. While we use force consistency as a representative metric for structure prediction when conformer ensembles are unavailable for individual molecular graphs, this benchmark enables a direct assessment of Zatom-2's conformer prediction capabilities against established methods. We compare against GO-Flow \citep{liu2026goflow}, a specialized conformer prediction model, and test whether Zatom-2's structure generation sampler can recover a conformer ensemble when atom identities and molecular connectivity are fixed. Identical GEOM-Drugs training, validation, and test data splits are used for both Zatom-2 and GO-Flow.

Table~\ref{tab:geom-drugs-zatom2-structure} reports coverage (COV) and matching
(MAT) metrics following \citet{liu2026goflow}, using RDKit symmetry-aware RMSD after removing hydrogens. For each molecule, COV-R (recall) is
the percentage of reference conformers with a generated conformer within
$1.25$~\AA{} RMSD, and COV-P (precision) is the percentage of generated conformers
with a reference conformer within the same threshold. MAT-R is the mean minimum
RMSD from each reference conformer to the generated ensemble, and MAT-P reverses
this direction. Higher COV and lower MAT indicate better performance. We report
the mean of these per-molecule scores across the test set. Zatom-2 achieves higher precision than
GO-Flow \citep{liu2026goflow}, but lower recall. Higher precision is
valuable under a limited sampling budget: a larger fraction of generated
candidates lies close to reference conformations; however, the model does not cover the entire conformational diversity of the reference ensemble as well.

  \begin{table}[!htbp]
    \centering
    \small
    \caption{\textbf{GEOM-Drugs conformer generation evaluation.}
    Values are means across test molecules; COV is reported as a percentage
    and MAT as RMSD in \AA. GO-Flow values are
    from Table~1 of \citet{liu2026goflow}.}
    \label{tab:geom-drugs-zatom2-structure}
    \vspace{0.4em}
    \begin{tabular}{@{}lrrrr@{}}
      \toprule
      Model & COV-R mean $\uparrow$ & MAT-R mean $\downarrow$
      & COV-P mean $\uparrow$ & MAT-P mean $\downarrow$ \\
      \midrule
      GO-Flow & \textbf{94.82} & \textbf{0.7971} & 70.13 & 1.1068 \\
      Zatom-2 & 86.09 & 0.8459 & \textbf{83.07} & \textbf{0.9083} \\
      \bottomrule
    \end{tabular}
  \end{table}
\FloatBarrier

\subsection{MP20 material structure prediction}
\label{app:mp20-material-structure-prediction}

We also benchmark Zatom-2 on periodic material structure prediction using the MP20 dataset. We compare against Crystalite \citep{hadziveljkovic2026crystalite} and report
Pymatgen's \texttt{StructureMatcher} match rate (MR) and normalized RMS displacement (RMSE) \citep{ong2013pymatgen} across matched structures. 

In this setting, Zatom-2 was conditioned on chemical composition, and we used the model to infer lattice geometry (lengths and angles) and fractional coordinates. To calculate match rates, we use PyMatGen's \texttt{StructureMatcher} with
$(\mathrm{ltol},\mathrm{stol},\mathrm{angle\_tol})=(0.3,0.5,10^\circ)$ and
volume rescaling enabled \citep{ong2013pymatgen}. As shown in Table~\ref{tab:mp20-structure-prediction}, without hyperparameter tuning, Zatom-2 approaches Crystalite's match rate (64.00\% vs.\ 66.09\%), but its higher RMSE
(0.1156 vs.\ 0.0337) indicates lower geometric accuracy among matched structures.

\begin{table}[h]
  \centering
  \small
  \caption{\textbf{MP20 structure prediction evaluation.}
  We report the Crystalite result from \citet{hadziveljkovic2026crystalite}.}
  \label{tab:mp20-structure-prediction}
  \vspace{0.4em}
  \setlength{\tabcolsep}{3pt}
  \begin{tabular}{@{}lrr@{}}
    \toprule
    Model & MR (\%) $\uparrow$
    & RMSE $\downarrow$ 
    \\
    \midrule
    Crystalite& \textbf{66.09}
      & \textbf{0.0337} \\
    Zatom-2 & 64.00
      & 0.1156\\
    \bottomrule
  \end{tabular}
\end{table}

\FloatBarrier

\clearpage

\section{Visualization}
\label{app:visualization}

In this section, we visualize Zatom-2's generated samples using MolStar (Mol*, version~5.11.0) \citep{sehnal2021molstar}. Figures~\ref{fig:generated-molecules-molstar}--\ref{fig:generated-proteins-molstar} illustrate molecular geometries, periodic structures, and protein backbones.
Appendix~\ref{app:visualization-embeddings} additionally examines the model's latent representations of reference OMol25 and OMat24 validation examples.

\subsection{Generated molecules}
\label{app:visualization-molecules}

\begin{figure}[!htbp]
  \centering
  \includegraphics[width=\linewidth]{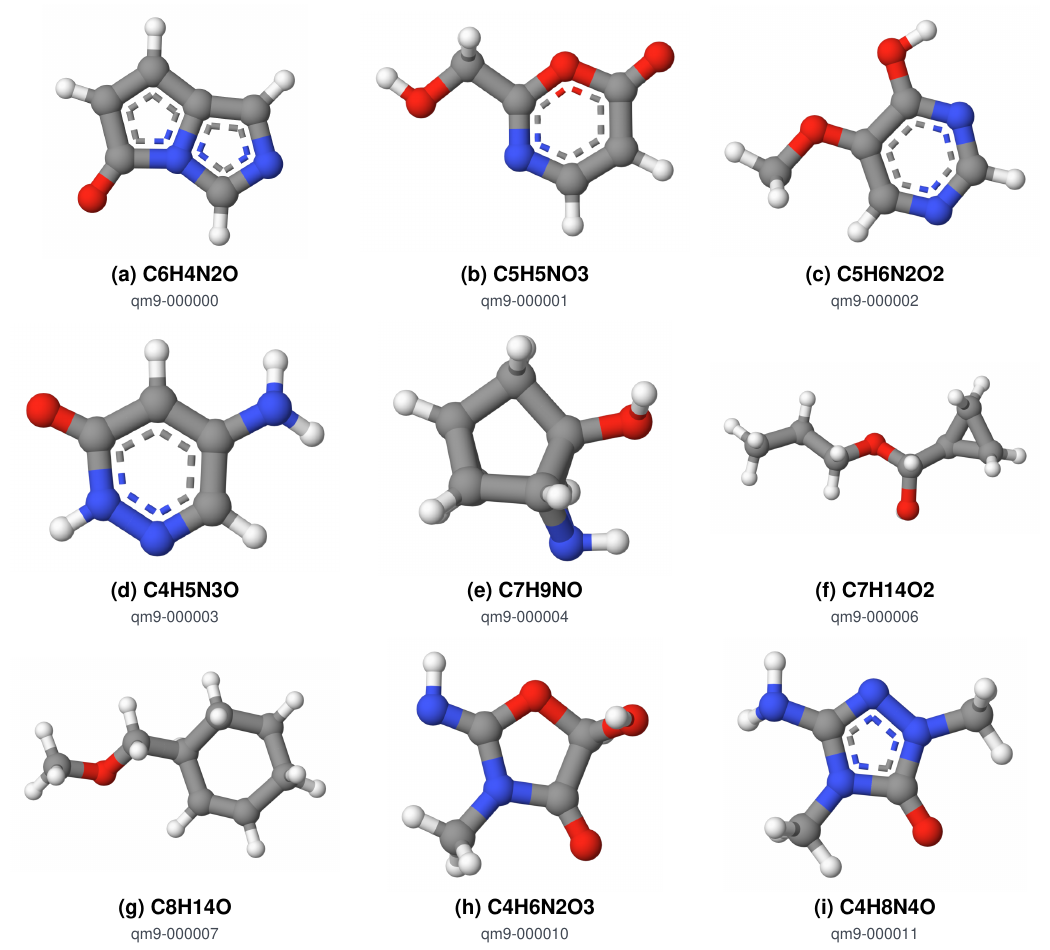}
  \caption{\textbf{Generated QM9 molecules.} MolStar ball-and-stick views of nine Zatom-2 molecules. Carbon is gray, hydrogen white, nitrogen blue, and oxygen red. Bonds are perceived by MolStar from the saved structures (note that coordinates are unchanged).}
  \label{fig:generated-molecules-molstar}
\end{figure}

\paragraph{QM9.}
Figure~\ref{fig:generated-molecules-molstar} shows QM9 samples from the Zatom-2 checkpoint trained jointly on QM9 and MP20, corresponding to the results in Appendix~\ref{app:complete-generation-benchmarks}. We select the first nine distinct samples that yield connected, sanitizable neutral RDKit graphs from their saved coordinates. This selection filter is used only for this illustration.

\begin{figure}[!htbp]
  \centering
  \includegraphics[width=\linewidth,height=0.65\textheight,keepaspectratio]{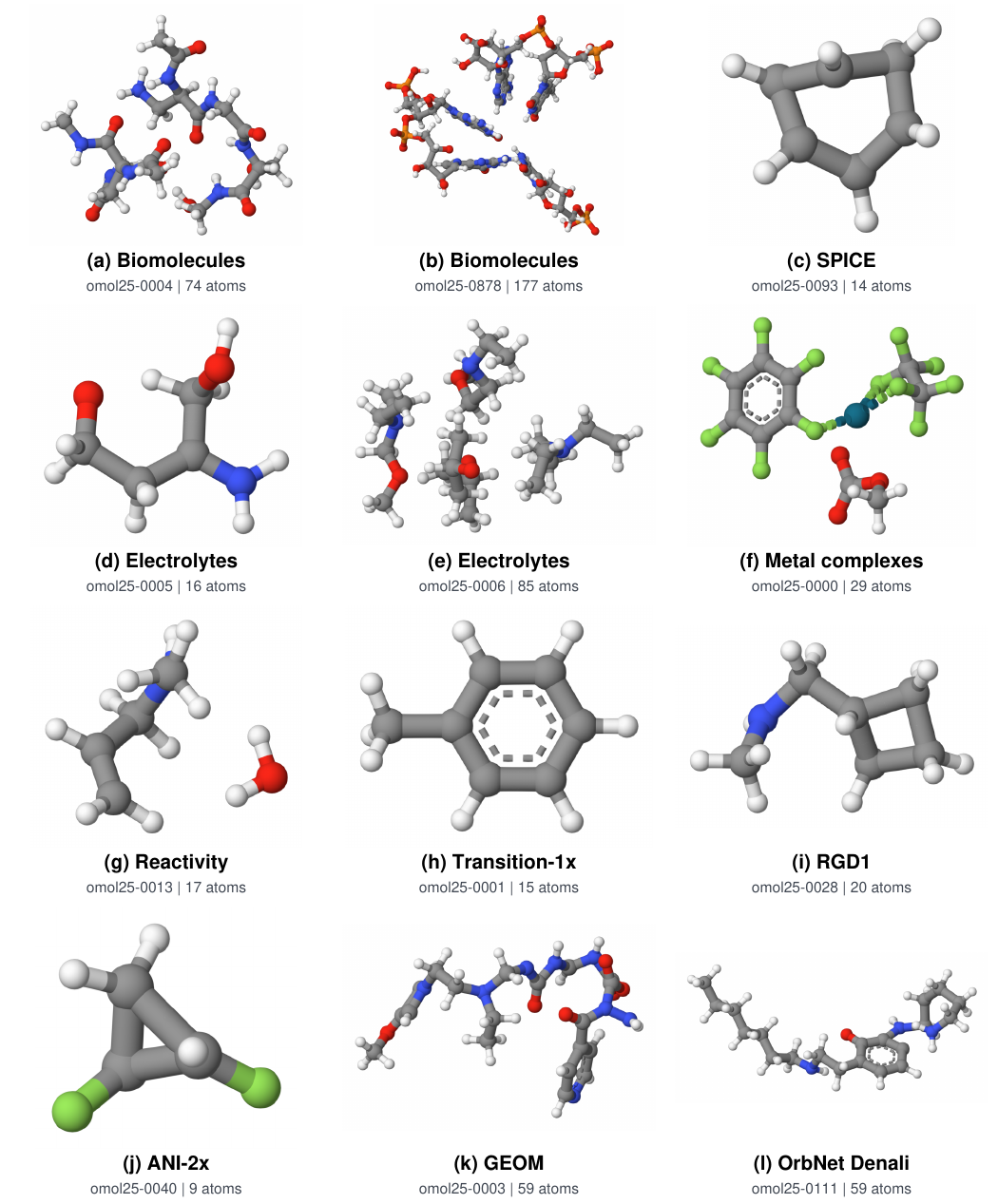}
  \caption{\textbf{Generated OMol25 molecules.} MolStar ball-and-stick views spanning biomolecules, electrolytes, metal complexes, reactivity, and community datasets (labels identify the source subset). Notably, Zatom-2's source category and chemical composition for each sample are not fixed during generation but rather are (re)produced. Note that Transition-1x and RGD1 supply metadata for known reaction path examples, but Zatom-2's generated samples for these subsets are not verified transition states. Atoms are colored by element type, and bonds are perceived by MolStar.}
  \label{fig:generated-omol25-molstar}
\end{figure}

\clearpage
\paragraph{OMol25.}
Figure~\ref{fig:generated-omol25-molstar} shows \emph{force-conditioned} samples from the Zatom-2 model pretrained via joint OMol25/OMat24 generation + structure + MLIP tasks. Here, we stratify by the OMol25 subset supplying each sample's atom count, charge, spin, and mean atomic force norm conditioning value \citep{levine2025omol25}. The twelve generated samples below cover all ten OMol25 subsets in a 1,000-sample batch of molecules drawn from Zatom-2. Original generated atom identities and coordinates are illustrated, without sanitization or post hoc optimization.

\clearpage
\subsection{Generated materials}
\label{app:visualization-materials}

\paragraph{MP20.}
Figure~\ref{fig:generated-materials-molstar} uses the same joint QM9 and MP20 Zatom-2 checkpoint as adopted in Appendix~\ref{app:visualization-molecules}. Here, we select the first nine distinct unit cell compositions considered valid according to LeMat-GenBench's evaluation protocol. 

\begin{figure}[!htbp]
  \centering
  \includegraphics[width=\linewidth]{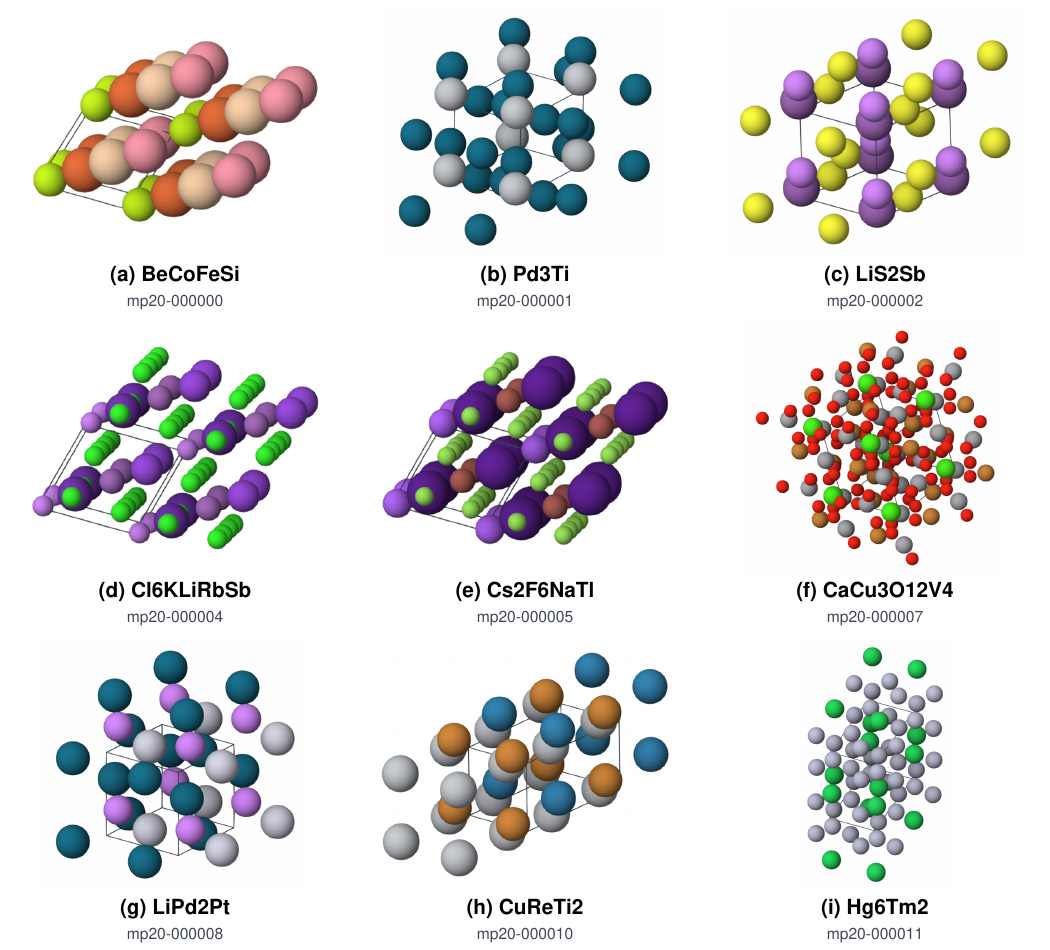}
  \caption{\textbf{Generated MP20 materials.} MolStar views of unrelaxed
  Zatom-2 materials, with atoms colored by element. Each panel displays a
  $2\!\times\!2\!\times\!2$ periodic repetition of the saved cell; a gray
  outline marks one unit cell. Labels report the composition of a single
  saved cell (not a reduced formula) and its sample ID.}
  \label{fig:generated-materials-molstar}
\end{figure}

\begin{figure}[!htbp]
  \centering
  \includegraphics[width=\linewidth]{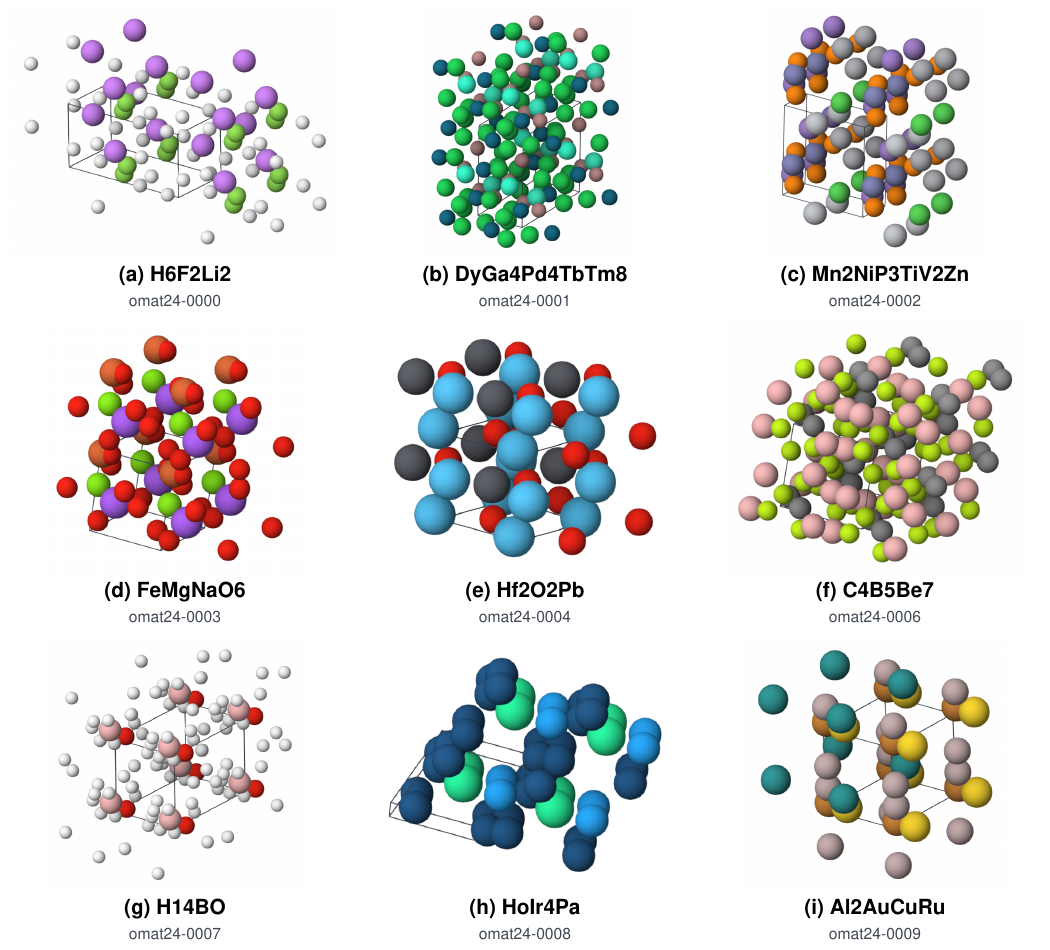}
  \caption{\textbf{Generated OMat24 materials.} MolStar views of nine force-conditioned Zatom-2 samples, with atoms colored by element. Each panel displays a $2\!\times\!2\!\times\!2$ periodic repetition of the generated cell, with one cell outlined in gray. Labels give the composition of the saved cell and an abbreviated sample ID. No relaxation, space group refinement, or symmetrization is applied.}
  \label{fig:generated-omat24-molstar}
\end{figure}

\clearpage
\paragraph{OMat24.}
Figure~\ref{fig:generated-omat24-molstar} shows \emph{force-conditioned} generated materials from the same joint OMol25/OMat24 model as Figure~\ref{fig:generated-omol25-molstar}. For the sake of legibility, we select the first nine distinct unit cell compositions with at most 40 atoms. Unlike the MP20 gallery in Figure~\ref{fig:generated-materials-molstar}, these examples are not filtered by LeMat-GenBench's validity (or MetaSUN) criteria.

\clearpage
\subsection{Generated proteins}
\label{app:visualization-proteins}

\begin{figure}[!htbp]
  \centering
  \includegraphics[width=\linewidth]{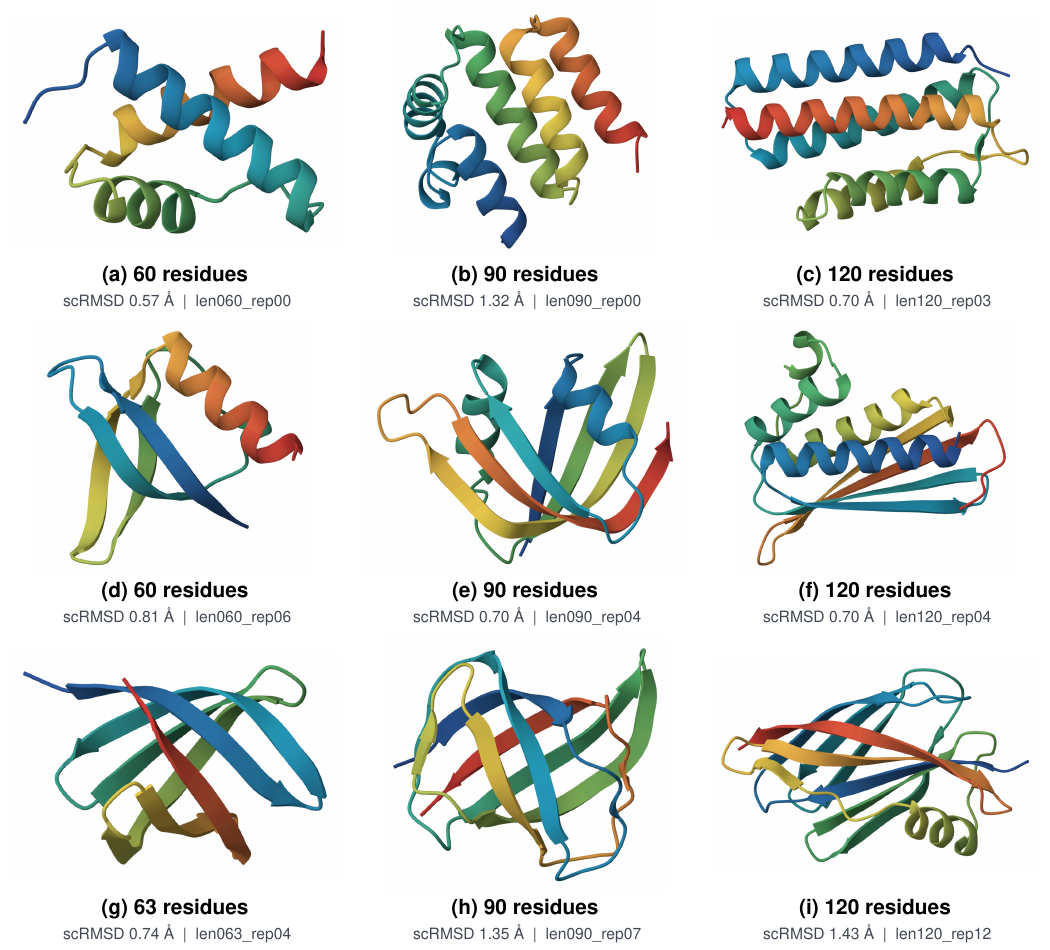}
  \caption{\textbf{Generated SCOPe-2k protein backbones.} MolStar cartoons
  colored from the N-terminus (blue) to the C-terminus (red).
  Rows show helix-rich, mixed, and sheet-rich examples, respectively.
  Labels display backbone length, best-of-eight C$\alpha$ scRMSD in \AA{}, and
  abbreviated sample IDs. The coordinates rendered for each sample are the generated backbones, not their ESMFold2 refolded structures.}
  \label{fig:generated-proteins-molstar}
\end{figure}

\paragraph{SCOPe.}
Figure~\ref{fig:generated-proteins-molstar} shows generated backbones from the joint generation + structure + MLIP pretrained Zatom-2 model (with force conditioning) after 30,000 SCOPe-2k finetuning steps (see Section~\ref{sec:results-protein-transfer} of the main text). Samples come from Section~\ref{sec:results-protein-transfer}'s 1,027-backbone evaluation with 13 samples at each length from 50 to 128 residues. We select designable examples near lengths 60, 90, and 120 in three broad secondary-structure categories: helix-rich, mixed, and sheet-rich. Designability here uses the samples' precomputed best-of-eight C$\alpha$ scRMSD threshold of 1.5~\AA{} (Appendix~\ref{app:metrics-protein}).

\clearpage
\subsection{Latent representations of OMol25 and OMat24}
\label{app:visualization-embeddings}

\begin{figure}[p]
  \centering
  \includegraphics[width=\linewidth,height=0.85\textheight,keepaspectratio]{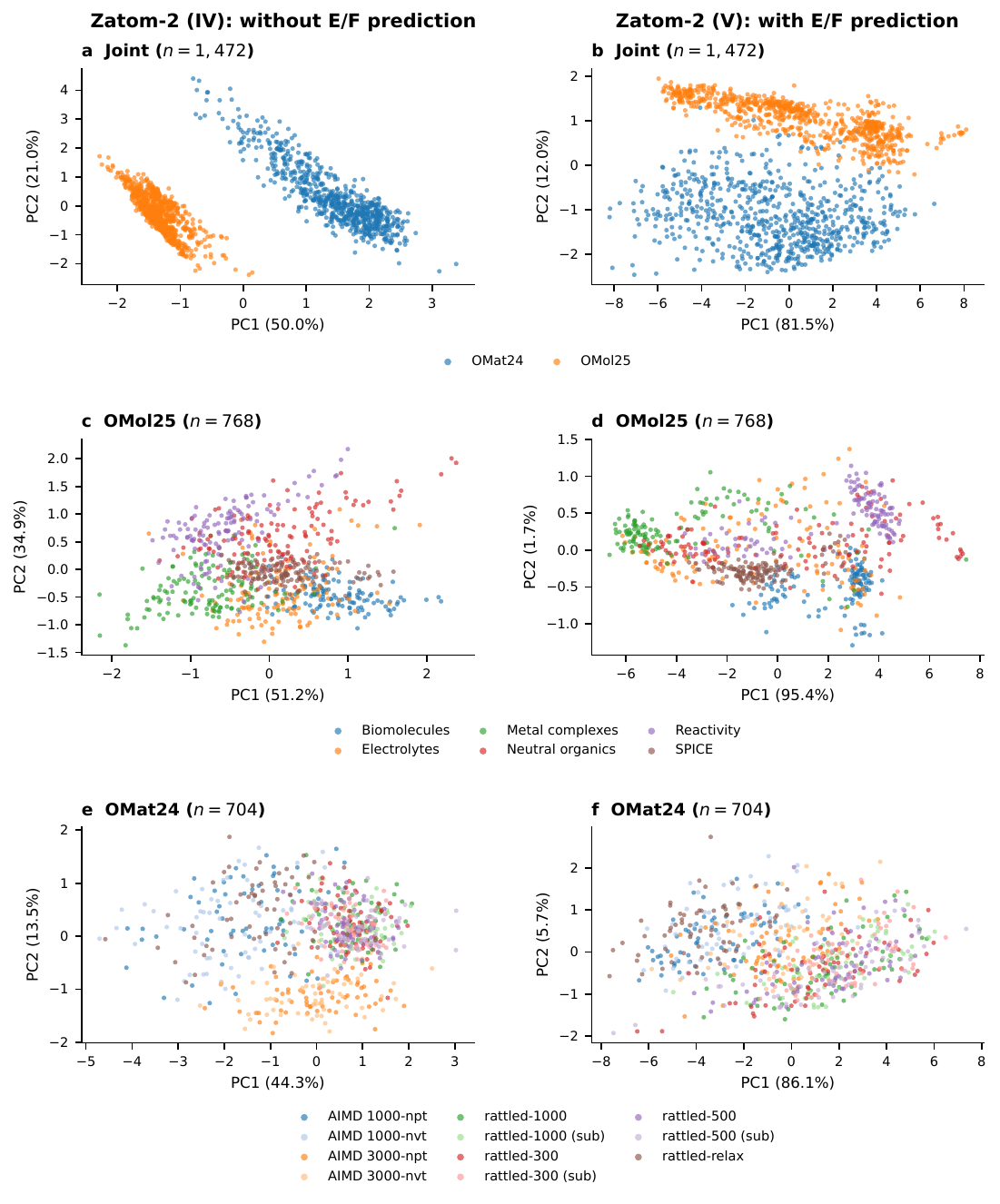}
  \caption{\textbf{Base Zatom-2 system embeddings across OMol25 and OMat24.}
  Left: Zatom-2 (IV), without energy/force prediction; right: Zatom-2 (V), with it.
  Each point represents the same validation example in both columns, mean-pooled
  over all atoms. Rows show results for the joint datasets (a,b), OMol25 subsets (c,d), and
  OMat24 subsets (e,f), with matching colors across checkpoints. Each panel has
  an independent PCA fit; axis percentages report variance. Note that axes
  are not aligned between models and that ``(sub)'' denotes subsampled OMat24 data.}
  \label{fig:subset-embeddings-pca}
\end{figure}

\paragraph{System representations.} 
To visualize how Zatom-2 organizes its molecular and material latent space with and without energy/force prediction during pretraining, we extract embeddings from the base-sized models, Zatom-2 (IV) and (V) of Table~\ref{tab:pretraining-quality} in the main text, using their EMA weights and the same reference OMol25 and OMat24 validation examples. Notably, we evaluate both models under the clean-input conditions used for MLIP prediction, where we provide a clean flow endpoint $t=1$ with known atom identities, Euclidean coordinates, and ground-truth charge/spin metadata and run one model forward pass. Here, force conditioning is disabled. We extract the final backbone representations before the task heads, allowing the same procedure to be used for model (IV), which has no energy/force prediction heads. Each embedding data point then averages the final 128-dimensional atom representations $\mathbf{q}_{s,i}$ (n.b., $Q_L$ elsewhere in the text) over all $N_s$ atoms of example $s$, including hydrogens:
\begin{equation}
  \mathbf{h}_s = \frac{1}{N_s}\sum_{i=1}^{N_s}\mathbf{q}_{s,i}.
\end{equation}
Regarding data composition, we sample examples from these validation datasets without replacement, allowing 2--512 atoms total: 128 examples per OMol25 subset and 64 per OMat24 subset, totaling 768 molecules and 704 materials. Note that the ``neutral organics'' subset here pools the \texttt{ani2x}, \texttt{orbnet\_denali}, and \texttt{geom\_orca6} subsets into one subset for the sake of visualization. Importantly, both models receive identical examples and preprocessing. Figure~\ref{fig:subset-embeddings-pca} uses separate centered principal component analysis (PCA) fits for each of its panels, applying equal weight to each example as well as no channel standardization or vector normalization.

For joint PCA projections in panels (a,b) of Figure~\ref{fig:subset-embeddings-pca}, most molecules and materials are separated visually in both models' latent spaces, while Zatom-2 \emph{with} energy/force prediction notably shrinks the distance between molecules and materials and even partially overlaps a subset of these inputs. To better understand this observation, in panels (c,d) OMol25 subsets occupy partially distinct regions, whereas in panels (e,f) OMat24 subsets overlap substantially (n.b., these material subsets describe different sampling procedures rather than distinct material types). Comparing the two columns, energy/force prediction pretraining is associated with a greater concentration of variance in the first principal component: 95.4\% versus 51.2\% for OMol25 and 86.1\% versus 44.3\% for OMat24, with versus without energy/force prediction. In the next section, we investigate this phenomenon in more atomistic detail.

\begin{figure}[p]
  \centering
  \includegraphics[width=\linewidth,height=0.80\textheight,keepaspectratio]{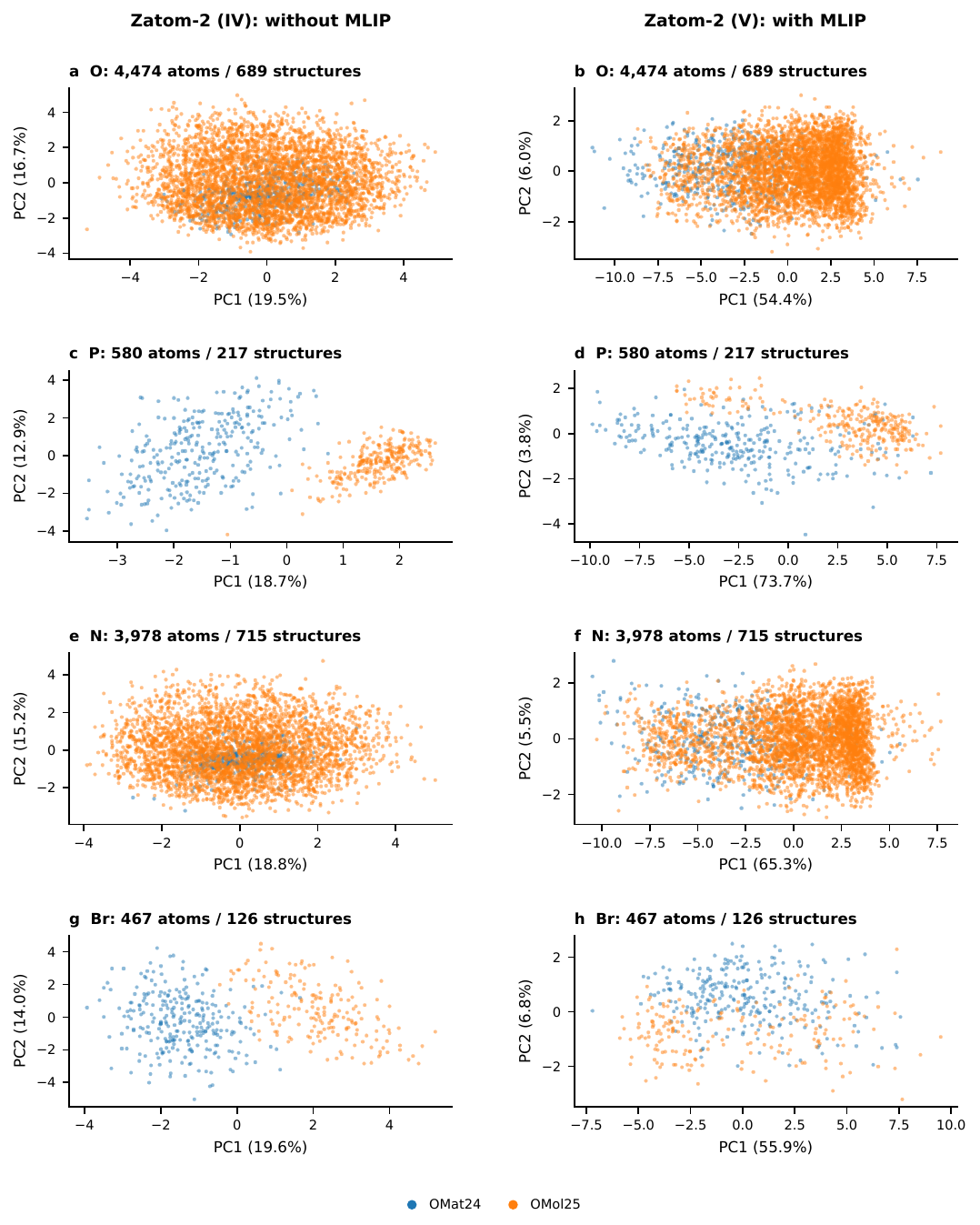}
  \caption{\textbf{Base Zatom-2 atom embeddings across OMol25 and OMat24.}
  Atom-level counterpart of Figure~\ref{fig:subset-embeddings-pca}(a,b), separately for O, P, N, and Br, selected by shared structure coverage.
  Left: Zatom-2 (IV), without MLIP supervision; right: Zatom-2 (V), with it.
  Each point is one atom, colored by domain, with identical atoms in both columns.
  Titles give sampled atom and structure counts. Each panel fits centered PCA independently; axis percentages report variance, and axes are not aligned between models.}
  \label{fig:element-embeddings-joint}
\end{figure}

\begin{figure}[p]
  \centering
  \includegraphics[width=\linewidth,height=0.80\textheight,keepaspectratio]{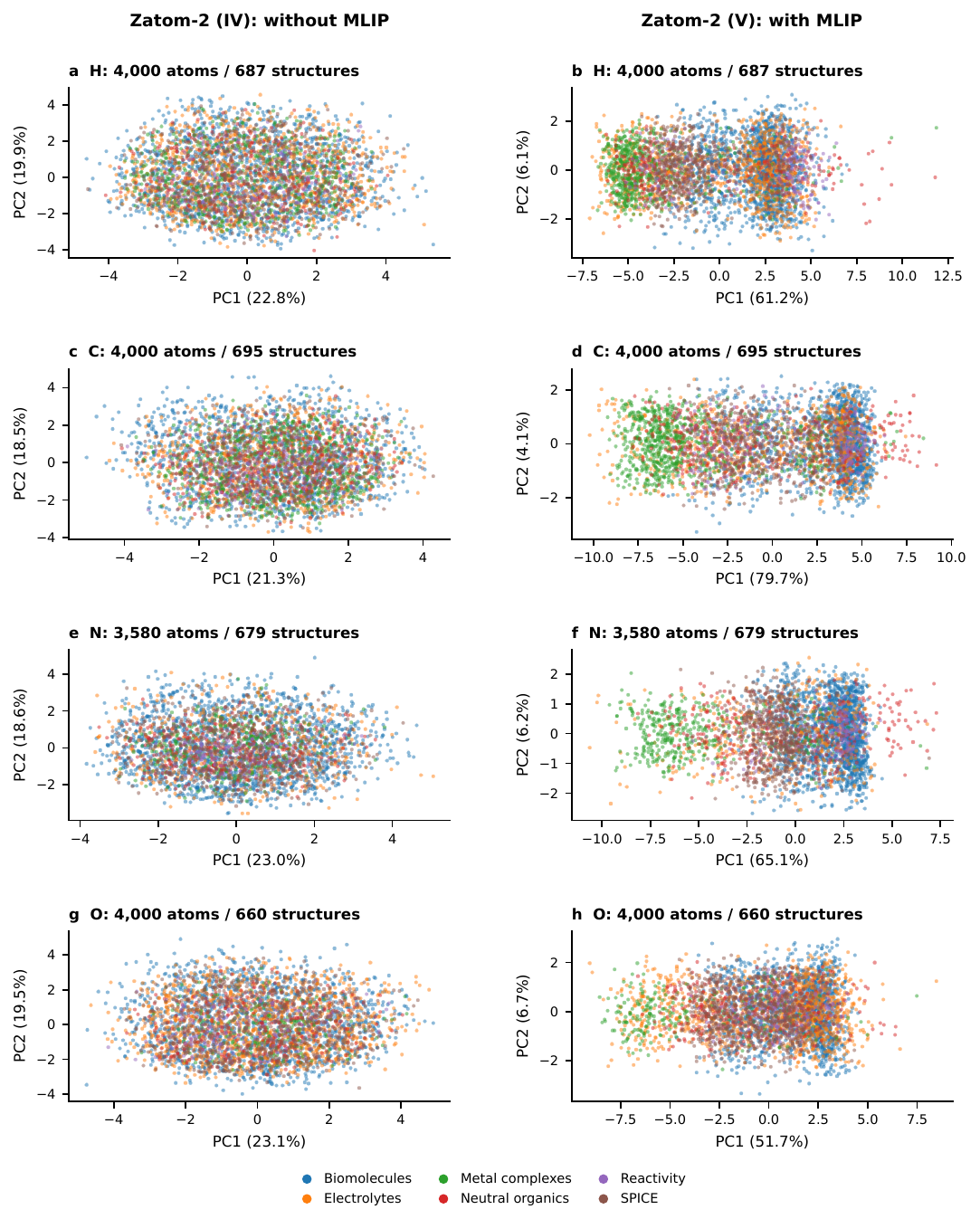}
  \caption{\textbf{Base Zatom-2 atom embeddings within OMol25.}
  Atom-level counterpart of Figure~\ref{fig:subset-embeddings-pca}(c,d) for H, C, N, and O, the four elements present in the most sampled molecules.
  Left: without MLIP supervision; right: with it. Colors indicate the same OMol25 subsets as in Figure~\ref{fig:subset-embeddings-pca}.
  Each point is an individual atom, with an independent PCA fit per panel.}
  \label{fig:element-embeddings-omol25}
\end{figure}

\begin{figure}[p]
  \centering
  \includegraphics[width=\linewidth,height=0.80\textheight,keepaspectratio]{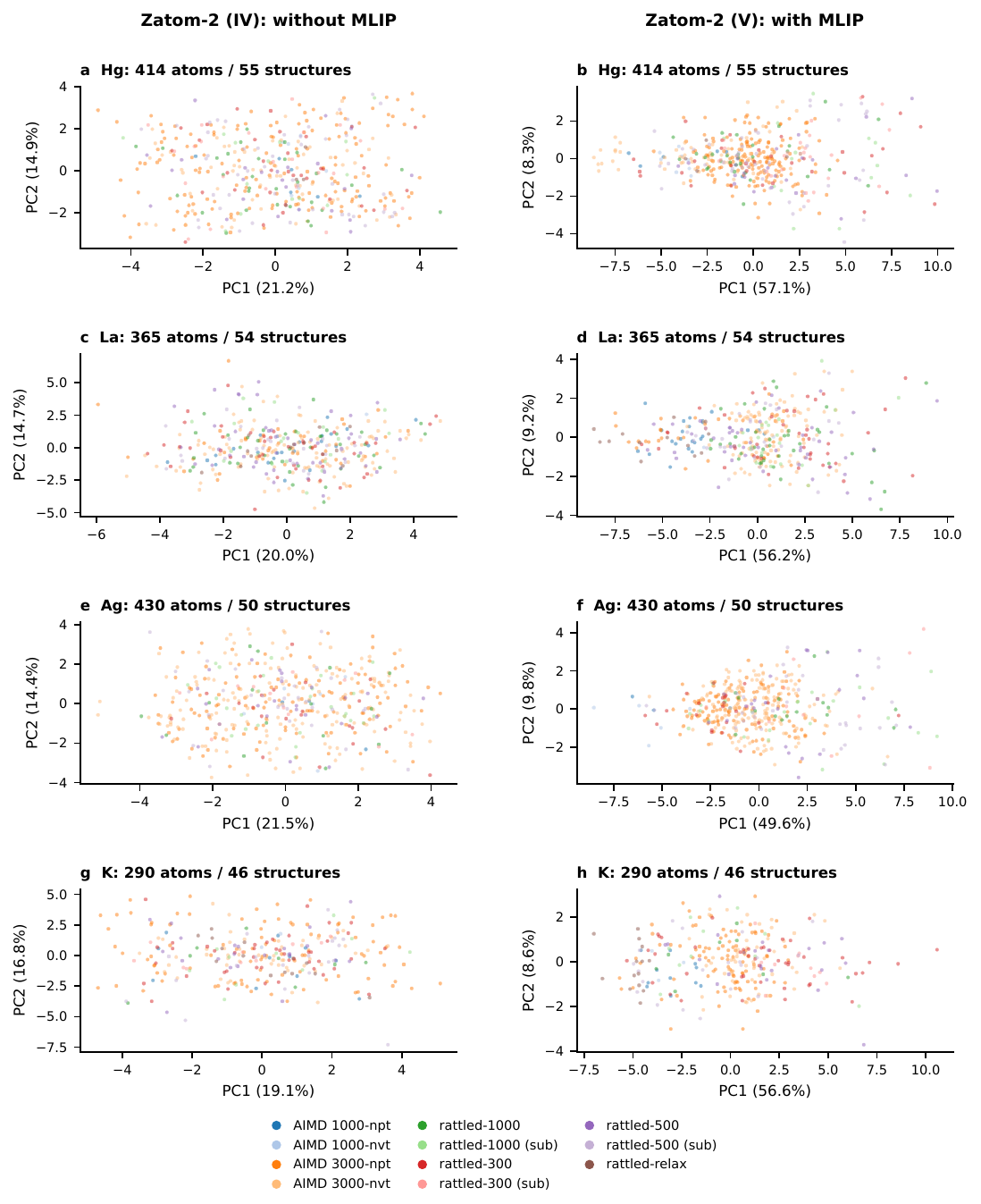}
  \caption{\textbf{Base Zatom-2 atom embeddings within OMat24.}
  Atom-level counterpart of Figure~\ref{fig:subset-embeddings-pca}(e,f) for Hg, La, Ag, and K, selected by structure coverage in the sampled validation materials.
  Left: without MLIP supervision; right: with it. Each point is an atom, colored by the corresponding OMat24 subset; ``(sub)'' denotes a subsampled subset.
  Identical atoms appear in both columns, with independent PCA fits.
  MLIP supervision concentrates more variance in PC1.}
  \label{fig:element-embeddings-omat24}
\end{figure}

\clearpage
\paragraph{Atom representations.}
To examine how energy/force prediction pretraining affects the organization of individual atom representations, we visualize the final backbone vectors $\mathbf{q}_{s,i}$ separately for each element, using the same models, validation examples, and MLIP clean-input conditions as above. Figures~\ref{fig:element-embeddings-joint}--\ref{fig:element-embeddings-omat24} show the per-element embeddings for OMol25 and OMat24 jointly, for OMol25 subsets, and for OMat24 subsets, respectively. Each comparison uses identical atoms across models and independent PCA fits, with element types selected by how common they are in the validation examples. Separating elements allows us to examine atomistic embedding organization while retaining variation that is averaged out in the system (e.g., per-molecule) representations.

In the joint dataset projections of Figure~\ref{fig:element-embeddings-joint}, molecular and material atom representations partially overlap, with the degree of overlap varying by element type and pretraining configuration. Energy/force prediction pretraining concentrates more variance in the first principal component, as observed for the system representations in Figure~\ref{fig:element-embeddings-joint}. However, most molecular and material atoms remain largely distinguishable in the full representation space with energy/force supervision, indicating that energy/force supervision may promote less obfuscation of atoms from different atomistic domains.

The clearest pretraining effect appears within OMol25 (Figure~\ref{fig:element-embeddings-omol25}), where energy/force prediction pretraining produces more distinct subset-associated regions for H, C, N, and O. This organization also extends beyond the PCA projections: in the original 128-dimensional latent space, the fraction of ten nearest neighbors sharing an atom's subset label increases by 12.6--14.9\% for these elements, considering only same-element neighbors from other validation examples and giving each query example equal weight. 

\clearpage

\section{Broader Impacts \& Limitations}
\label{app:broader-impacts-and-limitations}

\subsection{Broader impacts}
\label{app:broader-impacts}

Shared atomistic representations could accelerate the discovery of medicines,
catalysts, and energy materials and reduce the need to train separate models
for each task. These benefits must be weighed against the energy cost of
training and oracle evaluation, unequal access to computing resources, and
biases in the chemical and structural coverage of the training data.
Generative chemistry and protein design capabilities also carry dual-use
risks, including the design of harmful compounds or biological agents.
Responsible use calls for application-specific safety assessment, appropriate
screening of proposed designs, and experimental validation before deployment.

\subsection{Limitations}
\label{app:limitations}

This work limits training to the 4M and 1M subsets of OMol25 and OMat24, respectively. The full datasets contain 100M examples each, and we leave training on the entire data for future work. Secondly, training with force conditioning enables Zatom-2 to distinguish clearly between low- and high-force regimes, but as the experiments in Section~\ref{sec:results-pretraining} highlight, within-regime force conditioning is not well captured by our force conditioning method on all subsets of the data. We leave improvements on force conditioning mechanisms, including conditioning on the per-atom force vector, to future work. Moreover, the UMA-based metric employed for force consistency is an error-prone machine learning method; however, we do wish to highlight that the MAEs between UMA-calculated mean force norms of OMol25 and OMat24 examples and their dataset mean force norm labels are 0.00587 and 0.02462~eV\,\AA$^{-1}$, respectively (see Appendix~\ref{app:force-sc-metrics}). We also note that our protein generation experiments are performed on a small, although diverse, dataset of proteins, and that future experiments can explore transfer to large-scale protein datasets via finetuning. Lastly, while we treat force and energy prediction as auxiliary tasks for generative training and do not intend to match specialized MLIPs' performance, results in Appendix~\ref{sec:results-mlip-finetuning} leave room for improvement, which we leave for future work. 

\end{document}